\documentclass[10pt,journal,compsoc]{IEEEtran}
\ifCLASSOPTIONcompsoc
  \usepackage[nocompress]{cite}
\else
  \usepackage{cite}
\fi
\ifCLASSINFOpdf
\else
\fi
\usepackage{times}
\usepackage{epsfig}
\usepackage{graphicx}
\usepackage{amsmath}
\usepackage{amssymb}

\usepackage{amsmath,amsfonts,bm}

\def\eqref#1{equation~\ref{#1}}

\def\1{\bm{1}}

\newcommand{\test}{\mathcal{D_{\mathrm{test}}}}

\DeclareMathAlphabet{\mathsfit}{\encodingdefault}{\sfdefault}{m}{sl}
\SetMathAlphabet{\mathsfit}{bold}{\encodingdefault}{\sfdefault}{bx}{n}

\usepackage{amsfonts}
\usepackage{array}
\usepackage{booktabs}
\usepackage{mathtools}
\usepackage{microtype}
\usepackage{multirow}
\usepackage{nicefrac}
\usepackage{url}
\usepackage{amsfonts}
\usepackage{xcolor}
\usepackage{pifont}
\usepackage{float}

\usepackage{tabulary}

\usepackage{graphbox}

\newcommand{\cmark}{\ding{52}}%
\newcommand{\xmark}{\ding{56}}%

\newcommand{\vect}[1]{\mathbf{#1}}

\def\eg{\textit{e.g.}~}
\def\ie{\textit{i.e.}}

\newcolumntype{x}[1]{>{\centering\arraybackslash}p{#1pt}}
\newcolumntype{Y}{>{\centering\arraybackslash}X}

\usepackage{tabularx}
\newcolumntype{Y}{>{\centering\arraybackslash}X}

\usepackage{graphicx,overpic,xcolor}
\usepackage{longtable}
\usepackage{colortbl}
\usepackage[caption=false]{subfig}
\usepackage{wrapfig}

\usepackage{wrapfig}

\newlength\savewidth\newcommand\shline{\noalign{\global\savewidth\arrayrulewidth
  \global\arrayrulewidth 1pt}\hline\noalign{\global\arrayrulewidth\savewidth}}
  
\newcommand{\tablestyle}[2]{\setlength{\tabcolsep}{#1}\renewcommand{\arraystretch}{#2}\centering\footnotesize}

\begin{document}
%
\title{Dynamic Graph Message Passing Networks \\
for Visual Recognition}
%
%
%
%

\author{Li Zhang,
        Mohan Chen,
        Anurag Arnab,
        Xiangyang Xue,
        and~Philip H.S. Torr
\IEEEcompsocitemizethanks{
\IEEEcompsocthanksitem Li Zhang is the corresponding author with the School of Data Science, Fudan University.\protect\\
E-mail: lizhangfd@fudan.edu.cn
\IEEEcompsocthanksitem 
Anurag Arnab is with Google Research. 
Mohan Chen and Xiangyang Xue are with School of Computer Science, Fudan University. 
Philip H.S. Torr is with the Department of Engineering Science, University of Oxford.}
}

%
%

\markboth{
}%
{Shell \MakeLowercase{\textit{et al.}}: Bare Demo of IEEEtran.cls for Computer Society Journals}
%



\IEEEtitleabstractindextext{%
\begin{abstract}
Modelling long-range dependencies is critical for scene understanding tasks in computer vision. Although convolution neural networks (CNNs) have excelled in many vision tasks, they are still limited in capturing long-range structured relationships as they typically consist of layers of local kernels. A fully-connected graph, such as the self-attention operation in Transformers, is beneficial for such modelling, however, its computational overhead is prohibitive. In this paper, we propose a dynamic graph message passing network, that significantly reduces the computational complexity compared to related works modelling a fully-connected graph. This is achieved by adaptively sampling nodes in the graph, conditioned on the input, for message passing. Based on the sampled nodes, we dynamically predict node-dependent filter weights and the affinity matrix for propagating information between them. This formulation allows us to design a self-attention module, and more importantly a new Transformer-based backbone network, that we use for both image classification pretraining, and for addressing various downstream tasks (\eg object detection, instance and semantic segmentation). Using this model,  we show significant improvements with respect to strong, state-of-the-art baselines on four different
tasks. Our approach also outperforms fully-connected graphs while using substantially fewer floating-point operations and parameters. Code and models will be made publicly available at \url{https://github.com/fudan-zvg/DGMN2}.
\end{abstract}

}

\maketitle

\IEEEdisplaynontitleabstractindextext

%
\IEEEpeerreviewmaketitle

\IEEEraisesectionheading{\section{Introduction}
\label{sec:intro}}

\IEEEPARstart{C}{apturing} long-range dependencies is crucial for complex scene understanding tasks such as image classification, semantic segmentation, instance segmentation and object detection.
Although convolutional neural networks (CNNs) have excelled in a wide range of scene understanding tasks \cite{alexnet, vgg, resnet}, they are still limited by their ability to capture these long-range interactions. 
To improve the capability of CNNs in this regard, a recent, popular model Non-local networks~\cite{wang2018nonlocal} proposes a generalisation of the attention model of~\cite{vaswani2017attention} and achieves significant advance in several computer vision tasks. 

\begin{figure*}
	\centering
	\includegraphics[width=0.92\linewidth]{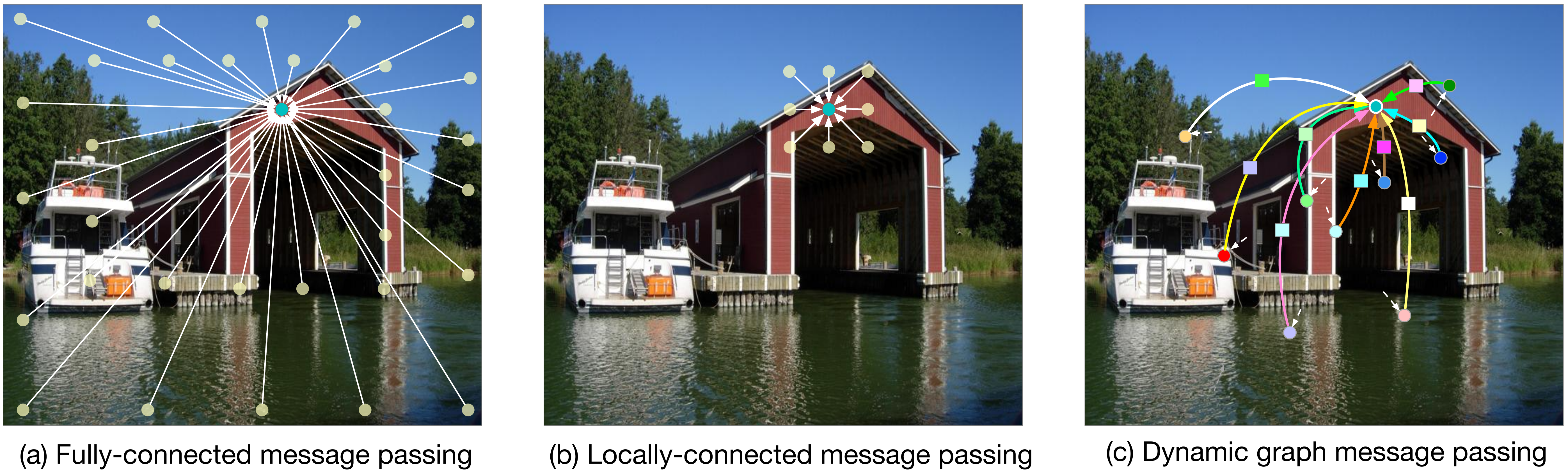}
	\caption{\small Contextual information is crucial for complex scene understanding tasks. To recognise the ``boathouse'', one needs to consider the ``boat'' and the ``water'' next to it.
	Fully-connected message passing models (a) are able to obtain this information, but are prohibitively expensive. Furthermore, they capture a lot of redundant information (\ie ``trees'' and ``sky'').
	Locally-connected models (b) are more efficient, but miss out on important context.
	Our proposed approach (c), dynamically samples a small subset of relevant feature nodes based on a \textit{learned} dynamic sampling scheme, 
	\ie~the \textit{learned} position-specific random walk (indicated by the white dashed arrow lines), and also dynamically predicts filter weights and affinities (indicated by unique edge and square colors.), which are both conditioned on the sampled feature nodes.
	}
	\label{fig:teaser}
\end{figure*}

Non-local networks essentially model pairwise structured relationships among all feature elements in a feature map to produce the attention weights which are used for feature aggregation.
Considering each feature element as a node in a graph, Non-local networks effectively model a fully-connected feature graph and thus have a quadratic inference complexity with respect to the number of the feature elements.
This is infeasible for dense prediction tasks on high-resolution imagery, as commonly encountered in semantic segmentation~\cite{Cityscapes}.
Moreover, in dense prediction tasks, capturing relations between all pairs of pixels is usually unnecessary due to the redundant information contained within the image (Fig.~\ref{fig:teaser}).
Simply subsampling the feature map to reduce the memory requirements is also suboptimal, as such na\"ive subsampling would result in smaller objects in the image not being represented adequately.

Graph convolution networks (GCNs)~\cite{kipf2016semi, gilmer2017neural} -- which propagate information along graph-structured input data -- can alleviate the computational issues of non-local networks to a certain extent.
However, this stands only if local neighbourhoods are considered for each node.
Employing such local-connected graphs means that the long-range contextual information needed for complex vision tasks such as segmentation and detection~\cite{rabinovich2007objects,oliva2007, bai2009geodesic} 
will only be partially captured.
Along this direction, GraphSAGE~\cite{hamilton2017inductive} introduced an efficient graph learning model based on graph sampling.
However, the proposed sampling method considered a uniform sampling strategy along the spatial dimension of the input, and was independent of the actual input.
Consequently, the modelling capacity was restricted as it assumed a static input graph where the neighbours for each node were fixed and filter weights were shared among all nodes.

To address the aforementioned shortcomings, we propose a novel dynamic graph message passing network (DGMN) model, targeting effective and efficient deep representational learning with joint modeling of two key dynamic properties as illustrated in Fig.~\ref{fig:teaser}. Our contribution is twofold: (i) We dynamically sample the neighbourhood of a node from the feature graph, conditioned on the node features. 
Intuitively, this learned sampling allows the network to efficiently gather long-range context by only selecting a subset of the most relevant nodes in the graph; 
(ii) Based on the nodes that have been sampled, we further dynamically predict node-dependant, and thus \textit{position specific}, filter weights and also the affinity matrix, which are used to propagate information among the feature nodes via message passing. 
The dynamic weights and affinities are especially beneficial to specifically model each sampled feature context, leading to more effective message passing.
Both of these dynamic properties are jointly optimised in a single model, and we modularise the DGMN as a network layer for simple deployment into existing networks.

We demonstrate the proposed model on the tasks of image classification, semantic segmentation, object detection and instance segmentation on the challenging ImageNet~\cite{imagenet}, Cityscapes~\cite{Cityscapes} and COCO~\cite{lin2014microsoft} datasets.
We achieve significant performance improvements over the fully-connected model~\cite{wang2018nonlocal,vaswani2017attention}, while using substantially fewer floating point operations (FLOPs).
Specifically, ``plugging'' our module into existing networks, we show considerable improvements with respect to strong, state-of-the-art baselines on three different vision downstream tasks and backbone architectures.
Furthermore, we design a new Transformer backbone network based on the building block of our formulation.
It achieves leading performance on Imagenet classification task and the pretrained backbone network also supports various vision downstream tasks.

A preliminary version of this work was published in~\cite{zhang2020dynamic}.
We have extended our conference version as follows:
\textbf{(i)}
We introduce a modified process of dynamic message passing calculation and present a simplified version of our prior formulation;
We further recast our formulation to take a Transformer perspective and implement the message updating process into an efficient and effective Transformer layer.
\textbf{(ii)}
Based on the building block of our formulation, we propose a new backbone network DGMN2 with multi-scale feature representation which is very compatibility in supporting various downstream vision tasks.
The experiment on upstream ImageNet~\cite{imagenet} classification task show that all the variants of DGMN2 (\ie,~DGMN2-Tiny, DGMN2-Small, DGMN2-Medium and DGMN2-Large respectively) can surpass the conventional convolution-based backbones such as ResNet/ResNeXt~\cite{resnet,xie2017aggregated}, and Transformer-based backbone such as DeiT~\cite{touvron2020training} and PVT~\cite{wang2021pyramid}.
\textbf{(iii)}
With the pretrained backbone network DGMN2, we show considerable improvements with respect to strong, state-of-the-art baselines on three different vision downstream tasks such as object detection, instance segmentation and semantic segmentation.

\section{Related work}
\label{sec:related}

An early technique for modelling context for computer vision tasks involved conditional random fields.
In particular, the DenseCRF model~\cite{krahenbuhl2011efficient} was popular as it modelled interactions between all pairs of pixels in an image.
Although such models have been integrated into neural networks~\cite{zheng2015conditional,arnab2016higher,arnab2018conditional, xu2017learning}, they are limited by the fact that the pairwise potentials are based on simple handcrafted features. 
Moreover, they mostly model discrete label spaces, and are thus not directly applicable in the feature learning task since feature variables are typically continuous. 
Coupled with the fact that CRFs are computationally expensive, CRFs are no longer used for most computer vision tasks.

A complementary technique for increasing the receptive field of CNNs was to use dilated convolutions~\cite{deeplabv1,yu2015multi}.
With dilated convolutions, the number of parameters does not change, while the receptive field grows exponentially if the dilation rate is linearly increased in successive layers.
Other modifications to the convolution operation include deformable convolution~\cite{dai2017deformable,zhu2018deformable}, which learns the offset with respect to a predefined grid from which to select input values.
However, the weights of the deformable
convolution filters do not depend on the selected input, and are in fact shared across all different positions.
In contrast, our dynamic sampling aims to sample over the whole feature graph to obtain a large receptive field, and the predicted affinities and the weights for message passing are \textit{position specific} and \textit{conditioned} on the dynamically sampled nodes. 
Our model is thus able to better capture position-based semantic context to enable more effective message passing among feature nodes.

Another line of related research targets the graph learning~\cite{scarselli2008graph,kazi2022differentiable,wang2019dynamic,fey2018splinecnn,monti2017geometric,li2019deepgcns} on the irregular structured and geometric input.
A neural network module EdgeConv is proposed in~\cite{wang2019dynamic} to learn dynamic graphs with k-nearest neighbor (kNN) on the irregular point clouds data.
DeepGCN~\cite{li2019deepgcns} proposes to enlarge the receptive field on graph networks with dilated convolution. 
DGM~\cite{kazi2022differentiable} introduces a differentiable graph module that learns dynamic adjacency matrices.

The idea of sampling graph nodes has previously been explored in GraphSAGE~\cite{hamilton2017inductive}.
Crucially, 
GraphSAGE simply uniformly samples nodes. 
In contrast, our sampling strategy is \textit{learned} based on the node features. 
Specifically, we first sample the nodes uniformly in the spatial dimension, and then dynamically predict \textit{walks} of each node conditioned on the node features. 
Furthermore, GraphSAGE does not consider our second important property, \ie,~the dynamic prediction of the affinities and the message passing kernels.

We also note that \cite{jia2016dynamic} developed an idea of ``dynamic convolution'', that is predicting a dynamic convolutional filter for each feature position.
More recently,~\cite{wu2019pay} further reduced the complexity of this operation in the context of natural language processing with lightweight grouped convolutions.
Unlike~\cite{jia2016dynamic,wu2019pay}, we present a graph-based formulation, and jointly learn dynamic weights and dynamic affinities, which are conditioned on an \emph{adaptively sampled} neighbourhood for each feature node in the graph using the proposed dynamic sampling strategy for effective message passing.

Recently, there have been a number of variants of Transformer~\cite{vaswani2017attention,dosovitskiy2020image} in computer vision.
ViT~\cite{dosovitskiy2020image} adopts a pure-Transformer architecture with \textit{query}, \textit{key} and \textit{value} self-attention for image classification and demonstrates its effectiveness for replacing convolution neural networks.
However, there is a bottleneck in such Transformer based models, namely its quadratic complexity in both computation and memory usage.
PVT~\cite{wang2021pyramid} uses downsampling operation to shrink the number of \textit{key} and \textit{value} tokens.
Similarly, MViT~\cite{fan2021multiscale} also performs pooling to reduce the number of tokens on which self-attention is applied.
Swin Transformer~\cite{liu2021swin}, on the other hand, predefines a local window region for each \textit{query} point to linearlize the attention computation.
We also instantiate our formulation into an efficient and effective Transformer layer and present a new Transformer backbone network DGMN2 in
supporting various vision downstream tasks.
Different to above Transformer-based backbone networks, we introduce a graph-based formulation to adaptively sample feature nodes for effective message passing.
\section{Dynamic graph message passing networks}
\label{sec:method}
\subsection{Problem definition and notation}
Given an input feature map interpreted as a set of feature vectors, \ie,~$\vect{F} = \{\vect{f}_i\}_{i=1}^N$  with $\vect{f}_i \in \mathbb{R}^{1\times C}$, where $N$ is the number of pixels and $C$ is the feature dimension, our goal is to learn a set of refined latent feature vectors $\vect{H} = \{\vect{h}_i\}_{i=1}^N$ by utilising hidden structured information among the feature vectors at different pixel locations.
$\vect{H}$ has the same dimension as the observation $\vect{F}$. To learn such structured representations, we convert the feature map into a graph
$\mathcal{G}=\{\mathcal{V}, \mathcal{E}, A\}$ with $\mathcal{V}$ as its nodes, $\mathcal{E}$ as its edges and $A$ as its adjacency matrix.
Specifically, the nodes of the graph represent the latent feature vectors, \ie,~$\mathcal{V} = \{\vect{h}_i\}_{i=1}^N$, and $A \in \mathbb{R}^{N\times N}$ is a binary or learnable matrix with self-loops describing the connections between nodes.
In this work, we propose a novel dynamic graph message passing network~\cite{gilmer2017neural} for deep representation learning, which refines each graph feature node by passing messages on the graph $\mathcal{G}$.
Different from existing message passing neural networks considering a fully- or locally-connected static graph~\cite{wang2018nonlocal, gilmer2017neural}, we propose a dynamic graph network model with two dynamic properties, \ie,~\textit{dynamic sampling} of graph nodes to approximate the full graph distribution, 
and \textit{dynamic prediction} of node-conditioned filter weights and affinities, in order to achieve more efficient and effective message passing.
\subsection{Graph message passing neural networks for deep representation learning}
\label{sec:method_mpgraph}

Message passing neural networks (MPNNs)~\cite{gilmer2017neural} present a generalised form of graph neural networks such as graph convolution networks~\cite{kipf2016semi}, gated graph sequential networks~\cite{li2015gated} 
and graph attention networks~\cite{velivckovic2017graph}.
In order to model structured graph data, in which latent variables are represented as nodes on an undirected or directed graph, feed-forward inference is performed through a message passing phase followed by a readout phase upon the graph nodes.
The message passing phase usually takes $T$ iteration steps to update feature nodes, while the readout phase is for the final prediction, \eg, graph classification with updated nodes.
In this work, we focus on the message passing phase for learning efficient and effective feature refinement, since well-represented features are critical in all downstream tasks.
The message passing phase consists of two steps,  \ie,~a message calculation step $M^t$ and a message updating step $U^t$.
Given a latent feature node $\vect{h}_i^{(t)}$ at an iteration $t$, for computational efficiency, we consider a locally connected node field with $v_i \subset \mathcal{V}$ and $v_i \in \mathbb{R}^{(K\times C)}$, where $K \ll N$ is the number of sampled nodes in $v_i$. Thus we can define the message calculation step for node $i$ operated locally as
\begin{align}
\vect{m}^{(t+1)}_{i} &= M^t \left(A_{i,j},  \{\vect{h}_1^{(t)}, \cdots,  \vect{h}_K^{(t)} \}, \vect{w}_j \right) \nonumber \\ 
&= \sum_{j \in \mathcal{N}(i)} A_{i,j} \vect{h}_j^{(t)} \vect{w}_j, \ \,
\label{equ:messagecalculation}
\end{align}
where $A_{i,j} = A[i, j]$ describes the connection relationship \ie,~the affinity between latent nodes $\vect{h}_i^{(t)}$ and $\vect{h}_j^{(t)}$, $\mathcal{N}(i)$ denotes a self-included neighborhood of the node $\vect{h}_i^{(t)}$ which can be derived from $v_i$ and $\vect{w}_j \in \mathbb{R}^{C \times C}$ is a transformation matrix for message calculation on the hidden node $\vect{h}_j^{(t)}$.
The message updating function $U^t$ then updates the node $\vect{h}_i^{(t)}$ with a linear combination of the calculated message and the observed feature $\vect{f}_i$ at the node position $i$ as: 
\begin{equation}
\vect{h}_i^{(t+1)} = U^t \left(\vect{f}_i, \vect{m}^{(t+1)}_{i} \right) = \sigma \left( \vect{f}_i+ \alpha^m_i \vect{m}^{(t+1)}_{i} \right),
\label{messageupdating}
\end{equation}
where $\alpha^m_i$ of a learnable parameter for scaling the message, and the operation $\sigma(\cdot)$ is a non-linearity function, \eg, ReLU. By iteratively performing message passing on each node with $T$ steps, we obtain a refined feature map $\vect{H}^{(T)}$ as output.

\begin{figure*}
	\centering
	\includegraphics[width=0.85\linewidth]{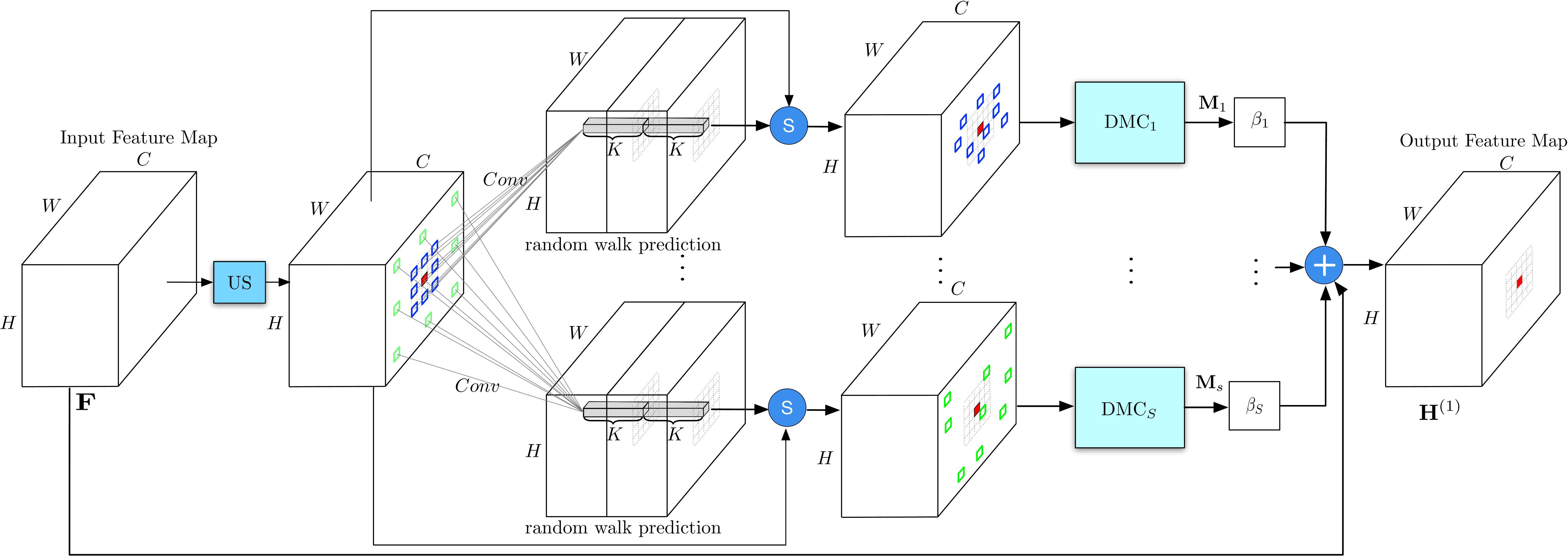}
	\caption{
	\small 
	Overview of our proposed dynamic graph message passing network (DGMN).
	The neighbourhood used to update the feature representation of each node (we show a single node with a red square) is predicted dynamically conditioned on each input.
	This is done by first uniformly sampling (denoted by ``US'') a set of $S$ neighbourhoods around each node.
	Each neighbourhood contains $K$ (\eg $3 \times 3$) sampled nodes.
	Here, the blue nodes were sampled with a low sampling rate, and the green ones with a high sampling rate.
	Walks are predicted (conditioned on the input) from these uniformly sampled nodes, denoted by the $\textcircled{s}$ symbol representing the random walk sampling operation described in Sec.~\ref{sec:method_sampling}.
	$\mathrm{DMC}_1, \cdots, \mathrm{DMC}_S$ and $\beta_1, \cdots, \beta_S$ denotes $S$ dynamic message calculation operations and $S$ message scaling parameters, respectively. 
	The DMC module is detailed in Figure~\ref{fig:dmc}.
	The symbol $\oplus$ indicates an element-wise addition operation.	
	}
	\label{fig:framework}
\end{figure*}
\subsection{From a fully-connected graph to a dynamic sampled graph}
\label{sec:method_sampling}

A fully-connected graph typically contains many connections and parameters, which, in addition to computational overhead, results in redundancy in the connections, and also makes the network optimisation more difficult especially when dealing with limited training data.
Therefore, as in Eq.~\ref{equ:messagecalculation}, a local node connection field is considered in the graph message passing network. 
However, in various computer vision tasks, such as detection and segmentation, learning deep representations capturing both local and global receptive fields is important for the model performance~\cite{rabinovich2007objects,oliva2007,li2019global,hou2020strip}.
To maintain a large receptive field while utilising much fewer parameters than the fully-connected setting, we further explore dynamic sampling strategies in our proposed graph message passing network.
We develop a uniform sampling scheme, which we then extend to a predicted random walk sampling scheme, aimed at reducing the redundancy found in a fully-connected graph.
This sampling is performed in a dynamic fashion, meaning that for a given node $\vect{h}_i$, we aim to sample an optimal subset of $v_i$ from $\mathcal{V}$ to update $\vect{h}_i$ via message passing as shown in Fig.~\ref{fig:framework}.

\noindent\textbf{Multiple uniform sampling for dynamic receptive fields.} 
Uniform sampling is a commonly used strategy for graph node sampling~\cite{leskovec2006sampling} based on Monte-Carlo estimation.
To approximate the distribution of $\mathcal{V}$, we consider a set of $S$ uniform sampling rates $\vect{\varphi}$ with $\vect{\varphi} = \{\rho_q\}_{q=1}^S$, where $\rho_q$ is a sampling rate.
Let us assume that the latent feature nodes are located in a $P$-dimensional space $\mathbb{R}^{P}$. 
For instance, $P = 2$  for images considering the $x$- and $y$-axes. 
For each latent node $\vect{h}_i$, a total of $K$ neighbouring nodes are sampled from $\mathbb{R}^P$. 
The receptive field of $v_i$ is thus determined by $\rho_q$ and $K$.
Note that the sampling rate $\rho_q$ corresponds to the ``dilation rate'' often used in convolution~\cite{yu2015multi} and is thus able to capture a large receptive field whilst maintaining a small number of connected nodes.
Thus we can achieve much lower computational overhead compared with fully-connected message passing in which typically all $N$ nodes are used when one of the nodes is updated. 
Each node receives $S$ complementary messages from distinct receptive fields for updating as
\begin{align}
\vect{m}^{(t+1)}_{i} &= \sum_{q} \sum_{j \in \mathcal{N}_{q}(i)} \beta_q A_{i,j}^{q} \vect{h}_j^{(t)} \vect{w}_j^q,
\label{equ:uniformsampling}
\end{align}
where $\beta_q$ is a weighting parameter for the message from the $q$-th sampling rate and $q = 1, \cdots, S$.
$A^q$ denotes an adjacency matrix formed under a sampling rate $\rho_q$, with 
$A_{i,j}^{q}$, $\vect{w}_j^q$ and $\mathcal{N}_{q}(i)$ defined analogously. 
The uniform sampling scheme acts as a linear sampler based on the spatial distribution while not considering the original feature distribution of the hidden nodes, \ie,~sampling independently of the node features. 
Eq.~\ref{messageupdating} can still be used to update the nodes.

\noindent\textbf{Learning position-specific random walks for node-dependant adaptive sampling.} 
To take into account the feature data distribution when sampling nodes, we further present a random walk strategy upon the uniform sampling. 
Walks around the uniformly sampled nodes could sample the graph in a non-linear and adaptive manner, and we believe that it could facilitate learning better approximation of the original feature distribution. 
The ``random'' here refers to the fact that the walks are predicted in a data-driven fashion from stochastic gradient descent.
Given a matrix, $v_i^q \in \mathbb{R}^{K\times C}$, constructed from $K$ uniformly sampled nodes under a sampling rate, $\rho_q$, the random walk of each node is further estimated based on the feature data of the sampled nodes.
Given the $P$-dimensional space where the nodes distribute ($P=2$ for images), let us denote $\bigtriangleup \vect{d}_j^q \in \mathbb{R}^{P\times 1}$ as predicted walks from a uniformly sampled node $\vect{h}_j$ with $j \in \mathcal{N}_{q}(i)$. 
The node walk prediction can then be performed using a matrix transformation as
\begin{equation}
\bigtriangleup \vect{d}_j^q = \vect{W}_{i,j}^q v_i^q + \vect{b}_{i,j}^q,
\label{equ:deformablewalk}
\end{equation}
where $\vect{W}_{i,j}^q \in \mathbb{R}^{P\times (K\times C)}$ and $\vect{b}_{i,j}^q \in \mathbb{R}^{P\times 1}$ are the matrix transformation parameters, which are learned separately for each node $v_i^q$. 
With the predicted walks, we can obtain a new set of adaptively sampled nodes ${v'}_i^{q}$, 
and generate the corresponding adjacency matrix ${A'}^q$, which can be used to calculate the messages as
\begin{align}
\vect{m}^{(t+1)}_{i} &= \sum_{q} \sum_{j \in \mathcal{N}_{q}(i)} \beta_q {A'}_{i,j}^{q} \varrho \left(\vect{h'}_{j}^{(t)}~| ~\mathcal{V},~j,~\bigtriangleup \vect{d}_j^q \right) \vect{w}_j^q,  
\label{equ:deformablemessage}
\end{align}
where the function $\varrho(\cdot)$ is a bilinear sampler~\cite{jaderberg2015spatial} which samples a new feature node $\vect{h'}_{j}^{(t)}$ around $\vect{h}_{j}^{(t)}$ given the predicted walk $\bigtriangleup \vect{d}_j^q$ and the whole set of graph vertexes $\mathcal{V}$.

\begin{figure*}
	\centering
	\includegraphics[width=0.85\linewidth]{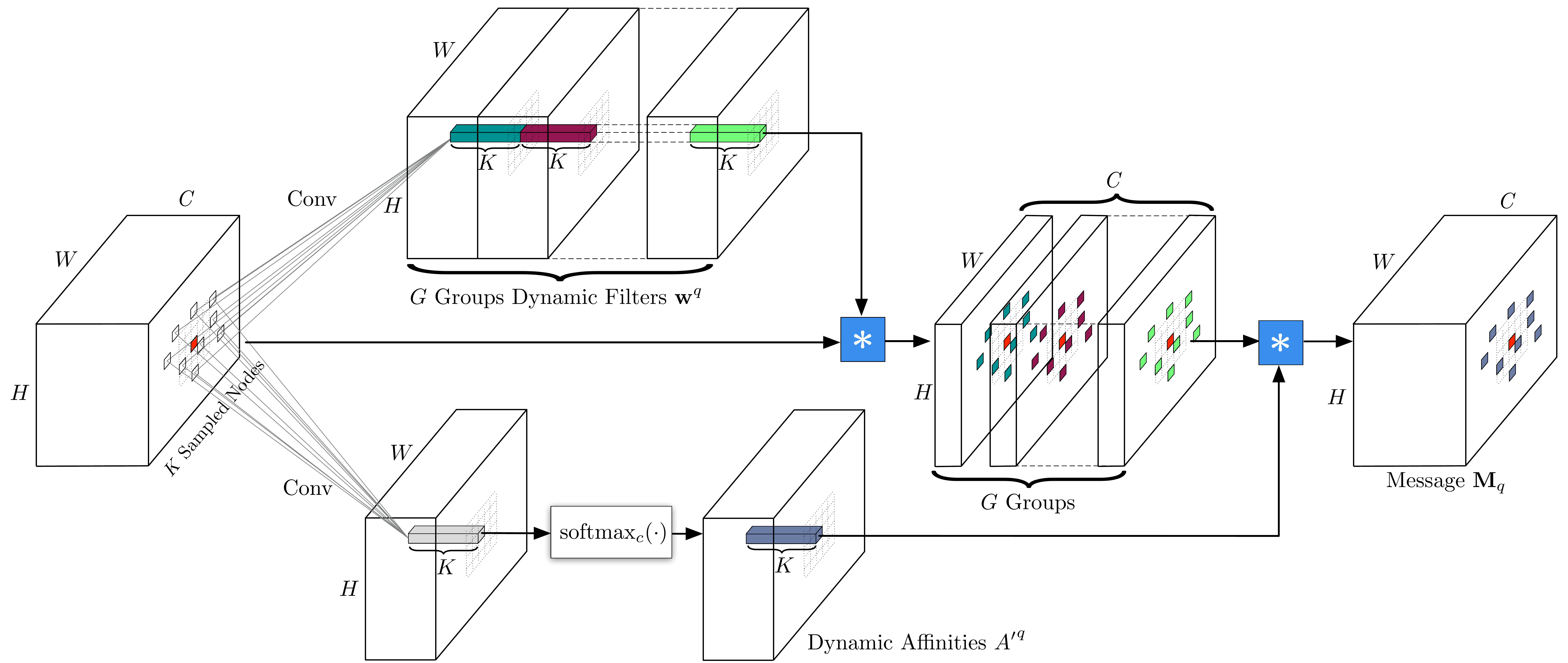}
	\caption{\small Schematic illustration of the proposed dynamic message passing calculation (DMC) module. 
	The small red square indicates the receiving node whose message is calculated from its neighbourhood, \ie~the sampled $K$
	(\eg $3 \times 3$) 
	features nodes. 
	The module accepts a feature map as input and produces its corresponding message map. 
	The symbol $\ast$ denotes group convolution operation using the dynamically predicted and position specific group kernels and affinities.}
	\label{fig:dmc}
\end{figure*}
\subsection{Joint learning of node-conditioned dynamic filters and affinities}
\label{sec:method_dynamic_weights}

In the message calculation formulated in Eq.~\ref{equ:deformablemessage}, the set of weights $\{\vect{w}_j^q\}_{j=1}^K$ of the filter is shared for each adaptively sampled node field ${v'}_i^q$. 
However, since each ${v'}_i^q$ essentially defines a node-specific local feature context, it is more meaningful to use a node-conditioned filter to learn the message for each hidden node $\vect{h'}_i^{(t)}$. 
In additional to the filters for the message calculation in Eq.~\ref{equ:deformablemessage}, the affinity ${A'}_{i,j}$ of any pair of nodes $\vect{h'}_i^{(t)}$ and $\vect{h'}_j^{(t)}$ could be also be predicted and should also be conditioned on the node field ${v'}_i^q$, since the affinity reweights the message passing only  in ${v'}_i^q$. 
As shown in Fig.~\ref{fig:dmc}, we thus use matrix transformations to simultaneously estimate the dynamic filter and affinity which are both conditioned on ${v'}_i^q$,
\begin{equation}
\{\vect{w}_j^q,~{A'}_{i,j}^q\} = \vect{W}_{i,j}^{k,A} {v'}_i^q + \vect{b}_{i,j}^{k,A}, 
\label{equ:dynamicweights}
\end{equation} 
\begin{equation}
{A'}_{i,j}^q = \mathrm{softmax}_{c} ({A'}_{i,j}^q) =\dfrac{\exp({{A'}_{i,j}^q})}{\sum_{l \in \mathcal{N}_q(i)} \exp({{A'}_{i,l}^q})},
\label{equ:aff_softmax}
\end{equation} 
where the function $\mathrm{softmax}_c(\cdot)$ denotes a softmax operation along the channel axis, which is used to perform a normalisation on the estimated affinity ${A'}_{i,j}^q \in \mathbb{R}^1$. 
$\vect{W}_{i,j}^{k,A} \in \mathbb{R}^{(G\times C + 1)\times (K \times C)}$ and $\vect{b}_{i,j}^{k,A} \in \mathbb{R}^{(G\times C + 1)}$ are matrix transformation parameters.
To reduce the number of the filter parameters, we consider grouped convolutions~\cite{chollet2017xception} with a set of $G$ groups split from the total $C$ feature channels, and $G \ll C$, \ie,~each group of $C/G$ feature channels shares the same set of filter parameters. 
The predicted dynamic filter weights and the affinities are then used in Eq.~\ref{equ:deformablemessage} for dynamic message calculation.

\subsection{Modular instantiation}

Fig.~\ref{fig:framework} and \ref{fig:dmc} shows how our proposed dynamic graph message passing network (DGMN) can be implemented in a neural network.
The proposed module accepts a single feature map $\vect{F}$ as input, which can be derived from any neural network layer.
$\vect{H}^{(0)}$ denotes an initial state of the latent feature map, $\vect{H}$, and is initialised with $\vect{F}$. 
$\vect{H}$ and $\vect{F}$ have the same dimension, \ie,~$\vect{F}, \vect{H} \in \mathbb{R}^{H\times W \times C}$, where $H$, $W$ and $C$ are the height, width and the number of feature channels of the feature map respectively.
We first define a set of $S$ uniform sampling rates (we show two uniform sampling rates in Fig.~\ref{fig:framework} for clarity).
The uniform and the random walk sampler sample the nodes from the full graph and return the node indices for subsequent dynamic message calculation (DMC) in Fig.~\ref{fig:dmc}. 
The matrix transformation $\vect{W}_{i,j}^q$ to estimate the node random walk in Eq.~\ref{equ:deformablewalk} is implemented by a 
$3 \times 3$ 
convolution layer~\cite{dai2017deformable}.
Note that other sampling strategies could also be flexibly employed in our framework.

The sampled feature nodes are processed along two data paths: one for predicting the node-dependant dynamic affinities ${A'}^q \in \mathbb{R}^{H\times W\times K}$ and another path for dynamic filters $\vect{w}^q \in \mathbb{R}^{H\times W \times K \times G}$ where $K$ 
(\eg, $3 \times 3$) 
is the kernel size for the receiving node.
The matrix transformation $\vect{W}_{i,j}^{k,A}$ used to jointly predict the dynamic filters and affinities in Eq~\ref{equ:dynamicweights} is implemented by a
$3 \times 3$ 
convolution layer.
Message $\vect{M}_q \in \mathbb{R}^{H\times W\times C}$ corresponding to the $q$-th sampling rate is 
then scaled to perform a linear combination with the observed feature map $\vect{F}$, to produce a refined feature map $\vect{H}^{(1)}$ as output. 
To balance performance and efficiency, as in existing graph-based feature learning models~\cite{wang2018nonlocal,graph_reason, lin2015deeply, zhang2019dual}, we also perform $T = 1$ iteration of message updating. 

\subsection{Discussion}
Our approach is related to deformable convolution~\cite{dai2017deformable, zhu2018deformable} and Non-local~\cite{wang2018understanding}, but has several key differences:

A fundamental difference to deformable convolution is that it only learns the offset dependent on the input feature while the filter weights are fixed for all inputs. 
In contrast, our model learns the random walk, weight and affinity as all being dependent on the input.
This property makes our weights and affinities position-specific whereas deformable convolution shares the same weight across all convolution positions in the feature map.
Moreover, \cite{dai2017deformable,zhu2018deformable} only consider $3 \times 3$ local neighbours at each convolution position.
In contrast, our model, for each position, learns to sample a set of $K$ nodes (where $K \gg 9$) for message passing globally from the whole feature map. 
This allows our model to capture a larger receptive field than deformable convolution.

Whilst Non-local also learns to refine deep features, it uses a self-attention matrix to guide the message passing between each pair of feature nodes.
In contrast, our model learns to sample graph feature nodes to capture global feature information efficiently.
This dynamic sampling reduces computational overhead, whilst still being able to improve upon the accuracy of Non-local across multiple tasks as shown in our experiments (Section~\ref{sec:exp}.
We now detail the extension of our model to transformer architectures next.

\begin{figure*}
	\centering
	\subfloat[Schematic illustration of the DGMN2 layer.
	The red arrowline denotes \textit{dynamic weights} and 
	the blue arrowline
	denotes \textit{dynamic affinities}.]
	{
	\includegraphics[width=0.5\linewidth]{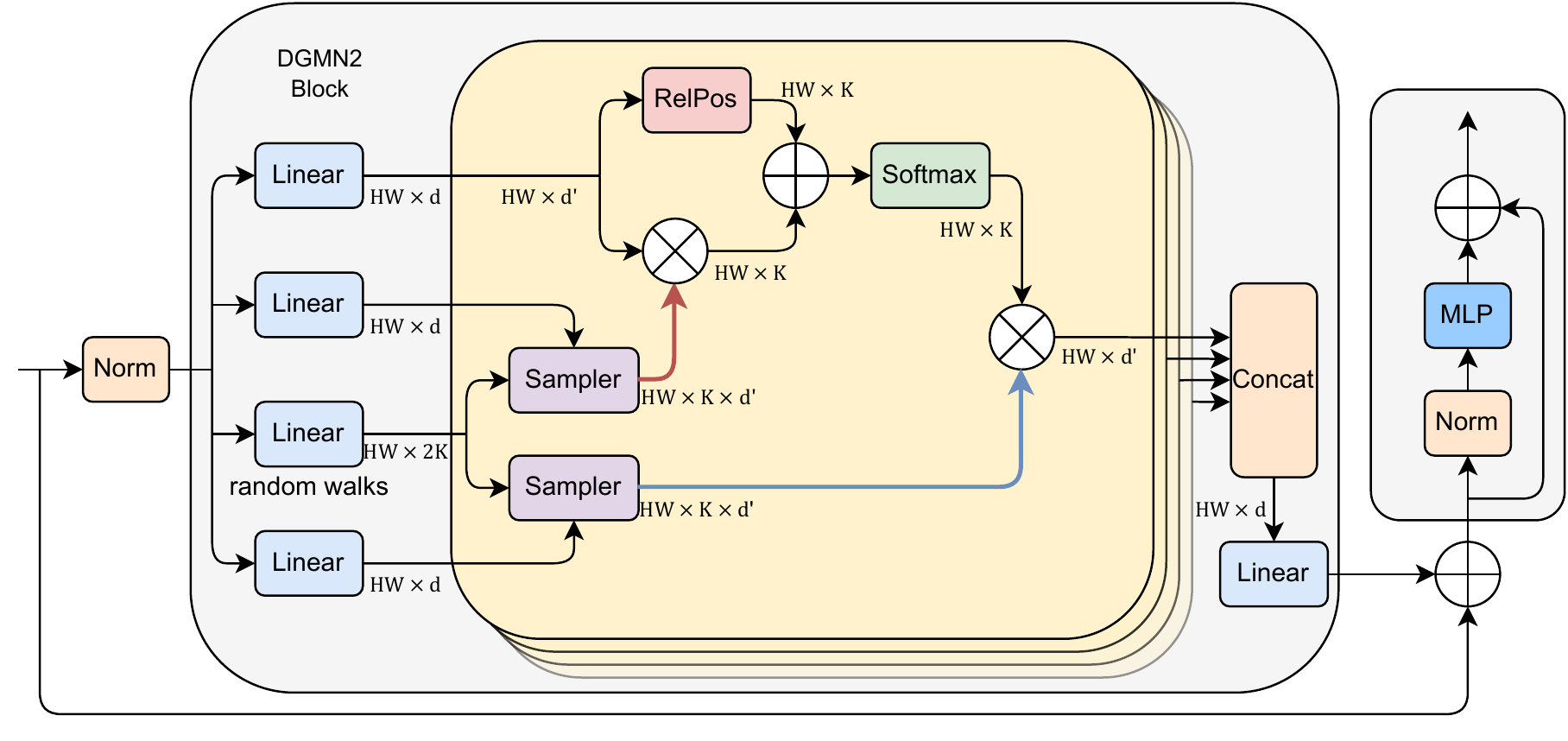}
	\label{fig:dgmn2_framework_block}
	}
	\hspace{8mm}
	\subfloat[The RelPos operation in DGMN2 block.]{
	\includegraphics[width=0.43\linewidth]{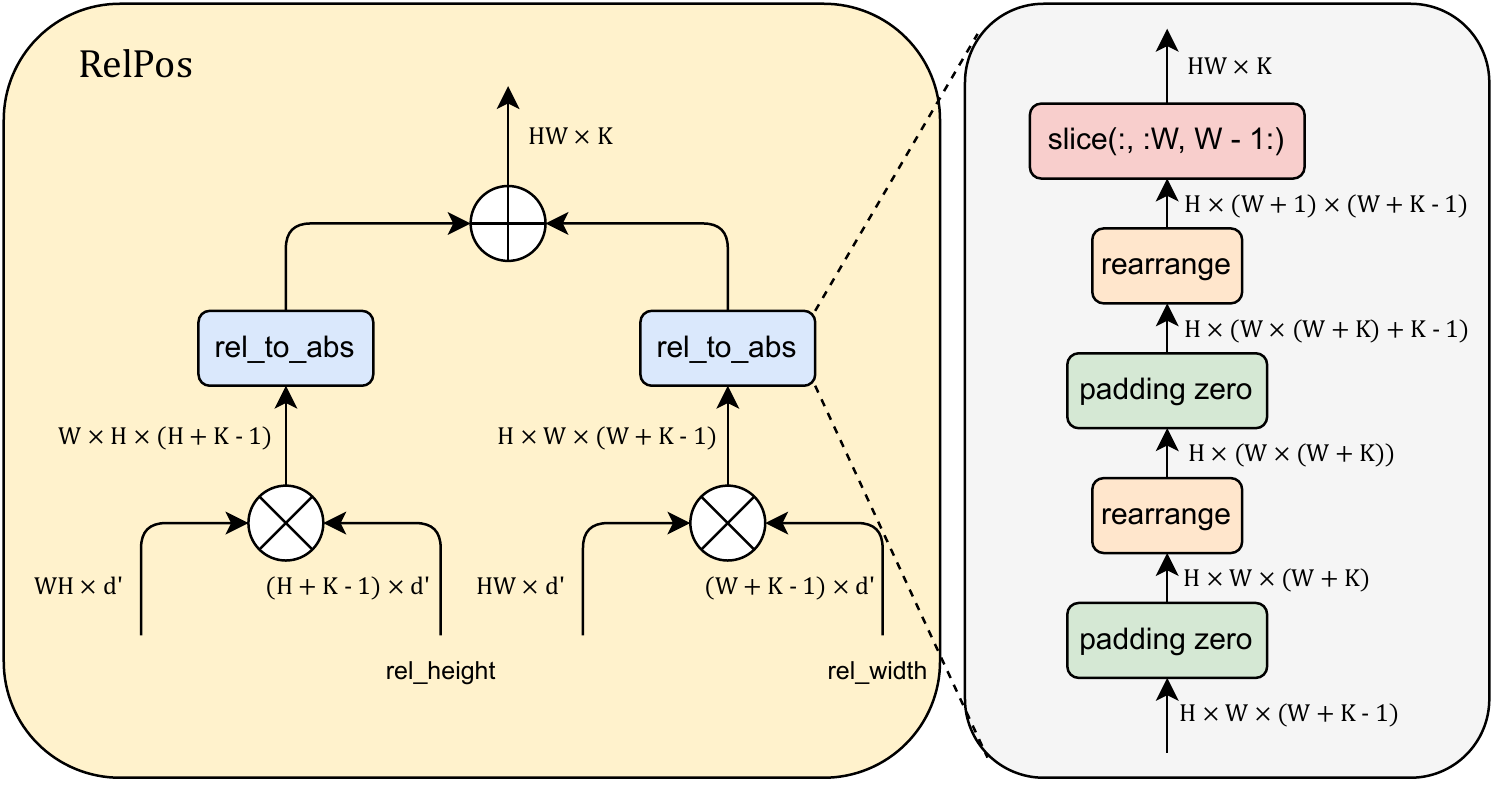}
	\label{fig:dgmn2_framework_relpos}
	} \\
	\subfloat[Schematic illustration of DGMN2 backbone network.]{
	\includegraphics[width=0.95\linewidth]{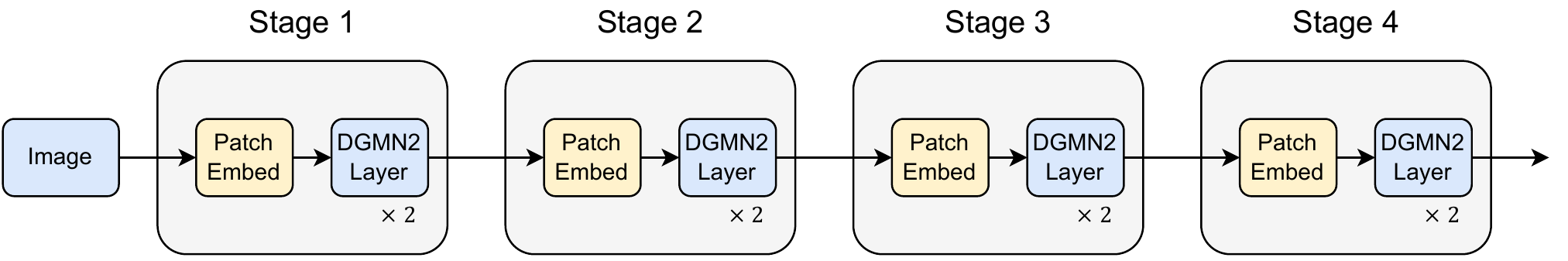}
	\label{fig:dgmn2_framework_arch}
	}
	\caption{
	\small 
	The architecture of DGMN2 (DGMN2-Tiny). 
	``$\rm d'$'' denotes the embedding dimension of tokens in each attention head. 
	``RelPos'' denotes the relative position encoding.
	$\bigoplus$ and $\bigotimes$ denotes element-wise sum and matrix multiplication respectively.
	}
	\label{fig:dgmn2_framework}
\end{figure*}

\section{Extending DGMN to Transformers}

\subsection{Interpretation as a Transformer layer}
We further instantiate our formulation into an efficient and effective Transformer layer and present a new Transformer backbone network DGMN2 to 
support various vision downstream tasks.

Specifically, we modify the process of dynamic message passing calculation and implement it as follows: 
With the predicted walk $\bigtriangleup \vect{d}_j^q$, we can obtain a new set of adaptively sampled nodes with graph vertexes $\mathcal{V}$.
Given the latent node $\vect{h}_{j}^{(t)}$, 
we can estimate the dynamic weight $\vect{w}_j^q$ 
and dynamic affinity $\vect{A'}_{i,j}^q$. 
This process can be defined as:
\begin{align}
\{\vect{w}_j^q,~\vect{A'}_{i,j}^q\} = \varrho \left(\vect{W}_{i,j}^{k,A} \vect{h}_{j}^{(t)} + \vect{b}_{i,j}^{k,A}~| ~\mathcal{V},~j,~\bigtriangleup \vect{d}_j^q \right),
\label{equ:dgmn2_sample}
\end{align}
where $\vect{W}_{i,j}^{k,A}$ are two sets of matrix transformations to transform the latent node $\vect{h}_{j}^{(t)}$.
Both of them are implemented by $1 \times 1$ convolution layers.
The function $\varrho(\cdot)$ is a bilinear sampler~\cite{jaderberg2015spatial}. 

With the estimated dynamic weight $\vect{w}_j^q$ 
and dynamic affinity $\vect{A'}_{i,j}^q$ on the sampled graph nodes, we can calculate the messages as:

\begin{equation}
\begin{aligned}
\vect{m}^{(t+1)}_{i} =& \sum_{q} \sum_{j \in \mathcal{N}_{q}(i)} \\ & \mathrm{softmax} \left(\vect{w}_j^q \left(\vect{W}_{i,j}^h \vect{h}_{j}^{(t)} + \vect{b}_{i,j}^h \right) + \vect{r}_j\right) {A'}_{i,j}^{q},
\label{equ:dgmn2_message}
\end{aligned}
\end{equation}
where the $\mathrm{softmax}(\cdot)$ denotes a softmax operation along the axis of number of sampled nodes.
$\vect{r}_j$ is a relative position encoding (Fig.~\ref{fig:dgmn2_framework_relpos}) and is added to each sampled node. 
$\vect{W}_{i,j}^{h}$ and $\vect{b}_{i,j}^h$ are matrix transformation to transform the latent feature $\vect{h}_{j}^{(t)}$.
It is implemented by a $1 \times 1$ convolution layer.
We set $q=1$ by default in our modified message passing process.
We then scale the message and add it with the observed feature map $\vect{F}$. 
We can perform dynamic message passing for $T$ steps and produce a refined feature map $\vect{H}^{(T)}$ as output.

Fig.~\ref{fig:dgmn2_framework_block} shows how our modified dynamic message passing networks (\textbf{DGMN2 block}) can be implemented in a neural network. 
We replace the self-attention block with our DGMN2 block in the traditional Transformer, that is, we stack a DGMN2 block with a feed forward residual block~\cite{dosovitskiy2020image} to form a dynamic message passing networks Transformer layer (\textbf{DGMN2 layer}).
Base on above formulation, we rephrase the dynamic message passing process from a Transformer perspective~\cite{vaswani2017attention,dosovitskiy2020image}. 
The node sampling process corresponds to the ``Sampler'' in Fig.~\ref{fig:dgmn2_framework_block}. 
``$K$'' denotes the number of sampled nodes and we set $K=9$ by default.
Specifically, we treat $(\vect{W}_{i,j}^h \vect{h}_{j}^{(t)} + b_{i,j}^h )$, $\vect{w}_j^q$ and ${A'}_{i,j}^{q}$ as the \textit{query}, \textit{key} and \textit{value} in the Transformer respectively.

\par\noindent\textbf{Relative position encoding.}
In order to make our DGMN2 \textit{position aware}, we add position encoding in each DGMN2 block.
Specifically, we consider the \textit{relative position}~\cite{srinivas2021bottleneck,shaw2018self,bello2019attention} between the K sampled nodes. 
We add relative height position and relative width position to obtain 2D relative position information. 
For each direction, we add position information relative to the sampled nodes. 
The specific implementation is shown in Fig.~\ref{fig:dgmn2_framework_relpos}. 
Typically, our DGMN2 is pretrained on ImageNet with resolution of 224 $\times$ 224.
For the downstream tasks with different input resolution, such as object detection and semantic segmentation, we perform bilinear interpolation for the position encoding to adapt to the input of various resolutions.

\par\noindent\textbf{Complexity.}
We summary the complexity of DGMN2 in space and time. For time complexity, it involves: 
(1) The linear layer to predict the random walks: $O(HWKd)$; 
(2) Sampling nodes from \textit{key} and \textit{value}: $O(HWK)$; (3) The multiplication of \textit{query} and \textit{key}: $O(HWKd)$; 
(4) The multiplication of attention matrix and \textit{value}: $O(HWKd)$. 
The total time complexity is $O(3HWKd + HWK)$. 
For space complexity, it involves: 
(1) The predicted random walks: $O(HWK)$; 
(2) Sampling nodes of \textit{key} and \textit{value}: $O(HWKd)$; (3) The attention matrix: $O(HWK)$; 
(4) The product of attention matrix and \textit{value}: $O(HWd)$. The total space complexity is $O(2HWK + HWKd + HWd)$.
As we keep $K=9$ ($K \ll  HW$) a fixed constant in our model, both time and space complexity are $O(HW)$, making DGMN2 a linear Transformer.

\subsection{Transformer network architecture}

\begin{table*}[ht]
\centering
\caption{\small 
Architecture specifications of DGMN2 variants. C33-BN-ReLU denotes three $3 \times 3$ Conv-BN-ReLU, with the stride of 2, 1, 2 respectively. C31-BN-ReLU denotes one $3 \times 3$ Conv-BN-ReLU, with a stride of 2. 
$d_i$ denotes the embedding dimension of tokens in stage $i$;
$h_i$ denotes the head number of the stage $i$. 
$e_i$ denotes the expansion ratio of the feed-forward layer in the stage $i$.
}
\label{tab:dgmn2_arch}
\scalebox{1.15}{
\begin{tabular}{x{48}|x{64}|x{64}|x{64}|x{64}|x{64}}
	  Stage & Layer Name & DGMN2-Tiny & DGMN2-Small & DGMN2-Medium  & DGMN2-Large \\
	\hline
	\multirow{2}{*}[-2.5ex]{Stage 1} & 
	Patch Embedding & \multicolumn{4}{c}{$\text{C33-BN-ReLU}, d_i=64$} \\
	\cline{2-6}
	& DGMN2 Layer  & 
	$\begin{bmatrix}
	\begin{array}{l}
	h_1=1 \\
	e_1=8 \\
	\end{array}
	\end{bmatrix} \times 2$ &
	$\begin{bmatrix}
	\begin{array}{l}
	h_1=1 \\
	e_1=8 \\
	\end{array}
	\end{bmatrix} \times 3$ &
	$\begin{bmatrix}
	\begin{array}{l}
	h_1=1 \\
	e_1=8 \\
	\end{array}
	\end{bmatrix} \times 3$ &
	$\begin{bmatrix}
	\begin{array}{l}
	h_1=1 \\
	e_1=8 \\
	\end{array}
	\end{bmatrix} \times 3$ \\
	\hline
	\multirow{2}{*}[-2.5ex]{Stage 2} & 
	Patch Embedding & \multicolumn{4}{c}{$\text{C31-BN-ReLU}, d_i=128$} \\
	\cline{2-6}
	&  DGMN2 Layer &
	$\begin{bmatrix}
	\begin{array}{l}
	h_2=2 \\
	e_2=8 \\
	\end{array}
	\end{bmatrix} \times 2$ &
	$\begin{bmatrix}
	\begin{array}{l}
	h_2=2 \\
	e_2=8 \\
	\end{array}
	\end{bmatrix} \times 4$ &
	$\begin{bmatrix}
	\begin{array}{l}
	h_2=2 \\
	e_2=8 \\
	\end{array}
	\end{bmatrix} \times 4$ &
	$\begin{bmatrix}
	\begin{array}{l}
	h_2=2 \\
	e_2=8 \\
	\end{array}
	\end{bmatrix} \times 8$\\
	\hline
	\multirow{2}{*}[-2.5ex]{Stage 3} & 
	Patch Embedding & \multicolumn{4}{c}{$\text{C31-BN-ReLU}, d_i=320$} \\
	\cline{2-6}
	& DGMN2 Layer &
	$\begin{bmatrix}
	\begin{array}{l}
	h_3=5 \\
	e_3=4 \\
	\end{array}
	\end{bmatrix} \times 2$ &
	$\begin{bmatrix}
	\begin{array}{l}
	h_3=5 \\
	e_3=4 \\
	\end{array}
	\end{bmatrix} \times 6$ &
	$\begin{bmatrix}
	\begin{array}{l}
	h_3=5 \\
	e_3=4 \\
	\end{array}
	\end{bmatrix} \times 18$ &
	$\begin{bmatrix}
	\begin{array}{l}
	h_3=5 \\
	e_3=4 \\
	\end{array}
	\end{bmatrix} \times 27$\\
	\hline
	\multirow{2}{*}[-2.5ex]{Stage 4} & 
	Patch Embedding & \multicolumn{4}{c}{$\text{C31-BN-ReLU}, d_i=512$} \\
	\cline{2-6}
	& DGMN2 Layer  & 
	$\begin{bmatrix}
	\begin{array}{l}
	h_4=8 \\
	e_4=4 \\
	\end{array}
	\end{bmatrix} \times 2$ &
	$\begin{bmatrix}
	\begin{array}{l}
	h_4=8 \\
	e_4=4 \\
	\end{array}
	\end{bmatrix} \times 3$ &
	$\begin{bmatrix}
	\begin{array}{l}
	h_4=8 \\
	e_4=4 \\
	\end{array}
	\end{bmatrix} \times 3$ &
	$\begin{bmatrix}
	\begin{array}{l}
	h_4=8 \\
	e_4=4 \\
	\end{array}
	\end{bmatrix} \times 3$\\
\end{tabular}
}

\end{table*}

By performing dynamic message passing for $T$ steps, we can further propose an efficient Transformer backbone network with $T$ DGMN2 layers.
Focusing on the general image recognition task and being able to support downstream vision tasks, we integrate our DGMN2 layer into the recent pyramidal Transformer architecture~\cite{wang2021pyramid,lu2021soft} to form our final model \textbf{DGMN2}.

We take DGMN2-Tiny as an example, Fig.~\ref{fig:dgmn2_framework_arch} illustrates the architecture of the DGMN2 backbone network.
DGMN2 has four stages to extract features at different scales.
For example, we use two DGMN2 layers after a patch embedding (\eg,~tokenization) layer in stage 1 of DGMN2-Tiny.
The patch embedding layers are used to downsample the token length to output stride of $\frac{1}{4}$, $\frac{1}{8}$, $\frac{1}{16}$, $\frac{1}{32}$ respectively in each of the four stages.
Specifically, 
the \texttt{STEM} is implemented by three units of \texttt{3$\times$3 Conv-BN-ReLU}, with the stride of 2, 1, 2 respectively; and
one such unit is applied to each of three following down-sampling operations with stride of 2 in the multi-stage architecture.
Table~\ref{tab:dgmn2_arch} details the family of our DGMN2 configurations with varying capacities: -Tiny, -Small, -Medium and -Large.
The architecture hyper-parameters of DGMN2 are listed as follows:
\begin{itemize}
    \item $d_i$: the embedding dimension of tokens in stage $i$;
    \item $h_i$: the head number in stage $i$;
    \item $e_i$: the expansion ratio of the feed-forward layer in stage $i$;
\end{itemize}

\subsection{Discussion}

Our DGMN2 is related to DAT~\cite{xia2022vision} and Deformable DETR~\cite{zhu2020deformable}, but has key differences:

When calculating attention, DAT~\cite{xia2022vision} learns a offset for each point on feature map and use the deformed points as the key and value. In contrast, our DGMN2 samples $K$ nodes of key and value and calculate the attention using sampled key and value. Thus our method need fewer FLOPs and is more efficient.

Deformable DETR~\cite{zhu2020deformable} propose an object detector that uses deformable attention. Deformable DETR samples a small number of points around a reference point, while our DGMN2 sampling nodes on the whole feature map without the assistance of reference points. Moreover, the attention weights of Deformable DETR are obatained via linear projection over the query feature. In contrast, our DGMN2 follows the form of dot-product attention and obtains attention weights via the multiplication of queries and keys.

\section{Experiments with convolutional architectures}
\label{sec:exp}
\begin{figure*}[!t]
	\def \imheight {2.50cm}
	\centering
	\includegraphics[height=\imheight]{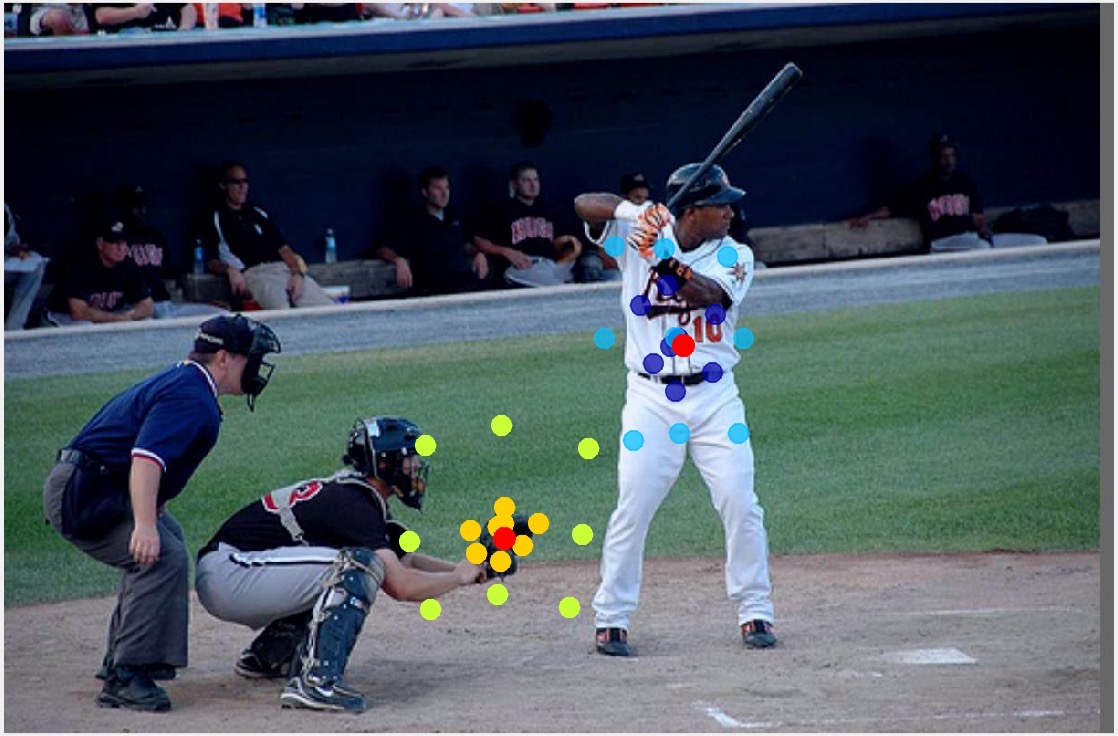}
	\includegraphics[height=\imheight]{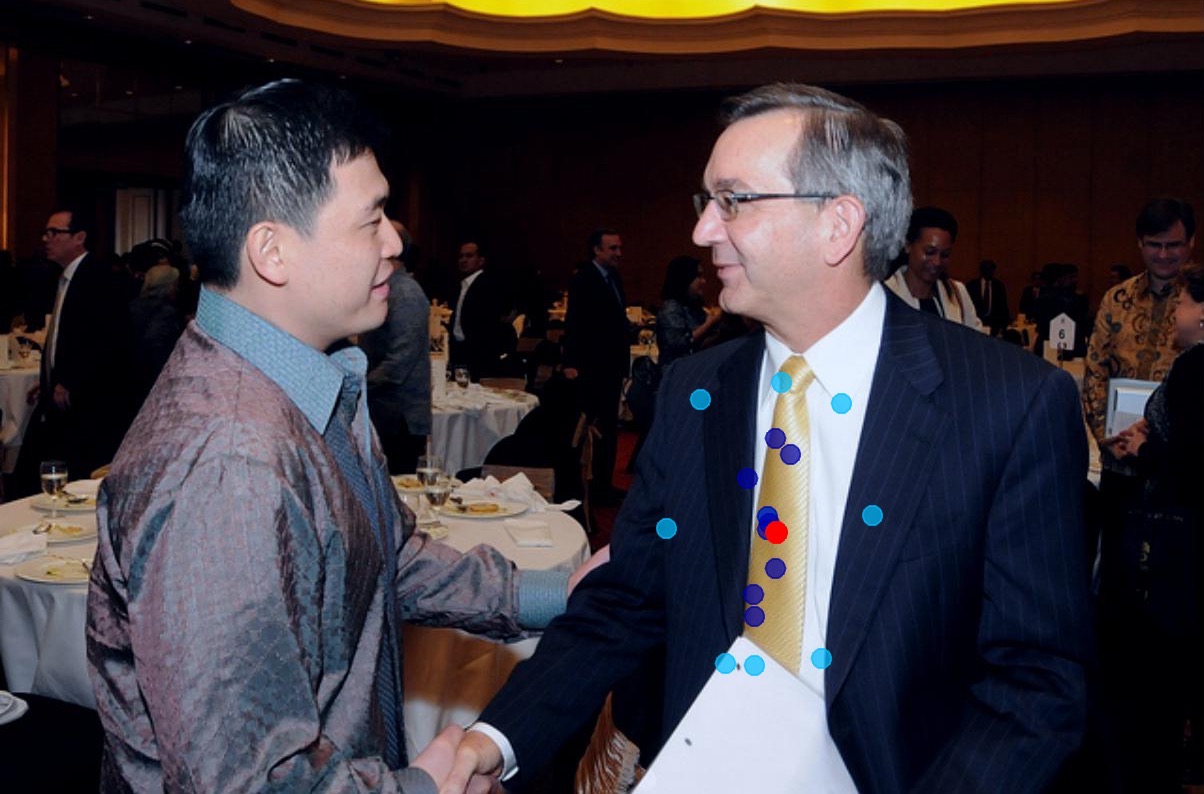}
	\includegraphics[height=\imheight]{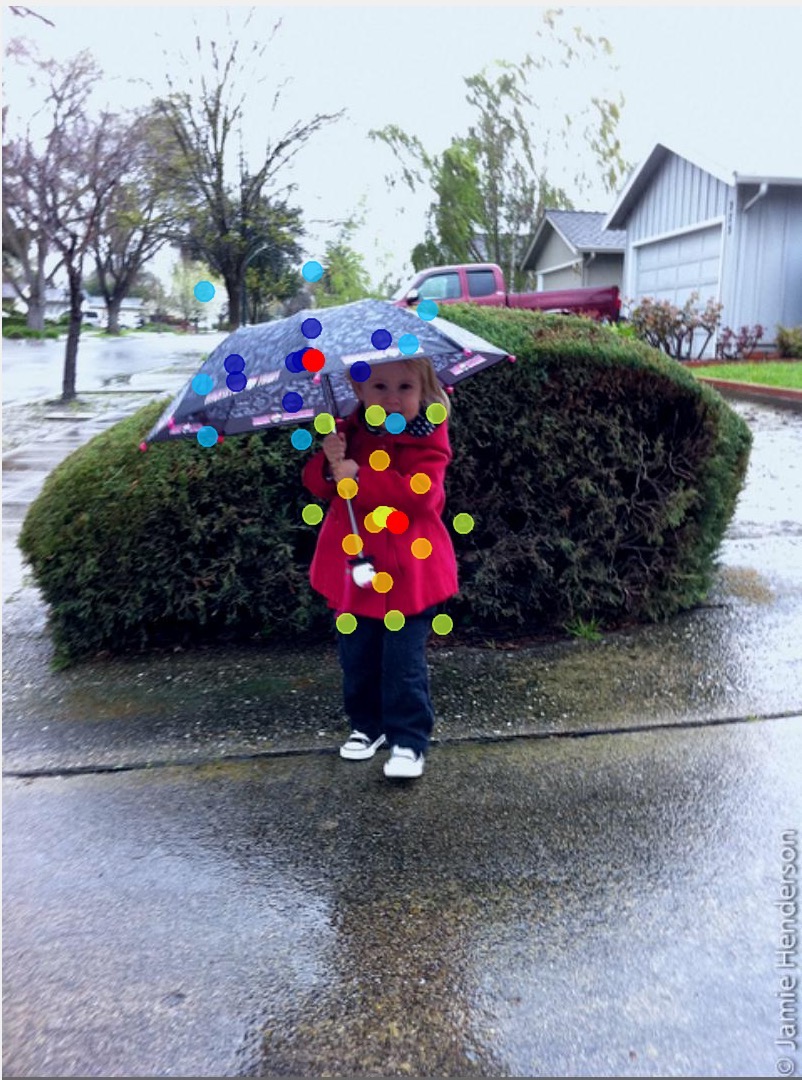} 
	\includegraphics[height=\imheight]{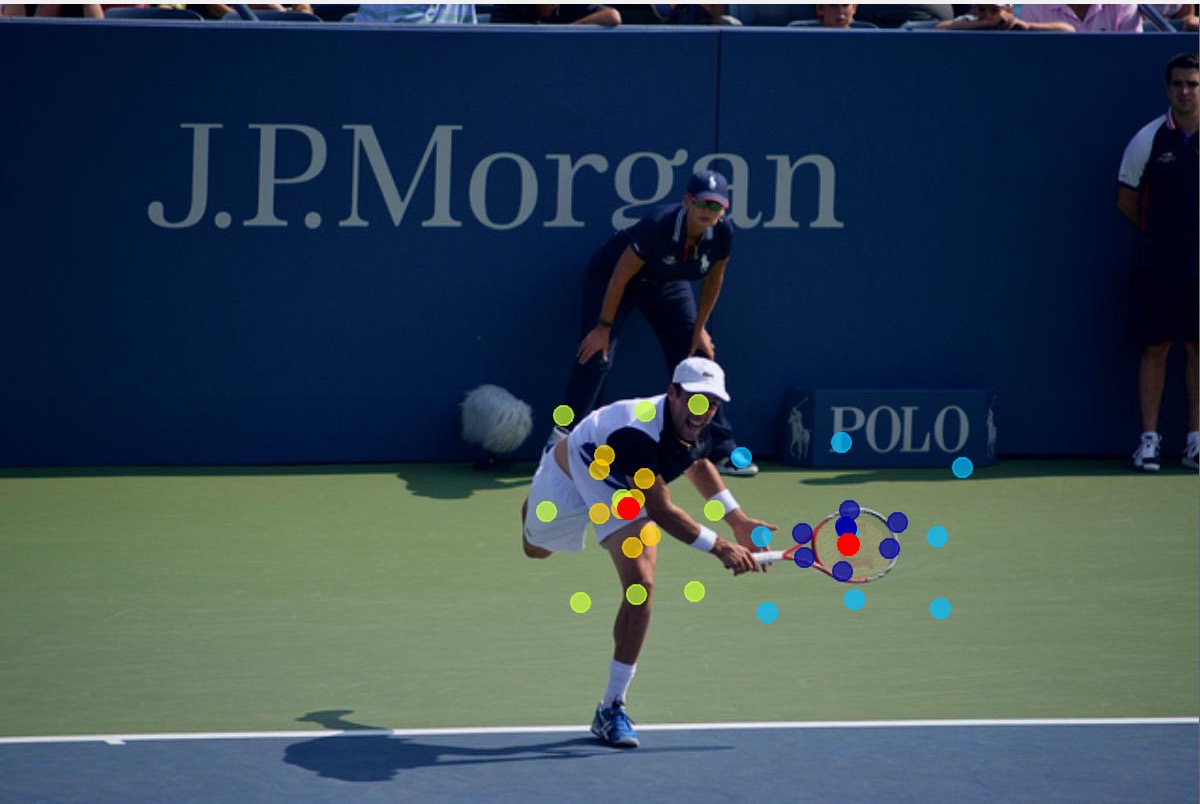}
	\includegraphics[height=\imheight]{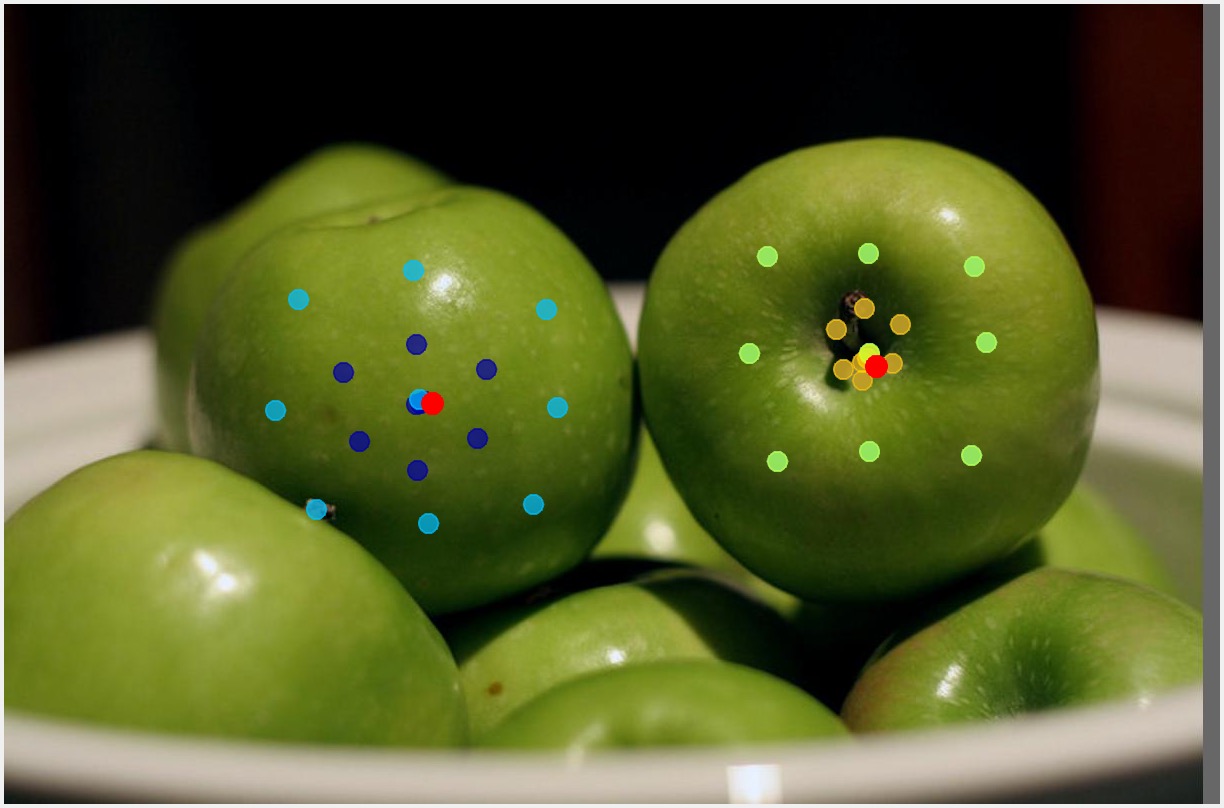}
	\caption{\small Visualisation of the nodes sampled via \textit{learning} the random walks with our network (trained for instance segmentation on COCO).  
	The red point indicates a receiving node $i$. 
	Different colour families (\ie~yellow and blue) indicate the learned position specific weights and affinities of the sampled nodes.
	Different colours in the same family show the sampled nodes with different sampling rates for the same receiving node. 
	}
	\label{fig:offset}
\end{figure*}
\subsection{Experimental setup}
\label{sec:exp_setup}

\par\noindent\textbf{Tasks and datasets.}
We evaluate our proposed model on three challenging public benchmarks, \ie,
~{Cityscapes}~\cite{Cityscapes} for semantic segmentation, and {COCO}~\cite{lin2014microsoft} for object detection and instance segmentation.
\par\noindent\textbf{Cityscapes:} Cityscapes~\cite{Cityscapes} has densely annotated semantic labels for 19 categories in urban road scenes, and contains a total of 5000 finely annotated images, divided into 2975, 500, and 1525 images for training, validation and testing respectively. 
In addition, it provides 19,998 coarse annotated images for model training.
The images of this dataset have a high resolution of $1024 \times 2048$. 
Following the standard evaluation protocol~\cite{Cityscapes}, the metric of mean Intersection over Union (mIoU) averaged over all classes is reported. 
\par\noindent\textbf{COCO:} COCO 2017~\cite{lin2014coco} consists of 80 object classes with a training set of 118,000 images, a validation set of 5000 images, and a test set of 2000 images. 
We follow the standard COCO evaluation metrics \cite{lin2014microsoft} to evaluate the performance of object detection and instance segmentation, employing the metric of mean average-precision (mAP) at different box and mask IoUs respectively.

We follow the standard protocol and evaluation metrics used by these public benchmarks. 

\par\noindent\textbf{Baseline models.}
For semantic segmentation on Cityscapes, our baseline is Dilated-FCN~\cite{yu2015multi} with a ResNet-101 backbone pretrained on ImageNet.
A randomly initialised $3 \times 3$ convolution layer, together with batch normalisation and ReLU is used after the backbone to produce a dimension-reduced feature map of 512 channels which is then fed into the final classifier.
For the task of object detection and instance segmentation on COCO, our baseline is Mask-RCNN~\cite{maskrcnn,massa2018mrcnn} with FPN and ResNet/ResNeXt~\cite{resnet,xie2017aggregated} as a backbone architecture.
Unless otherwise specified, we use a {single scale} and test with a {single model} for all experiments without using other complementary performance boosting ``tricks''. 
We train models on the COCO training set and test on the validation and test-dev sets.

Across all tasks and datasets, we consider Non-local networks~\cite{wang2018nonlocal} as an additional baseline. 
To have a direct comparison with the Deformable Convolution method~\cite{dai2017deformable,zhu2018deformable}, we consider two baselines: 
(i) ``deformable message passing'', which is a variant of our model using randomly initialised deformable convolutions for message calculation, but without using the proposed dynamic sampling and dynamic weights/affinities strategies, and 
(ii) the original deformable method which replaces the convolutional operations as in~\cite{dai2017deformable,zhu2018deformable}.

\par\noindent\textbf{Implementation details.} 
For our experiments on Cityscapes, our DGMN module is randomly initialised and inserted between the $3 \times 3$ convolution layer and the final classifier.
For the experiments on COCO, we insert one or multiple randomly initialised DGMN modules into the backbone for deploying our approach for feature learning.
Our models and baselines for all of our COCO experiments are trained with the typical ``$1\mathrm{x}$'' training settings from the public Mask R-CNN benchmark~\cite{massa2018mrcnn}.

When predicting dynamic filter weights, we used the grouping parameter $G=4$. 
For our experiments on Cityscapes, the sample rates are set to $\varphi = \{1, 6, 12, 24, 36\}$.
For experiments on COCO, we use smaller sampling rates of $\varphi = \{1, 4, 8, 12\}$.	
The effect of this hyperparameter 
and additional implementation details are described in the supplementary material.

\subsection{Model analysis}
\label{sec:exp_ablation}

To demonstrate the effectiveness of the proposed components of our model, we conduct ablation studies of: 
(i) DGMN w/ DA, which adds the dynamic affinity (DA) strategy onto the DGMN base model; 
(ii) DGMN w/ DA+DW, which further adds the proposed dynamic weights (DW) prediction; 
(iii) DGMN w/ DA+DW+DS, which is our full model with the dynamic sampling (DS) scheme added upon DGMN w/ DA+DW.
Note that DGMN Base was described in Sec.~\ref{sec:method_mpgraph}, DS in Sec.~\ref{sec:method_sampling}, and DA and DW in \ref{sec:method_dynamic_weights}.

\begin{table}[t]
	\centering
	\caption{Ablation study on the Cityscapes validation set for semantic segmentation. All models have a ResNet-101 backbone and are evaluated at a single scale.}
	\label{tab:ablation_study_cityscapes}
	{\tablestyle{2pt}{1.2}
		\begin{tabular}{ l|x{34}x{34}x{34} }
		&  mIoU (\%)  &   Params &   FLOPs \\
			\shline
			Dilated FCN~\cite{yu2015multi}  & 75.0 &  -- &  --\\
            \hline
             + Deformable~\cite{zhu2018deformable} & 78.2 & +1.31M  &+12.34G \\
             + ASPP~\cite{deeplabv3} &78.9&  +4.42M & +38.45G \\
             + Non-local~\cite{wang2018nonlocal}     &  79.0 & +2.88M & +73.33G\\
            \hline
             + DGMN w/ DA   & 76.5   & +0.57M & +5.32G\\
             + DGMN w/ DA+DW   & 79.1   & +0.73M & +6.88G\\
             + DGMN w/ DA+DW+DS   & \textbf{80.4}   & +2.61M & +24.55G\\
		\end{tabular}
	}
\end{table}

\par\noindent\textbf{Effectiveness of the dynamic sampling strategy.} 
Table~\ref{tab:ablation_study_cityscapes} shows quantitative results of semantic segmentation on the Cityscapes validation set. 
DGMN w/ DA+DW+DS clearly outperforms DGMN w/ DA+DW on the challenging segmentation task, meaning that the feature-conditioned adaptive sampling based on learned random walks is more effective compared to a spatial uniform sampling strategy when selecting nodes. 
More importantly, all variants of our module for both semantic segmentation and object detection that use dynamic sampling (Tab.~\ref{tab:ablation_study_cityscapes} and Tab.~\ref{tab:ablation_study_coco}) achieve higher performance than a fully-connected model (\ie,~Non-local~\cite{wang2018nonlocal}) with substantially fewer FLOPs. This emphasises the performance benefits of our dynamic graph sampling model.
Visualisations of the nodes dynamically sampled by our model are shown in Fig.~\ref{fig:offset}.

\begin{table}
	\centering
	\caption{ 
    \small Quantitative results of different models on the COCO 2017 validation set for object detection ($\mathrm{AP^{b}}$) and instance segmentation ($\mathrm{AP^{m}}$). 
    C5 denotes inserting DGMN after all $3 \times 3$ convolutional layers in \emph{res5}.
    All methods are based on the Mask R-CNN with ResNet-50 as backbone. 
    }
    \label{tab:ablation_study_coco}
	\tablestyle{1.2pt}{1}
    \begin{tabular}{l|x{21}x{21}x{21}|x{21}x{21}x{21}}
     &\scriptsize $\mathrm{AP^{b}}$ &\scriptsize $\mathrm{AP^{b}_{50}}$ &\scriptsize $\mathrm{AP^{b}_{75}}$ &\scriptsize $\mathrm{AP^{m}}$ &\scriptsize $\mathrm{AP^{m}_{50}}$ &\scriptsize $\mathrm{AP^{m}_{75}}$  \\
    \shline
    \scriptsize Mask R-CNN baseline   &\scriptsize 37.8&\scriptsize 59.1 &\scriptsize 41.4 &\scriptsize 34.4  &\scriptsize 55.8 &\scriptsize 36.6 \\
    \hline
    \scriptsize + GCNet~\cite{cao2019gcnet} &\scriptsize 38.1 &\scriptsize  60.0 &\scriptsize41.2 &\scriptsize 34.9 &\scriptsize 56.5 &\scriptsize37.2  \\
    \scriptsize + Deformable Message Passing &\scriptsize 38.7 & \scriptsize 60.4 &\scriptsize42.4 &\scriptsize 35.0 &\scriptsize 56.9 &\scriptsize37.4  \\
    \scriptsize + Non-local~\cite{wang2018nonlocal} 
    &\scriptsize 39.0 &\scriptsize 61.1 &\scriptsize 41.9 &\scriptsize 35.5  &\scriptsize 58.0 &\scriptsize 37.4 \\
    \scriptsize + CCNet~\cite{huang2018ccnet}    &\scriptsize39.3 &\scriptsize- &\scriptsize- &\scriptsize36.1  &\scriptsize- &\scriptsize- \\
    \scriptsize+ DGMN &\scriptsize \bf 39.5 & \scriptsize \bf 61.0 &\scriptsize \bf 43.3&\scriptsize \bf 35.7&\scriptsize \bf 58.0 &\scriptsize \bf 37.9 \\ 
    \hline
    \scriptsize + GCNet (C5) ~\cite{cao2019gcnet}    &\scriptsize38.7 &\scriptsize61.1 &\scriptsize41.7 &\scriptsize35.2 &\scriptsize57.4 &\scriptsize37.4 \\ 
    \scriptsize + Deformable (C5) ~\cite{zhu2018deformable}     &\scriptsize39.9&\scriptsize- &\scriptsize- &\scriptsize34.9  &\scriptsize- &\scriptsize- \\ 
    \scriptsize+ DGMN (C5)  &\scriptsize\textbf{40.2}&\scriptsize\textbf{62.0} &\scriptsize\textbf{43.4} &\scriptsize\textbf{36.0}  &\scriptsize\textbf{58.3} &\scriptsize\textbf{38.2} \\ 
    \multicolumn{7}{c}{~}\\
    \end{tabular}
\end{table}

\begin{figure}[t]
	\centering
	\includegraphics[width=0.86\linewidth]{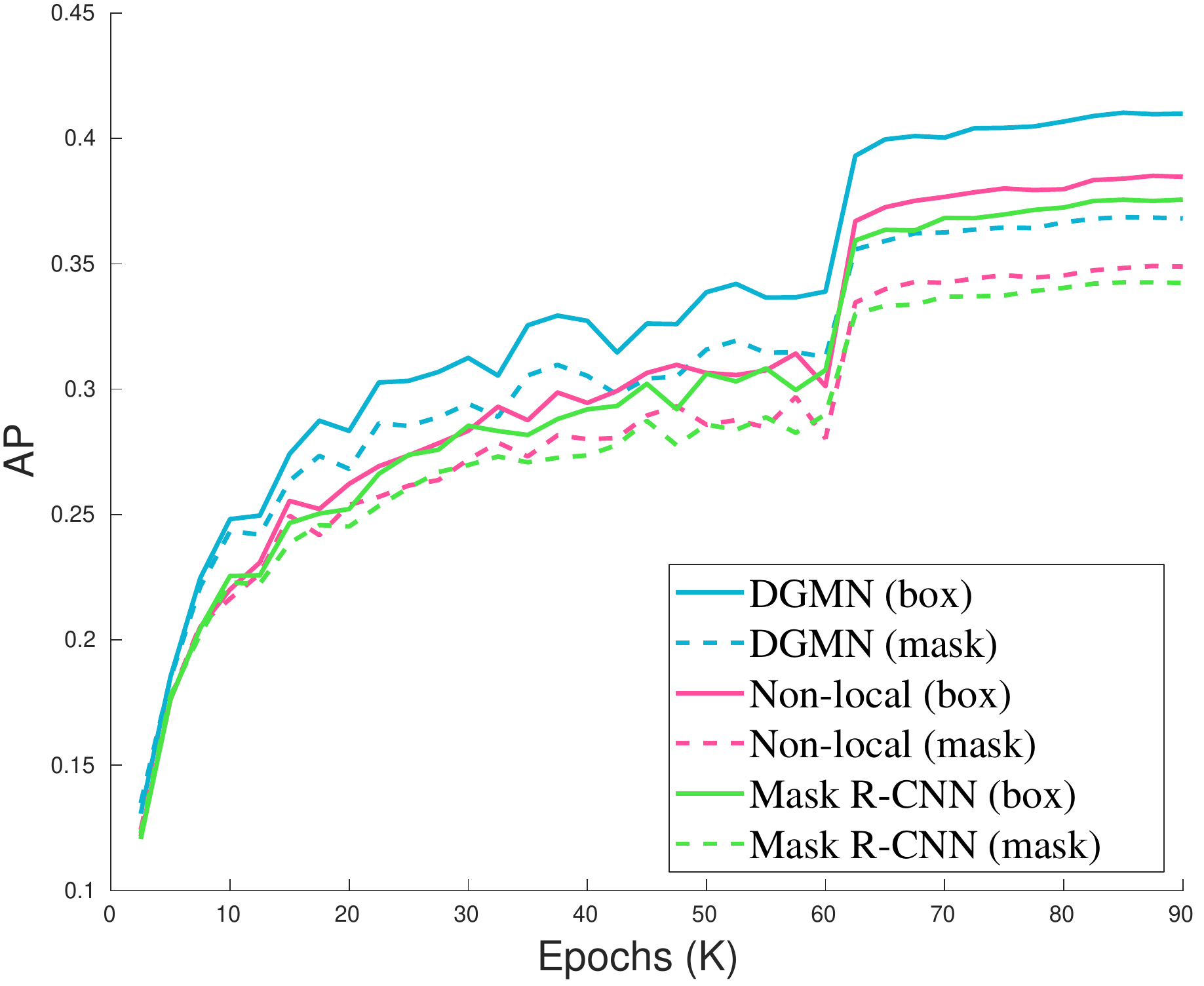}
	\caption{
	\small Validation curves of $\mathrm{AP^{box}}$ and $\mathrm{AP^{mask}}$ on COCO for Mask-RCNN baseline, Non-local and the proposed DGMN.
	The number of training epochs is 90K.
	}
	\label{fig:apcurve}
\end{figure}

\par\noindent\textbf{Effectiveness of joint learning the dynamic filters and affinities.} 
As shown in Tab.~\ref{tab:ablation_study_cityscapes}, DGMN w/ DA is 1.5 points better than Dilated FCN baseline with only a slight increase in FLOPs and parameters, showing the benefit of using predicted dynamic affinities for reweighting the messages in message passing. 
By further employing the estimated dynamic filter weights for message calculation, the performance increases substantially from a mIoU of 76.5\% to 79.1\%, which is almost the same as the 79.2\% of the Non-local model~\cite{wang2018nonlocal}.
Crucially, our approach only uses 9.4\% of the FLOPs and 25.3\% of the parameters compared to Non-local.
These results clearly demonstrate our motivation of jointly learning the dynamic filters and dynamic affinities from sampled graph nodes. 

\begin{figure*}[ht]
	\centering
	\def \imwidth {4cm}
	\small
	\begin{tabularx}{\linewidth}{YYYY}
	\includegraphics[width=\imwidth]{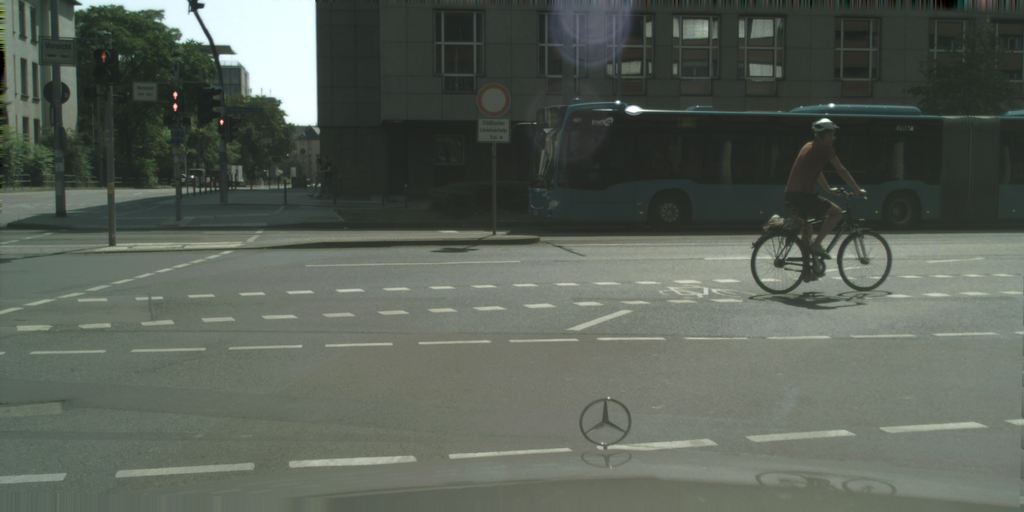} &
	\includegraphics[width=\imwidth]{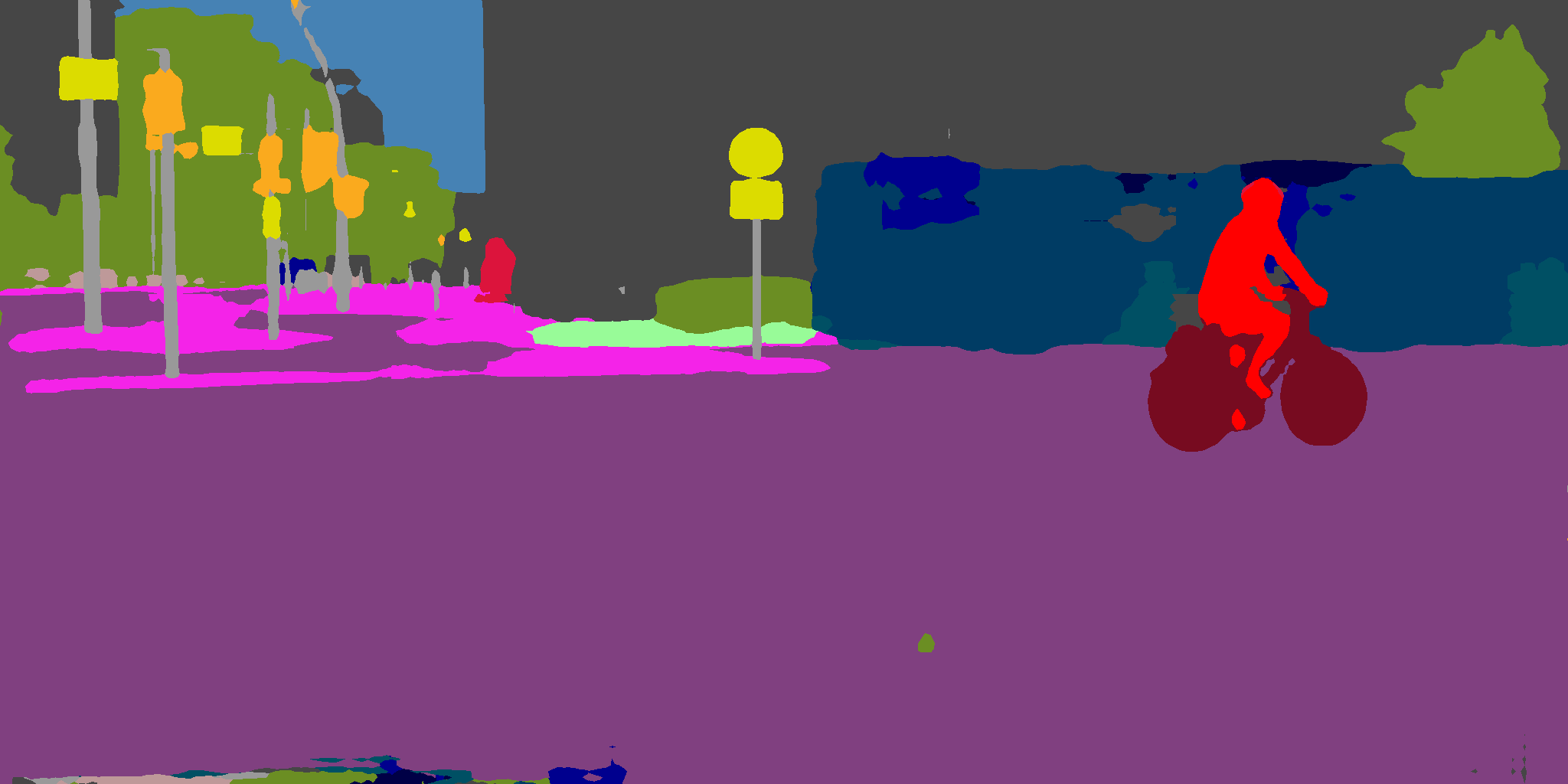} &
	\includegraphics[width=\imwidth]{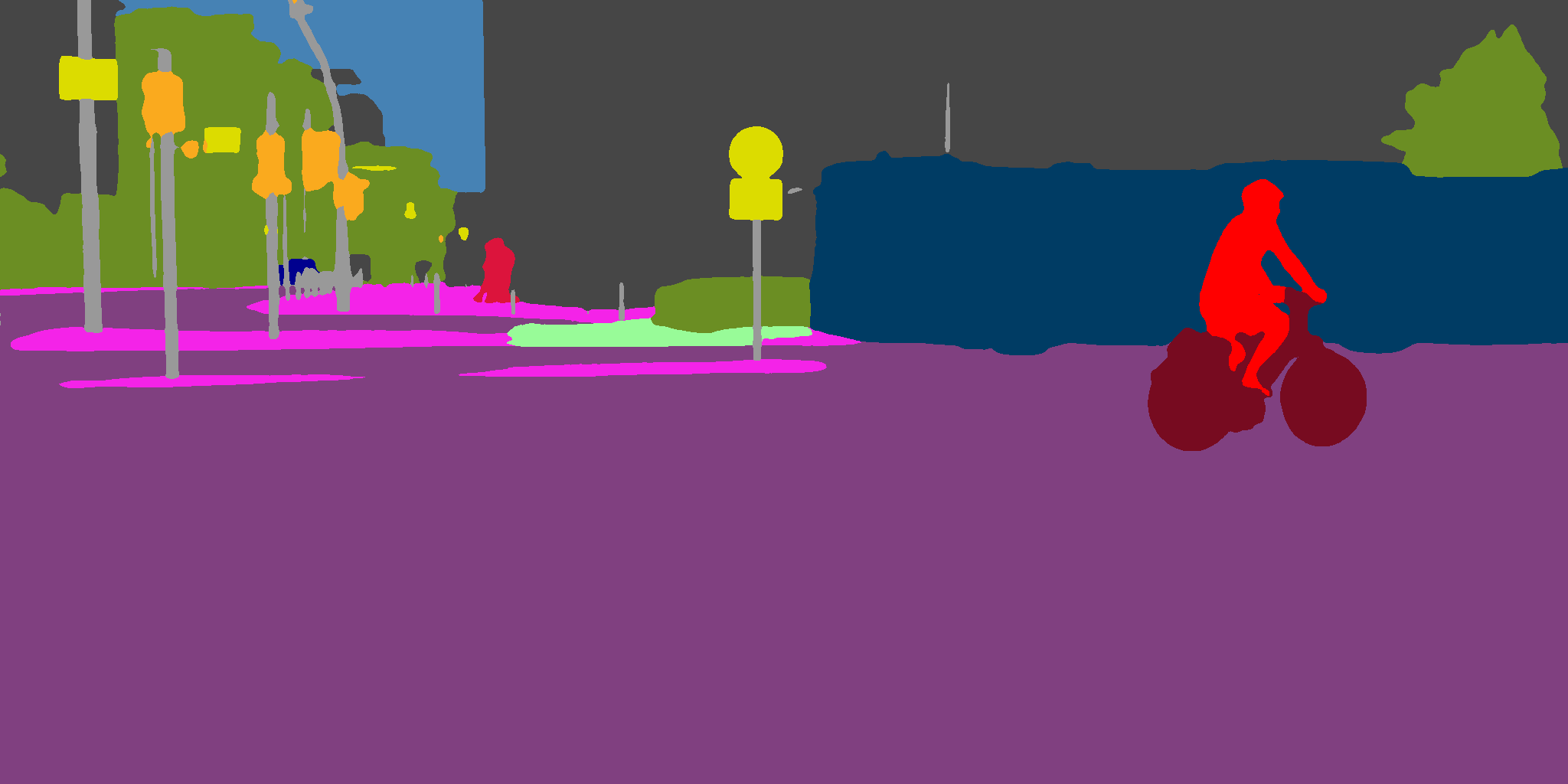} &
	\includegraphics[width=\imwidth]{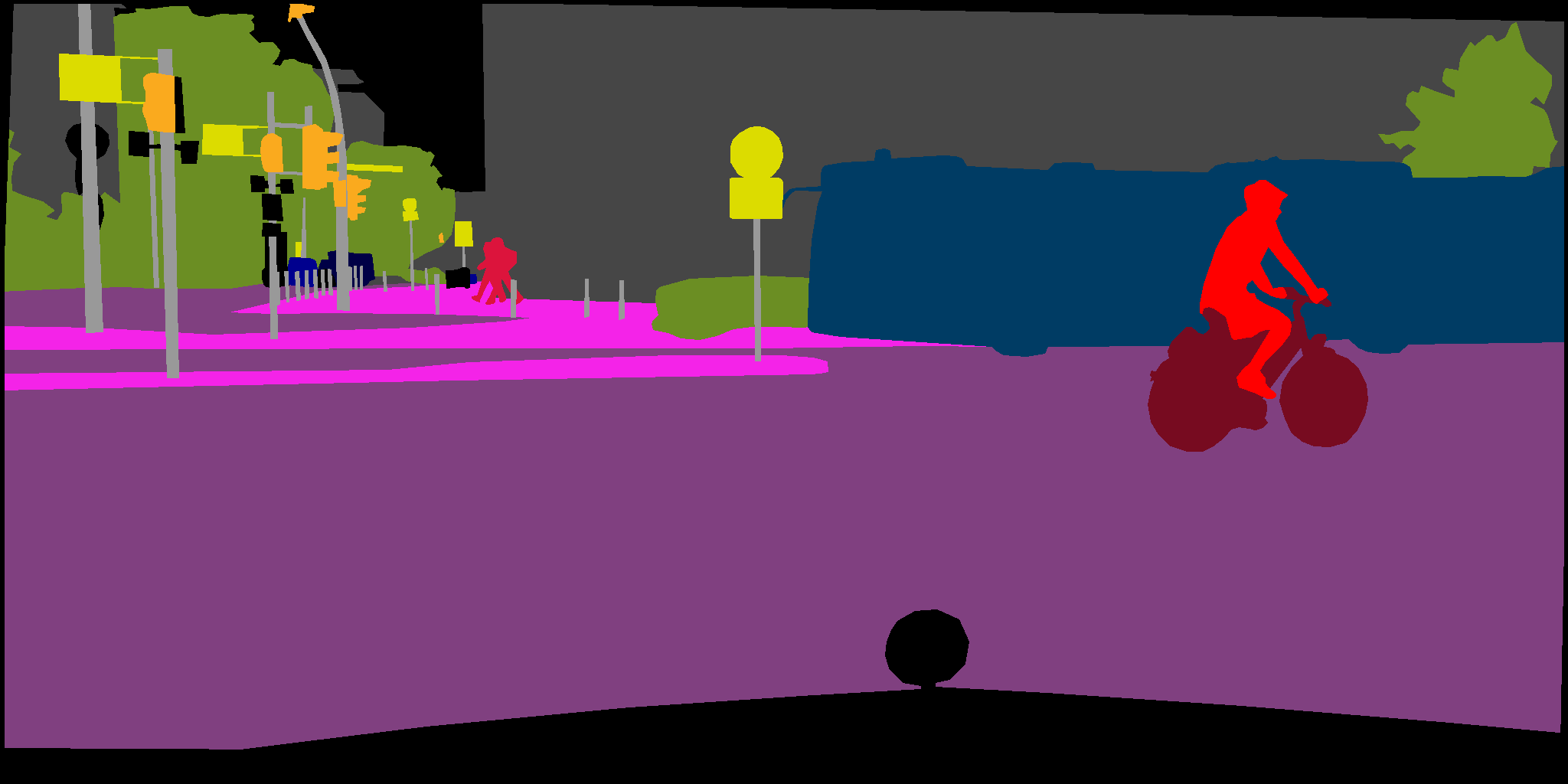}
	\\
	Input image & Dilated FCN & DGMN (ours) & Ground truth \\
	\includegraphics[width=\imwidth]{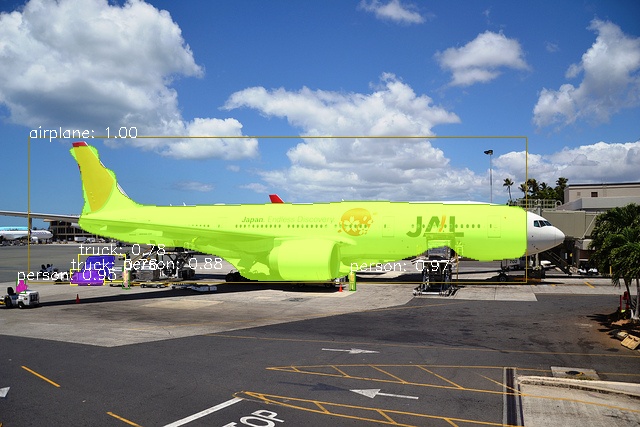} &
	\includegraphics[width=\imwidth]{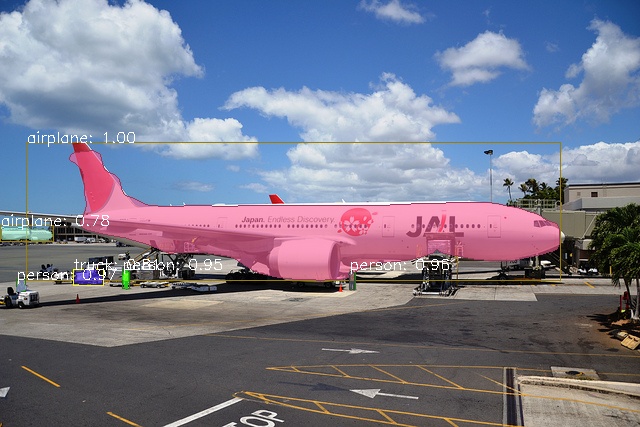} &
	\includegraphics[width=\imwidth]{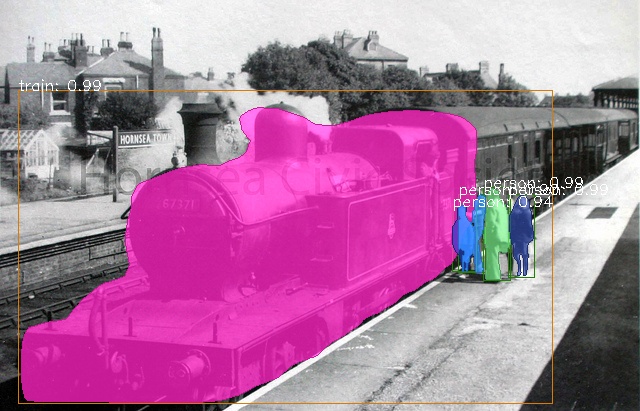} &
	\includegraphics[width=\imwidth]{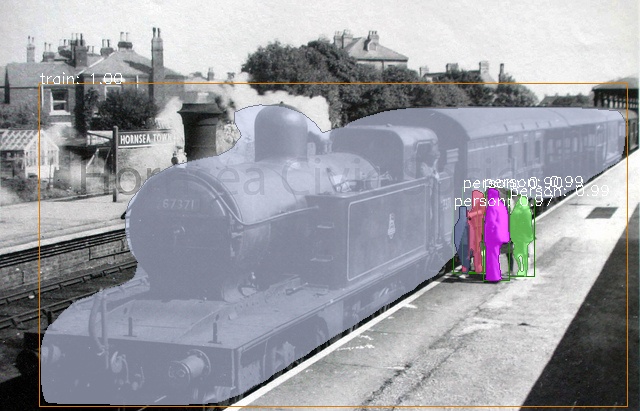} \\
	Mask R-CNN & DGMN (ours) & Mask R-CNN & DGMN (ours) \\
	\end{tabularx}	
	\caption{\small Qualitative examples of our results for  semantic segmentation on Cityscapes (first row), and object detection and instance segmentation on COCO (second row)
	}
	\label{fig:qualitative}
\end{figure*}

\par\noindent\textbf{Comparison with other baselines.} 
For semantic segmentation on Cityscapes, we clearly outperform the ASPP module of Deeplab v3 \cite{deeplabv3} which also increases the receptive field by using multiple dilation rates.
Notably, we improve upon the most related method, Non-local~\cite{wang2018nonlocal} which models a fully-connected graph, whilst using only 33\% of the FLOPs of~\cite{wang2018nonlocal}.
This suggests that a fully-connected graph models redundant information, and further confirms the performance and efficiency of our model.

For fair comparison with Non-local ~\cite{wang2018nonlocal} as well as other alternatives on COCO, we insert one randomly initialised DGMN module right before the last residual block of \textit{res4}~\cite{wang2018nonlocal} (Tab.~\ref{tab:ablation_study_coco}).
We also compare to GCNet~\cite{cao2019gcnet} and CCNet~\cite{huang2018ccnet} which both aim to reduce the complexity of the fully connected Non-local model.
Our proposed DGMN model substantially improves upon these strong baselines and alternatives. 
Fig.~\ref{fig:apcurve} further shows the validation curves of our method and different baselines using the standard $\text{AP}^{\text{Box}}$ and $\text{AP}^{\text{Mask}}$ measures for semantic and instance segmentation respectively.
Our method is consistently better than the Non-local and Mask R-CNN baselines throughout training.

\par\noindent\textbf{Effectiveness of multiple DGMN modules.} 
We further show the effectiveness of our approach for representation learning by inserting multiple of our DGMN modules into the ResNet-50 backbone. 
Specifically, we add our full DGMN module \textit{after} all $3 \times 3$ conv layers in \textit{res5} which we denote as ``C5''.
The second part of Tab.~\ref{tab:ablation_study_coco} shows that our model significantly improves upon the Mask R-CNN baseline 
with the improvements of 2.4 points for the AP$^{\mathrm{box}}$ on object detection, and 1.6 points for the AP$^{\mathrm{mask}}$ on instance segmentation.
Furthermore, when we insert the GCNet module~\cite{cao2019gcnet} in the same locations for comparison, our model achieves better performance too. 
A straightforward comparison to the deformable method of \cite{zhu2018deformable} is the deformable message passing baseline in which we disable the proposed dynamic sampling and the dynamic weights and affinities learning strategies. 
In Tab.~\ref{tab:ablation_study_coco},
our model significantly improves upon the deformable message passing method, which is a direct evidence of the effectiveness of jointly modeling the dynamic sampling and dynamic filters and affinities for feature learning.
Furthermore, we also consider inserting the improved Deformable Conv method~\cite{zhu2018deformable} in C5, which is complementary to our model since it is plugged \textit{before} $3 \times 3$ layers to replace the convolution operations.
Our approach also achieves better performance than it.

\begin{table}
	\centering
	\caption{
	Comparison to state-of-the-art for semantic segmentation on Cityscapes.
	All methods are trained with the finely-annotated data from the training and validation sets.
	}
	\label{tab:city_test}
	\label{tab:city_test}{
		\tablestyle{1pt}{1}
		\begin{tabular}{l |x{42}|x{42}}
			&\scriptsize Backbone &\scriptsize mIoU (\%) \\
			\shline
			\scriptsize PSPNet~\cite{pspnet}   &\scriptsize ResNet 101  &\scriptsize 78.4\\
            \scriptsize PSANet~\cite{psanet} &\scriptsize ResNet 101  &\scriptsize 80.1   \\
            \scriptsize DenseASPP~\cite{denseaspp} &\scriptsize DenseNet 161   &\scriptsize 80.6   \\
            \hline
            \scriptsize GloRe~\cite{graph_reason} &\scriptsize ResNet 101  &\scriptsize 80.9  \\
            \scriptsize Non-local~\cite{wang2018nonlocal} &\scriptsize ResNet 101 &\scriptsize 81.2 \\
            \scriptsize CCNet~\cite{huang2018ccnet} &\scriptsize ResNet 101 &\scriptsize 81.4 \\
            \scriptsize DANet~\cite{DAnet} &\scriptsize ResNet 101     &\scriptsize 81.5   \\
            \scriptsize DGMN (Ours)  &\scriptsize ResNet 101   &\scriptsize \bf 81.6   \\ 
		\end{tabular}}
	\vspace{-.2cm}
\end{table}

\begin{table}
\centering
\caption{\small Quantitative results via plugging our DGMN module on different backbones on the COCO 2017 \texttt{test-dev} set for object detection ($\mathrm{AP^{box}}$) and instance segmentation ($\mathrm{AP^{mask}}$).}
\label{tab:test_coco}
\scalebox{0.8}{
\tablestyle{1pt}{1}
\begin{tabular}{l|x{40}|x{32}x{32}x{32}|x{32}x{32}x{32}}
&\scriptsize Backbone &\scriptsize $\mathrm{AP^{box}}$ &\scriptsize $\mathrm{AP^{box}_{50}}$ &\scriptsize $\mathrm{AP^{box}_{75}}$ &\scriptsize $\mathrm{AP^{mask}}$ &\scriptsize $\mathrm{AP^{mask}_{50}}$ &\scriptsize $\mathrm{AP^{mask}_{75}}$ \\
\shline
\scriptsize Mask R-CNN &\scriptsize \multirow{3}{*}{ResNet 50} &\scriptsize 38.0 &\scriptsize59.7 &\scriptsize41.5 &\scriptsize 34.6 &\scriptsize56.5 &\scriptsize36.6\\
\scriptsize+ DGMN (C5) & &\scriptsize40.2  &\scriptsize 62.5 &\scriptsize 43.9 &\scriptsize 36.2 &\scriptsize 59.1 &\scriptsize  38.4\\
\scriptsize+ DGMN (C4, C5)& &\scriptsize \bf41.0 &\scriptsize \bf63.2 &\scriptsize \bf44.9 &\scriptsize \bf36.8 &\scriptsize \bf59.8 &\scriptsize \bf39.1 \\
\hline
\scriptsize Mask R-CNN & \scriptsize  \multirow{2}{*}{ResNet 101}   &\scriptsize40.2 &\scriptsize61.9 &\scriptsize44.0 &\scriptsize36.2  &\scriptsize58.6 &\scriptsize38.4\\
\scriptsize+ DGMN (C5)& &\scriptsize \bf 41.9  &\scriptsize \bf 64.1&\scriptsize \bf 45.9 &\scriptsize \bf 37.6 &\scriptsize \bf 60.9 &\scriptsize \bf 40.0 \\
\hline
\scriptsize Mask R-CNN&\scriptsize \multirow{2}{*}{ResNeXt 101}  &\scriptsize 42.6 &\scriptsize 64.9 &\scriptsize 46.6 &\scriptsize 38.3  &\scriptsize 61.6 &\scriptsize 40.8\\
\scriptsize+ DGMN (C5)& &\scriptsize \bf 44.3   &\scriptsize \bf 66.8 &\scriptsize \bf 48.4  &\scriptsize \bf 39.5 &\scriptsize \bf 63.3 &\scriptsize \bf 42.1 \\
\end{tabular}
}
\end{table}

\subsection{Comparison to state of the art}

\label{sec:exp_sota}
\textbf{Performance on Cityscapes test set.}
Table~\ref{tab:city_test} compares our approach with state-of-the-art methods on Cityscapes.
Note that all the methods are trained using only the fine annotations and evaluated on the public evaluation server as test-set annotations are withheld from the public.
As shown in the table, DGMN (ours) achieves an mIoU of 81.6\%, surpassing all previous works. 
Among competing methods, GloRe~\cite{graph_reason}, Non-local~\cite{wang2018nonlocal}, CCNet~\cite{huang2018ccnet} and DANet~\cite{DAnet} are the most related to us as they all based on graph neural network modules. 
Note that we followed common practice and employed several complementary strategies used in semantic segmentation to boost performance,
including Online Hard Example Mining (OHEM)~\cite{DAnet}, Multi-Grid~\cite{deeplabv3} and Multi-Scale (MS) ensembling~\cite{pspnet}. 
The contribution of each strategy on the final performance is reported in the supplementary. 

\par\noindent\textbf{Performance on COCO 2017 test set.} 
Table \ref{tab:test_coco} presents our results on the COCO test-dev set, where we inserted our module on multiple backbones. 
By inserting DGMN into all layers of C4 and C5, we substantially improve the performance of Mask R-CNN, observing a gain of of 3.0 and 2.2 points on the AP$^{\mathrm{box}}$ and the AP$^{\mathrm{mask}}$ of object detection and instance segmentation respectively.
We observe similar improvements when using the ResNet-101 or ResNeXt-101 backbones as well, showing that our proposed DGMN module generalises to multiple backbone architectures.

Note that the Mask-RCNN baseline with ResNet-101 has 63.1M parameters and 354 GFLOPs.
Our DGMN (C4, C5) model with ResNet-50 outperforms it whilst having only 51.1M parameters and 297.1 GFLOPs.
This further shows that the improvements from our method are due to the model design, and not only the increased parameters and computation.
Further comparisons to the state of the art are included in the supplementary.

\subsection{Qualitative results}
\label{sec:app_qualitative}

\begin{figure*}[t]
	\centering
	\includegraphics[width=1.0\linewidth]{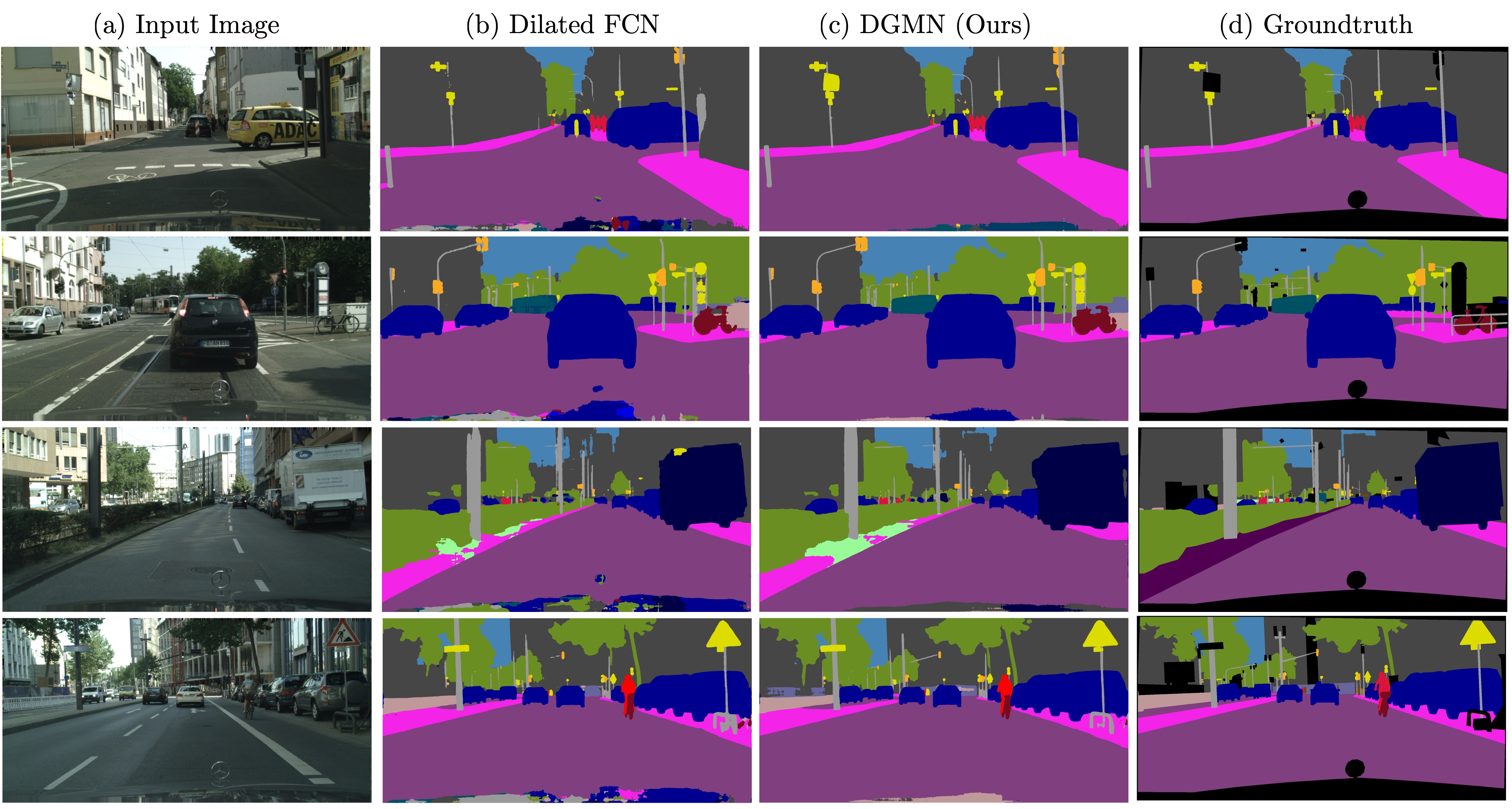}
	\caption{Qualitative results of the Dilated FCN baseline~\cite{yu2015multi} and our proposed DGMN model on the Cityscapes dataset.}
	\label{fig:cs_results}
\end{figure*}

\begin{figure*}[h]
	\def \imheight {3.6cm}
	\centering
	\rotatebox{90}{\textcolor{white}{------}Mask R-CNN} \hspace{0.05cm}
	\includegraphics[height=\imheight, align=b]{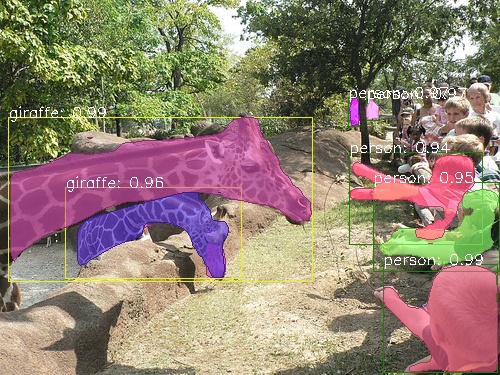}
	\includegraphics[height=\imheight, align=b]{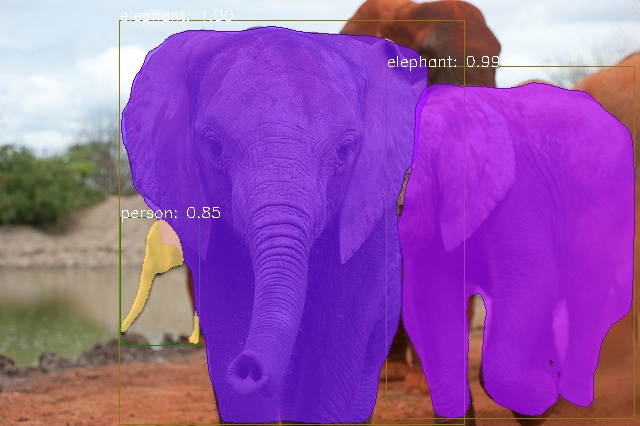}
	\includegraphics[height=\imheight, align=b]{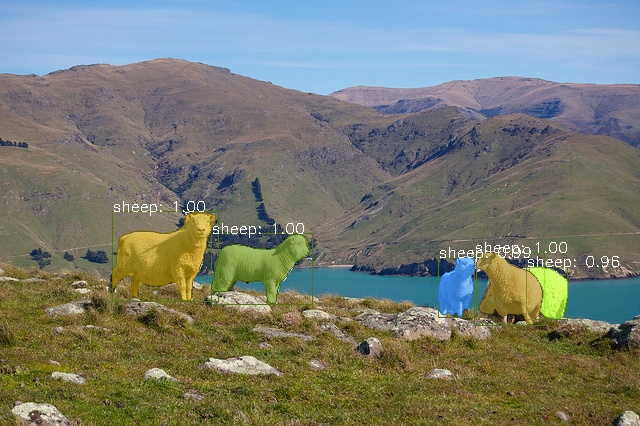}
	\\
	\rotatebox{90}{\textcolor{white}{------}DGMN (ours)} \hspace{0.05cm}
	\includegraphics[height=\imheight, align=b]{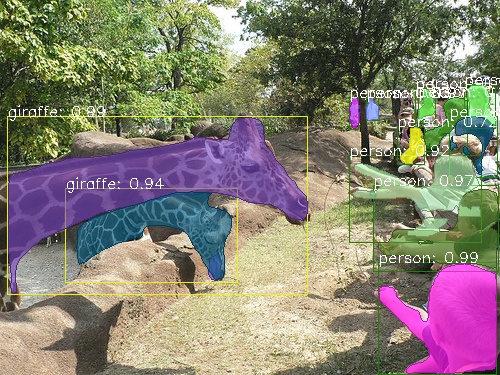}
	\includegraphics[height=\imheight, align=b]{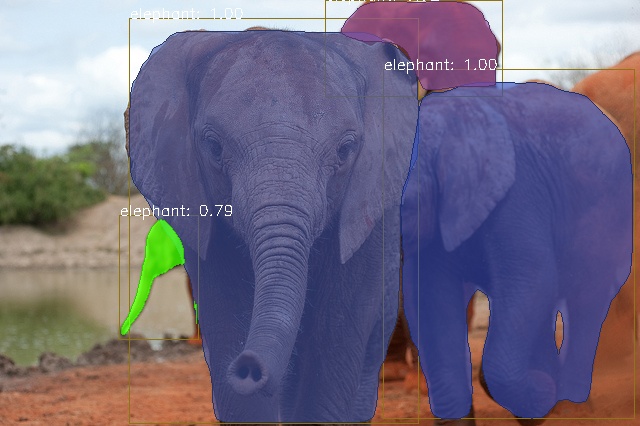}
	\includegraphics[height=\imheight, align=b]{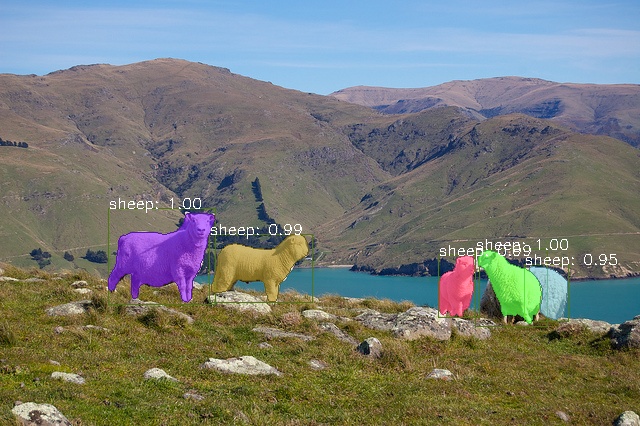}
	\\
	
	\rotatebox{90}{\textcolor{white}{------}Mask R-CNN} \hspace{0.05cm}
	\includegraphics[height=\imheight, align=b]{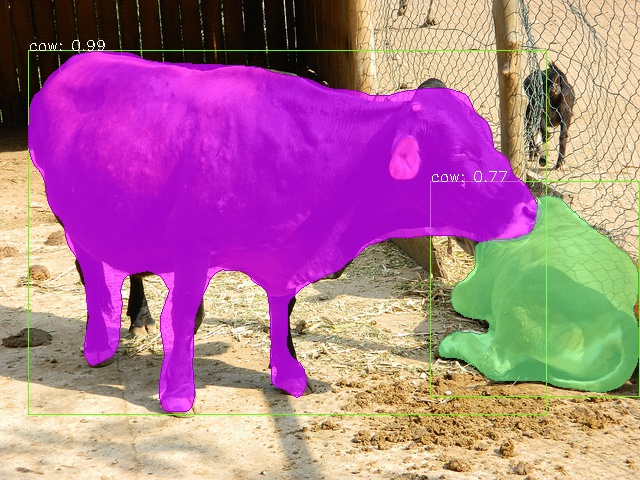}
	\includegraphics[height=\imheight, align=b]{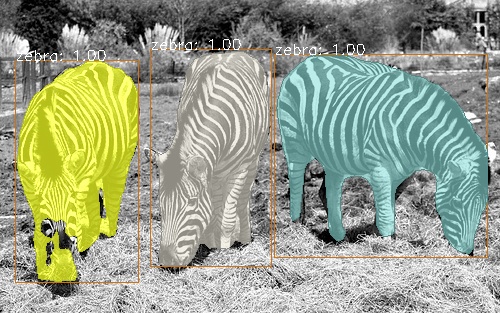}
	\includegraphics[height=\imheight, align=b]{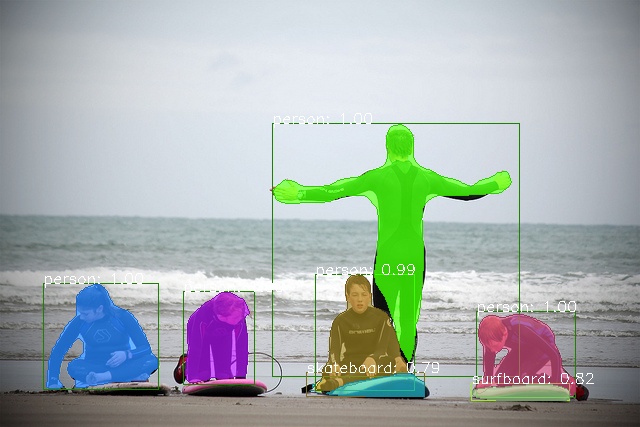}
	\\
	\rotatebox{90}{\textcolor{white}{------}DGMN (ours)} \hspace{0.05cm}
	\includegraphics[height=\imheight, align=b]{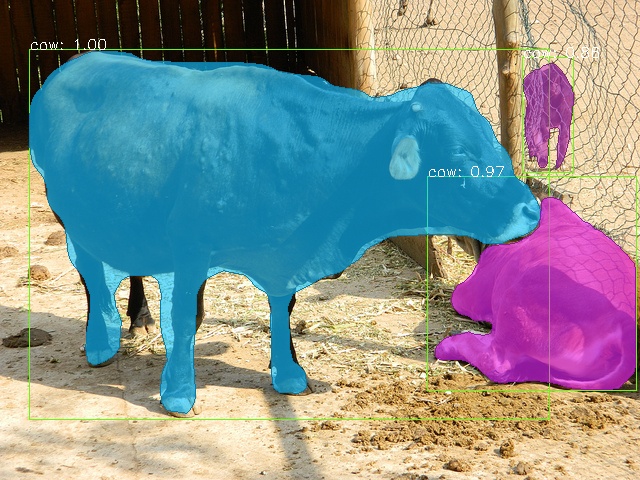}
	\includegraphics[height=\imheight, align=b]{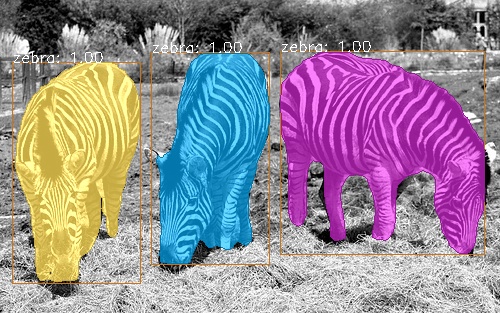}
	\includegraphics[height=\imheight, align=b]{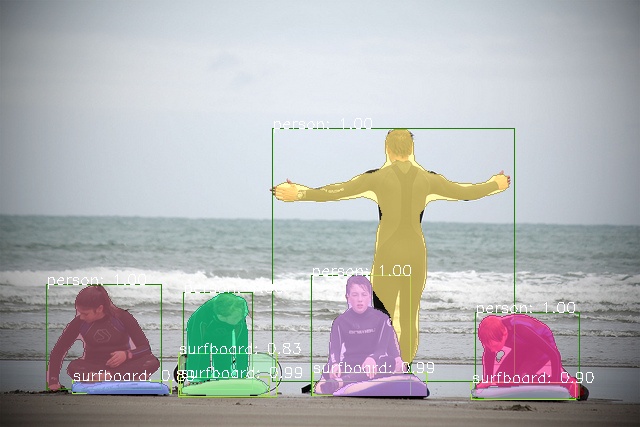}
	\\
	
	\rotatebox{90}{\textcolor{white}{------}Mask R-CNN} \hspace{0.05cm}
	\includegraphics[height=\imheight, align=b]{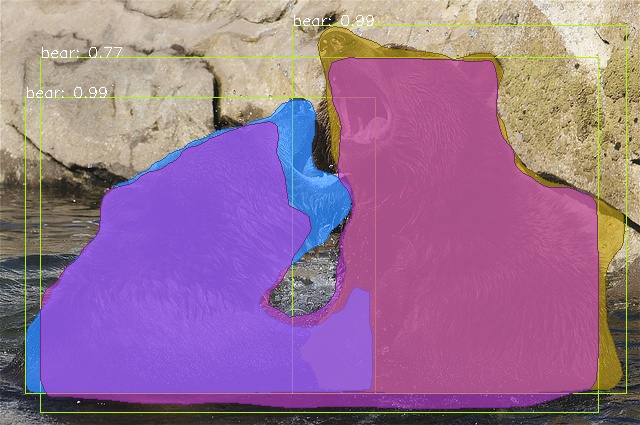}
	\includegraphics[height=\imheight, align=b]{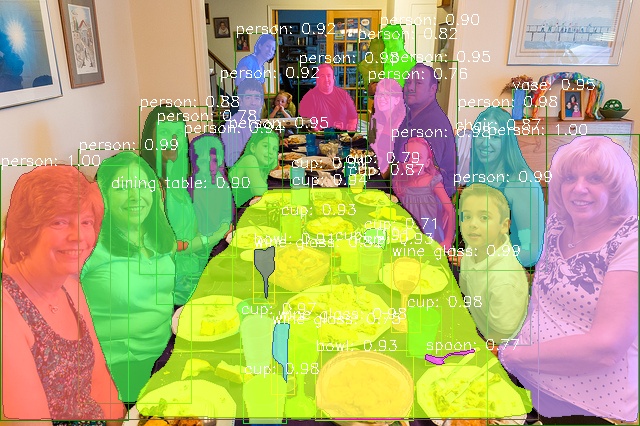}
	\includegraphics[height=\imheight, align=b]{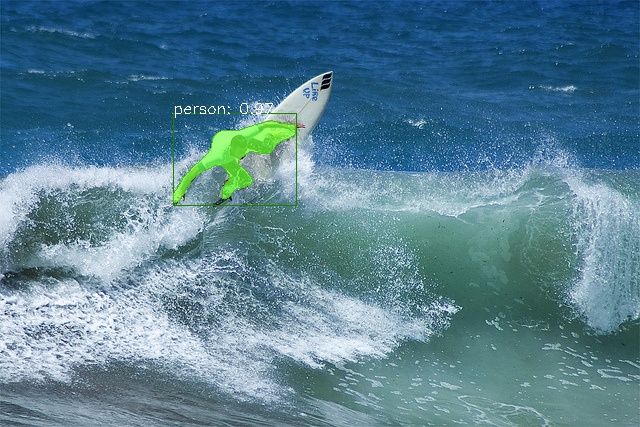}
	\\
	\rotatebox{90}{\textcolor{white}{------}DGMN (ours)} \hspace{0.05cm}
	\includegraphics[height=\imheight, align=b]{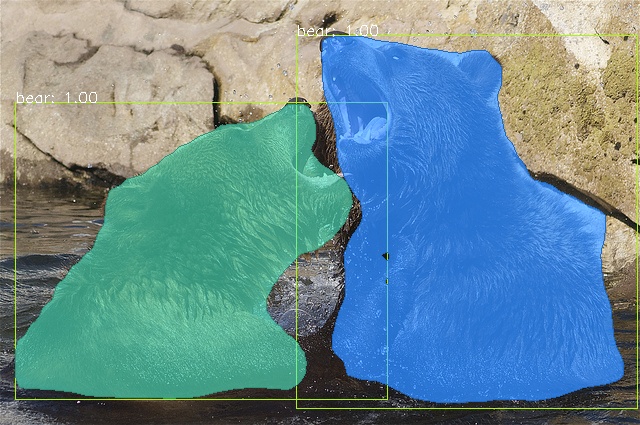}
	\includegraphics[height=\imheight, align=b]{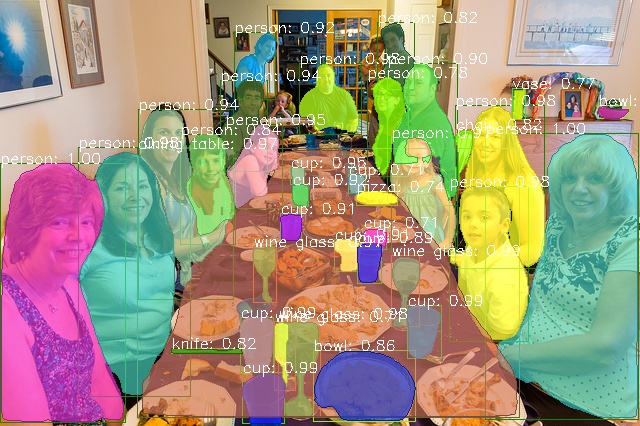}
	\includegraphics[height=\imheight, align=b]{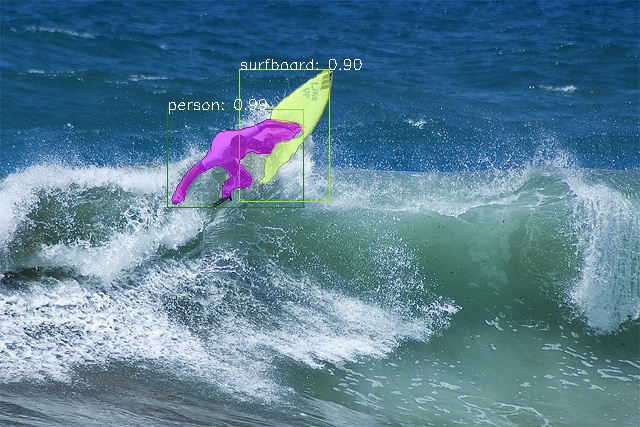}
	\\
	
	\caption{Qualitative examples of the instance segmentation task on the COCO validation dataset. 
	The odd rows are the results from the Mask R-CNN baseline~\cite{massa2018mrcnn,maskrcnn}. 
	The even rows are the results from our DGMN approach.
	Note how our approach often produces better segmentations and fewer false-positive and false-negative detections.
	}
	\label{fig:coco}
\end{figure*}

\begin{figure*}[htb]
	\def \imheight {3.6cm}
	\centering
	
	\rotatebox{90}{\textcolor{white}{------}Mask R-CNN} \hspace{0.05cm}
	\includegraphics[height=\imheight, align=b]{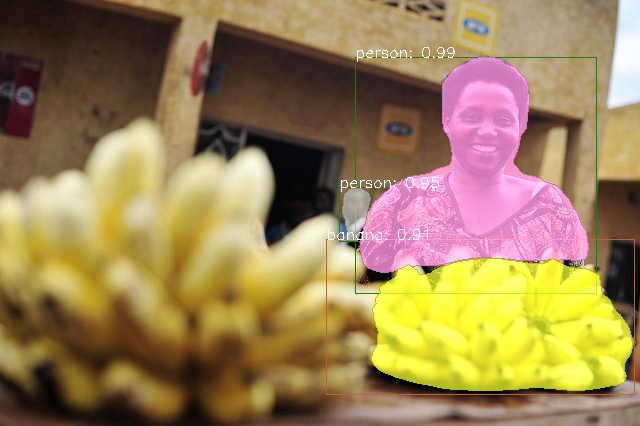}
	\includegraphics[height=\imheight, align=b]{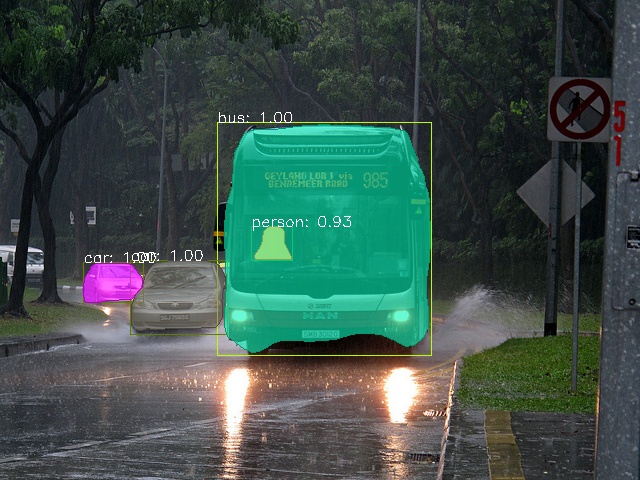}
	\includegraphics[height=\imheight, align=b]{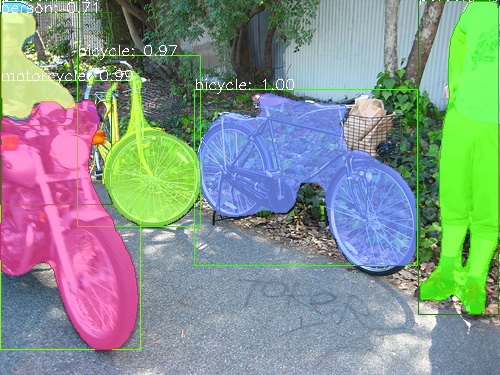}
	\\
	\rotatebox{90}{\textcolor{white}{------}DGMN (ours)} \hspace{0.05cm}
	\includegraphics[height=\imheight, align=b]{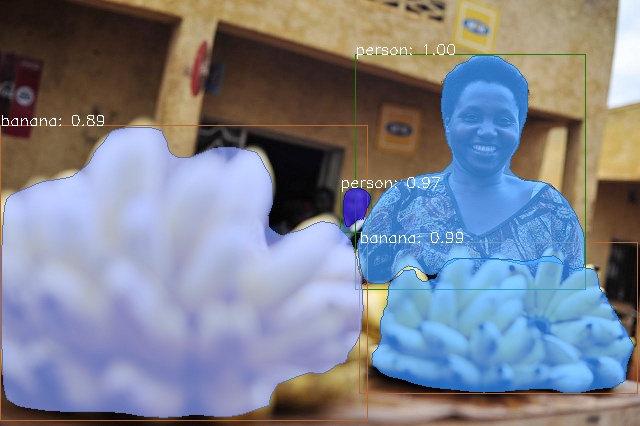}
	\includegraphics[height=\imheight, align=b]{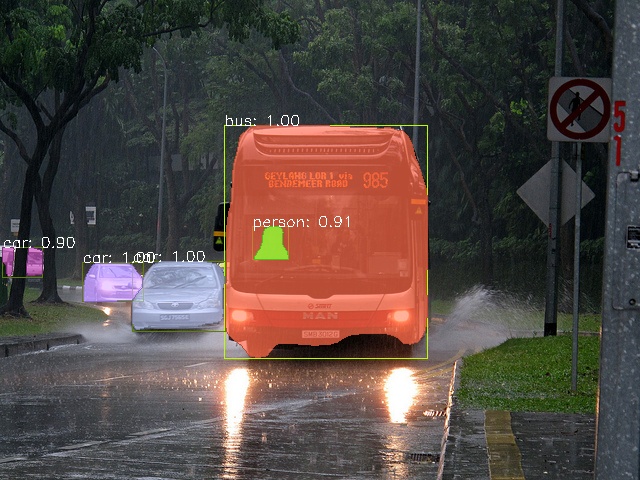}
	\includegraphics[height=\imheight, align=b]{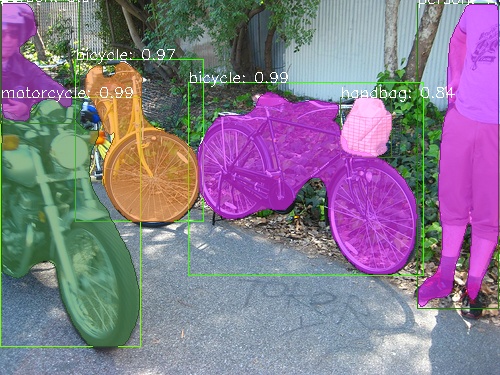}
	\\
	
	\rotatebox{90}{\textcolor{white}{------}Mask R-CNN} \hspace{0.05cm}
	\includegraphics[height=\imheight, align=b]{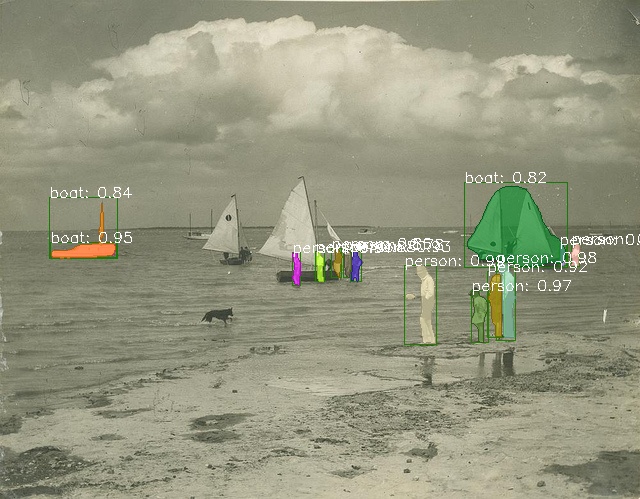}
	\includegraphics[height=\imheight, align=b]{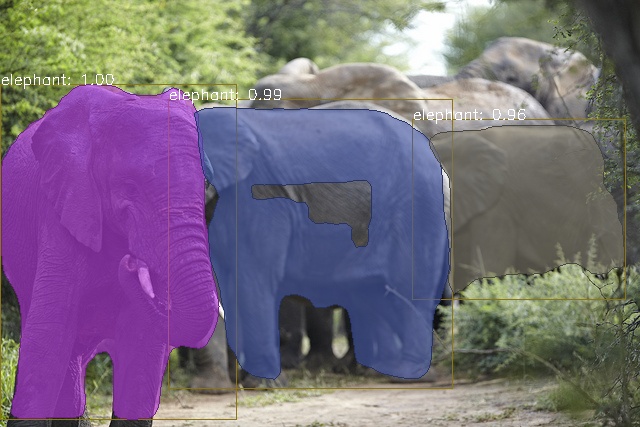}
	\includegraphics[height=\imheight, align=b]{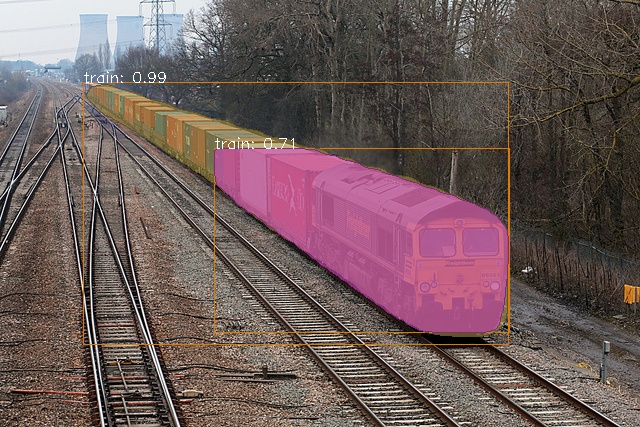}
	\\
	\rotatebox{90}{\textcolor{white}{------}DGMN (ours)} \hspace{0.05cm}
	\includegraphics[height=\imheight, align=b]{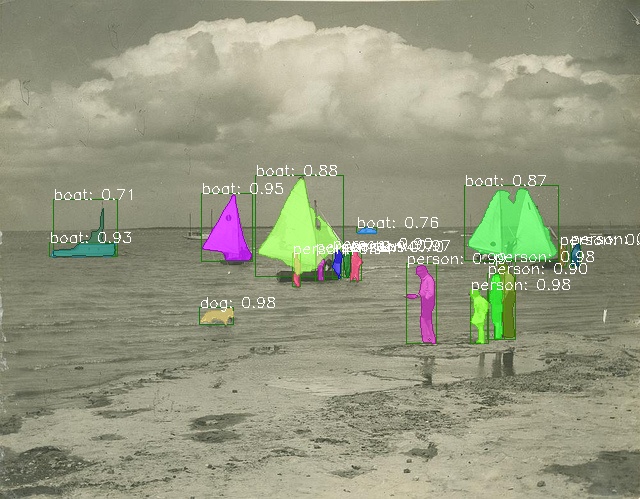}
	\includegraphics[height=\imheight, align=b]{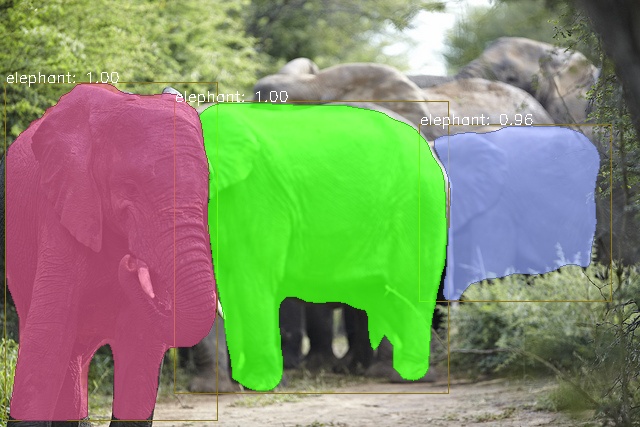}
	\includegraphics[height=\imheight, align=b]{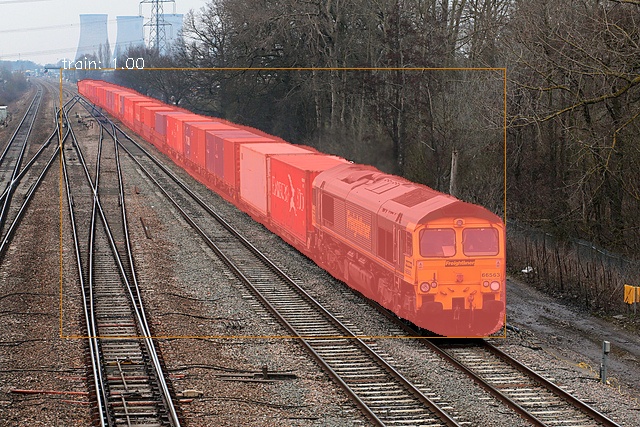}
	\\
	
	\rotatebox{90}{\textcolor{white}{------}Mask R-CNN} \hspace{0.05cm}
	\includegraphics[height=\imheight, align=b]{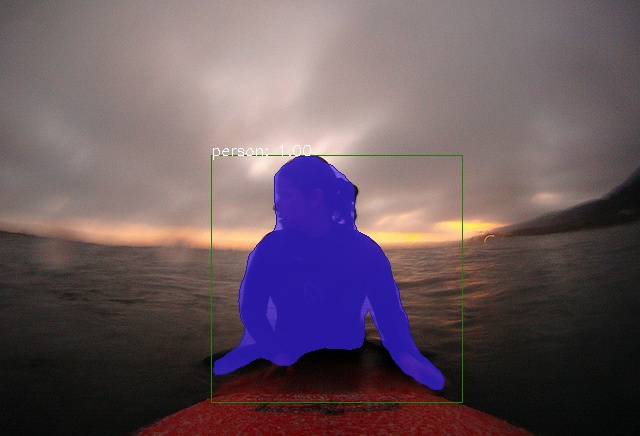}
	\includegraphics[height=\imheight, align=b]{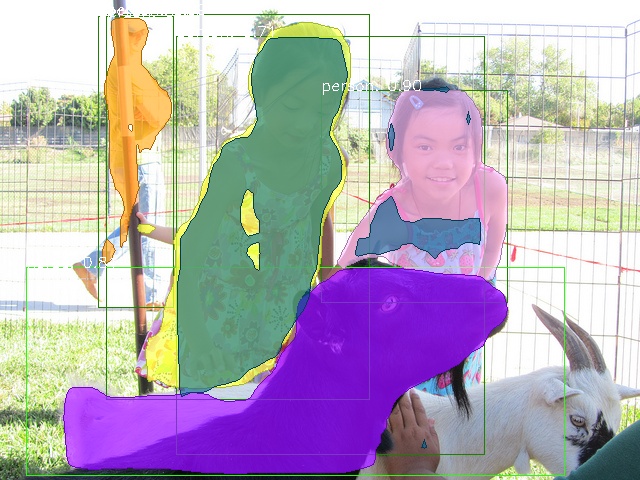}
	\includegraphics[height=\imheight, align=b]{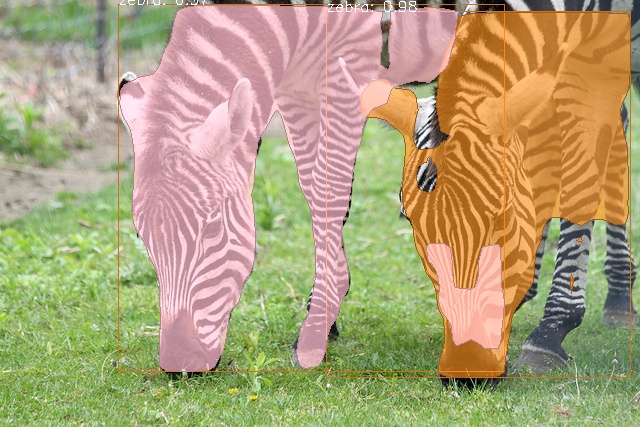}
	\\
	\rotatebox{90}{\textcolor{white}{------}DGMN (ours)} \hspace{0.05cm}
	\includegraphics[height=\imheight, align=b]{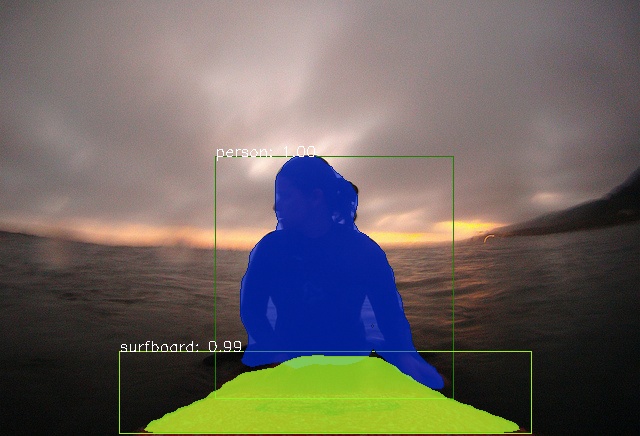}
	\includegraphics[height=\imheight, align=b]{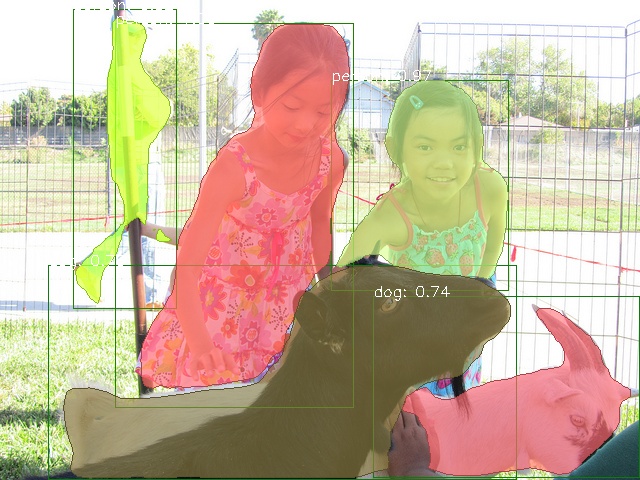}
	\includegraphics[height=\imheight, align=b]{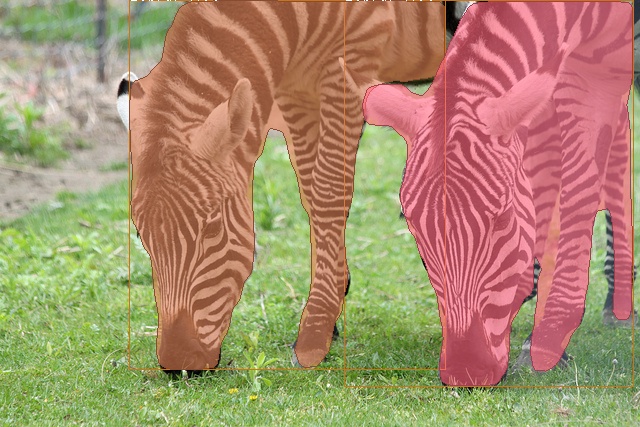}
	
	\caption{More qualitative examples of the instance segmentation task on the COCO validation dataset. 
	The odd rows are the results from the Mask R-CNN baseline~\cite{massa2018mrcnn,maskrcnn}. 
	The even rows are the detection results from our DGMN approach.
	Note how our approach often produces better segmentations and fewer false-positive and false-negative detections.
	}
	\label{fig:coco2}
\end{figure*}

Qualitative examples of our network's results can be found in Fig~\ref{fig:qualitative}.
Fig.~\ref{fig:cs_results} shows more qualitative results for semantic segmentation (on Cityscapes) and 
Fig.~\ref{fig:coco} and \ref{fig:coco2} show more qualitative results for object detection and instance segmentation (on COCO).

\section{Experiments with transformer architectures}

In this section,
we compare our DGMN2 model to representative backbones for the image classification, object detection, instance segmentation and semantic segmentation tasks.

\begin{table}
\centering
\caption{\small 
Evaluation results on ILSVRC-2012 ImageNet-1K~\cite{imagenet} validation set.
We report the results using the input size of 224x224 and center cropped from resized images with 256x256 pixels.
}
\label{tab:dgmn2_imagenet_val}
\scalebox{1}{
\tablestyle{1pt}{1}
\begin{tabular}{l|x{40}|x{40}|x{40}}
Method &\scriptsize Params (M) &\scriptsize FLOPs (G) &\scriptsize Top1 Acc (\%) \\
\shline
\scriptsize ResNet-18~\cite{resnet} &\scriptsize11.7 &\scriptsize1.8 &\scriptsize68.5 \\
\scriptsize DeiT-Tiny/16~\cite{touvron2020training} &\scriptsize \textbf{5.7} &\scriptsize \textbf{1.3} &\scriptsize72.2 \\
\scriptsize PVT-Tiny~\cite{wang2021pyramid} &\scriptsize13.2 &\scriptsize1.9 &\scriptsize75.1 \\
\scriptsize DGMN2-Tiny &\scriptsize12.1 &\scriptsize2.3 &\scriptsize \textbf{78.7} \\
\hline
\scriptsize ResNet-50~\cite{resnet} &\scriptsize25.6 &\scriptsize 4.1 &\scriptsize78.5 \\
\scriptsize DeiT-Small/16~\cite{touvron2020training} &\scriptsize22.1 &\scriptsize4.6 &\scriptsize79.9 \\
\scriptsize PVT-Small~\cite{wang2021pyramid} &\scriptsize24.5 &\scriptsize \textbf{3.8} &\scriptsize79.8 \\
\scriptsize Swin-T~\cite{liu2021swin} &\scriptsize29.0 &\scriptsize4.5 &\scriptsize81.3 \\
\scriptsize DGMN2-Small &\scriptsize \textbf{21.0} &\scriptsize4.3 &\scriptsize \textbf{81.7} \\
\hline
\scriptsize ResNet-101~\cite{resnet} &\scriptsize44.7 &\scriptsize7.9 &\scriptsize79.8 \\
\scriptsize DeiT-Medium/16~\cite{touvron2020training} &\scriptsize48.8 &\scriptsize9.9 &\scriptsize80.8 \\
\scriptsize PVT-Medium~\cite{wang2021pyramid} &\scriptsize44.2 &\scriptsize \textbf{6.7} &\scriptsize81.2 \\
\scriptsize DGMN2-Medium &\scriptsize \textbf{35.8} &\scriptsize7.1 &\scriptsize \textbf{82.5} \\
\hline
\scriptsize ResNeXt-101-64x4d~\cite{xie2017aggregated} &\scriptsize83.5 &\scriptsize15.6 &\scriptsize81.5 \\
\scriptsize ViT-Base/16~\cite{dosovitskiy2020image} &\scriptsize86.6 &\scriptsize17.6 &\scriptsize81.8 \\
\scriptsize DeiT-Large/16~\cite{touvron2020training} &\scriptsize86.6 &\scriptsize17.6 &\scriptsize81.8 \\
\scriptsize PVT-Large~\cite{wang2021pyramid} &\scriptsize61.4 &\scriptsize 9.8 &\scriptsize81.7 \\
\scriptsize Swin-S~\cite{liu2021swin} &\scriptsize50.0 &\scriptsize \textbf{8.7} &\scriptsize 83.0 \\
\scriptsize SOFT-Large~\cite{lu2021soft} &\scriptsize 64.0 &\scriptsize11.0 &\scriptsize 83.1 \\
\scriptsize DGMN2-Large &\scriptsize \textbf{48.3} &\scriptsize10.4 &\scriptsize \textbf{83.3} \\
\end{tabular}
}
\end{table}

\subsection{Image classification}
\par\noindent\textbf{Datasets.} ImageNet-1K~\cite{imagenet} is the most popular classification dataset, which contains 1.28M training images and 50K validation images with 1,000 classes.

\par\noindent\textbf{Implementation details.} 
We evaluate the proposed DGMN2 on ImageNet-1K~\cite{imagenet} for image classification task. 
Our models are trained with 1.28M training images and evaluate on 50K validation images. We report the Top-1 accuracy on validation set.
During training, we follow the training hyperparameters of recent work~\cite{touvron2020training, liu2021swin} and use random-size cropping to 224 $\times$ 224, random horizontal flipping~\cite{googlenet}, and mix up~\cite{zhang2017mixup} for data augmentation.
Optimization is performed using AdamW~\cite{loshchilov2017decoupled} with momentum of 0.9 and a batch size of 1024. 
The weight decay is set to $5 \times 10^{-2}$ by default. 
The initial learning rate is set to $1 \times 10^{-3}$ and decreases following the cosine schedule~\cite{loshchilov2016sgdr}. 
We conduct 300 epochs training from scratch using 8 NVIDIA Tesla V100 GPUs. 
FLOPs are calculated at batch size 1 with input of resolution 224 $\times$ 224.

\par\noindent\textbf{Results on ImageNet.}
We compare with state-of-the-art methods and report the Top-1 accuracy on the ImageNet-1K validation set. 
From Table~\ref{tab:dgmn2_imagenet_val} we can make following observations:
\textbf{(i)}
Transformer based models yield better classification accuracy over convolution counterparts;
\textbf{(ii)}
Our proposed DGMN2 achieves best performance among recent pure Transformer based backbones such as DeiT~\cite{touvron2020training}, the strong version of  ViT~\cite{dosovitskiy2020image};
\textbf{(iii)} 
We outperform the most similar (in architecture configuration) Transformer counterparts PVT~\cite{wang2021pyramid} at all variants and with fewer parameters and FLOPs.
\textbf{(iv)} 
We also beat the latest alternatives Swin Transformer~\cite{liu2021swin} and SOFT~\cite{lu2021soft} which are designed to address the efficiency limitation of Transformer based vision backbone.
Particularly, our DGMN2-Small is 0.4\% higher and 8M parameters fewer than Swin-T. Moreover, DGMN2-Large is 0.3\% higher and 1.7M parameters fewer than Swin-S.

\subsection{Object detection}

\begin{table}[t]
\centering
\caption{\small Object detection performance using our DGMN2 with RetinaNet~\cite{lin2017focal} on COCO validation set.
}
\label{tab:dgmn2_coco_object_detection_val}
\scalebox{0.85}{
\tablestyle{1pt}{1}
\begin{tabular}{l|x{36}|x{36}|x{20}x{20}x{20}x{20}x{20}x{20}}
 \scriptsize Backbone &\scriptsize Params (M) &\scriptsize FLOPs (G) &\scriptsize $\rm AP$ &\scriptsize $\rm AP_{50}$ &\scriptsize $\rm AP_{75}$ &\scriptsize $\rm AP_S$ &\scriptsize $\rm AP_M$ &\scriptsize $\rm AP_L$ \\
\shline
\scriptsize ResNet-18~\cite{resnet} &\scriptsize \textbf{21.3} &\scriptsize \textbf{198.2} &\scriptsize31.8 &\scriptsize49.6 &\scriptsize33.6 &\scriptsize16.3 &\scriptsize34.3 &\scriptsize43.2 \\
\scriptsize PVT-Tiny~\cite{wang2021pyramid} &\scriptsize23.0 &\scriptsize 230.2 &\scriptsize36.7 &\scriptsize56.9 &\scriptsize38.9 &\scriptsize22.6 &\scriptsize38.8 &\scriptsize50.0 \\
\scriptsize DGMN2-Tiny &\scriptsize21.8 &\scriptsize 206.8 &\scriptsize \textbf{39.7} &\scriptsize \textbf{60.5} &\scriptsize \textbf{42.2} &\scriptsize \textbf{23.5} &\scriptsize \textbf{43.1} &\scriptsize \textbf{52.1} \\
\hline
\scriptsize ResNet-50~\cite{resnet} &\scriptsize37.7 &\scriptsize 251.3 &\scriptsize36.3 &\scriptsize55.3 &\scriptsize38.6 &\scriptsize19.3 &\scriptsize40.0 &\scriptsize48.8 \\
\scriptsize PVT-Small~\cite{wang2021pyramid} &\scriptsize34.2 &\scriptsize 298.0 &\scriptsize40.4 &\scriptsize61.3 &\scriptsize43.0 &\scriptsize25.0 &\scriptsize42.9 &\scriptsize55.7 \\
\scriptsize DGMN2-Small &\scriptsize \textbf{30.7} &\scriptsize \textbf{248.2} &\scriptsize \textbf{42.5} &\scriptsize \textbf{63.3} &\scriptsize \textbf{45.3} &\scriptsize \textbf{25.3} &\scriptsize \textbf{46.1} &\scriptsize \textbf{56.5} \\
\hline
\scriptsize ResNet-101~\cite{resnet} &\scriptsize56.7 &\scriptsize 331.2 &\scriptsize38.5 &\scriptsize57.8 &\scriptsize41.2 &\scriptsize21.4 &\scriptsize42.6 &\scriptsize51.1 \\
\scriptsize PVT-Medium~\cite{wang2021pyramid} &\scriptsize53.9 &\scriptsize 389.3 &\scriptsize41.9 &\scriptsize63.1 &\scriptsize44.3 &\scriptsize25.0 &\scriptsize44.9 &\scriptsize57.6 \\
\scriptsize DGMN2-Medium &\scriptsize \textbf{45.6} &\scriptsize \textbf{312.0} &\scriptsize \textbf{43.7} &\scriptsize \textbf{64.8} &\scriptsize \textbf{46.6} &\scriptsize \textbf{26.3} &\scriptsize \textbf{47.2} &\scriptsize \textbf{58.3} \\
\hline
\scriptsize ResNeXt-101-64x4d~\cite{xie2017aggregated} &\scriptsize95.5 &\scriptsize 496.3 &\scriptsize41.0 &\scriptsize60.9 &\scriptsize44.0 &\scriptsize23.9 &\scriptsize45.2 &\scriptsize54.0 \\
\scriptsize PVT-Large~\cite{wang2021pyramid} &\scriptsize71.1 &\scriptsize 496.1 &\scriptsize42.6 &\scriptsize63.7 &\scriptsize45.4 &\scriptsize25.8 &\scriptsize46.0 &\scriptsize58.4 \\
\scriptsize DGMN2-Large &\scriptsize \textbf{57.9} &\scriptsize \textbf{382.1} &\scriptsize \textbf{44.7} &\scriptsize \textbf{65.4} &\scriptsize \textbf{48.0} &\scriptsize \textbf{26.6} &\scriptsize \textbf{48.7} &\scriptsize \textbf{60.6} \\
\end{tabular}
}
\end{table}

\begin{table}[t]
\centering
\caption{\small Instance segmentation performance using our DGMN2 with Mask R-CNN~\cite{maskrcnn} on COCO validation set. $\rm AP^b$ denotes bounding box AP and $\rm AP^m$ denotes mask AP.}
\label{tab:dgmn2_coco_instance_segmentation_val}
\scalebox{0.85}{
\tablestyle{1pt}{1}
\begin{tabular}{l|x{36}|x{36}|x{20}x{20}x{20}x{20}x{20}x{20}}
 \scriptsize Backbone &\scriptsize Params (M) &\scriptsize FLOPs (G) &\scriptsize $\rm AP^b$ &\scriptsize $\rm AP^b_{50}$ &\scriptsize $\rm AP^b_{75}$ &\scriptsize $\rm AP^m$ &\scriptsize $\rm AP^m_{50}$ &\scriptsize $\rm AP^m_{75}$ \\
\shline
\scriptsize ResNet-18~\cite{resnet} &\scriptsize \textbf{31.2} &\scriptsize \textbf{214.3} &\scriptsize34.0 &\scriptsize54.0 &\scriptsize36.7 &\scriptsize31.2 &\scriptsize51.0 &\scriptsize32.7 \\
\scriptsize PVT-Tiny~\cite{wang2021pyramid} &\scriptsize32.9 &\scriptsize 247.3 &\scriptsize36.7 &\scriptsize59.2 &\scriptsize39.3 &\scriptsize35.1 &\scriptsize56.7 &\scriptsize37.3 \\
\scriptsize DGMN2-Tiny &\scriptsize31.8 &\scriptsize 223.0 &\scriptsize \textbf{40.1} &\scriptsize \textbf{62.4} &\scriptsize \textbf{43.8} &\scriptsize \textbf{37.2} &\scriptsize \textbf{59.7} &\scriptsize \textbf{39.6} \\
\hline
\scriptsize ResNet-50~\cite{resnet} &\scriptsize44.2 &\scriptsize 269.8 &\scriptsize38.0 &\scriptsize58.6 &\scriptsize41.4 &\scriptsize34.4 &\scriptsize55.1 &\scriptsize36.7 \\
\scriptsize PVT-Small~\cite{wang2021pyramid} &\scriptsize44.1 &\scriptsize 315.1 &\scriptsize40.4 &\scriptsize62.9 &\scriptsize43.8 &\scriptsize37.8 &\scriptsize60.1 &\scriptsize40.3 \\
\scriptsize DGMN2-Small &\scriptsize \textbf{40.6} &\scriptsize \textbf{264.5} &\scriptsize \textbf{43.4} &\scriptsize \textbf{65.4} &\scriptsize \textbf{47.8} &\scriptsize \textbf{39.7} &\scriptsize \textbf{62.5} &\scriptsize \textbf{42.6} \\
\hline
\scriptsize ResNet-101~\cite{resnet} &\scriptsize63.2 &\scriptsize 349.7 &\scriptsize40.4 &\scriptsize61.1 &\scriptsize44.2 &\scriptsize36.4 &\scriptsize57.7 &\scriptsize38.8 \\
\scriptsize PVT-Medium~\cite{wang2021pyramid} &\scriptsize63.9 &\scriptsize 406.4 &\scriptsize42.0 &\scriptsize64.4 &\scriptsize45.6 &\scriptsize39.0 &\scriptsize61.6 &\scriptsize42.1 \\
\scriptsize DGMN2-Medium &\scriptsize \textbf{55.5} &\scriptsize \textbf{328.2} &\scriptsize \textbf{44.4} &\scriptsize \textbf{66.3} &\scriptsize \textbf{48.6} &\scriptsize \textbf{40.2} &\scriptsize \textbf{63.3} &\scriptsize \textbf{43.1} \\
\hline
\scriptsize ResNeXt-101-64x4d~\cite{xie2017aggregated} &\scriptsize101.9 &\scriptsize 514.7 &\scriptsize42.8 &\scriptsize63.8 &\scriptsize47.3 &\scriptsize38.4 &\scriptsize60.6 &\scriptsize41.3 \\
\scriptsize PVT-Large~\cite{wang2021pyramid} &\scriptsize81.0 &\scriptsize 513.1 &\scriptsize42.9 &\scriptsize65.0 &\scriptsize46.6 &\scriptsize39.5 &\scriptsize61.9 &\scriptsize42.5 \\
\scriptsize DGMN2-Large &\scriptsize \textbf{67.9} &\scriptsize \textbf{398.4} &\scriptsize \textbf{46.2} &\scriptsize \textbf{67.9} &\scriptsize \textbf{50.7} &\scriptsize \textbf{41.6} &\scriptsize \textbf{65.1} &\scriptsize \textbf{44.6} \\
\end{tabular}
}
\end{table}

\begin{table}[t]
\centering
\caption{\small Object detection performance using our DGMN2 with Deformable DETR~\cite{zhu2020deformable} on COCO validation set. ``*'' indicates that this result is from official GitHub repository, which has improved performance with a fixed bug. ``+'' indicates using iterative bounding box refinement and two-stage Deformable DETR.}
\label{tab:dgmn2_coco_object_detection_deformable_detr_val}
\scalebox{0.9}{
\tablestyle{1pt}{1}
\begin{tabular}{l|x{40}|x{40}|x{16}x{16}x{16}x{16}x{16}x{16}}
 \scriptsize Backbone &\scriptsize Params (M) &\scriptsize FLOPs (G) &\scriptsize $\rm AP$ &\scriptsize $\rm AP_{50}$ &\scriptsize $\rm AP_{75}$ &\scriptsize $\rm AP_S$ &\scriptsize $\rm AP_M$ &\scriptsize $\rm AP_L$ \\
\shline
\scriptsize ResNet-50~\cite{resnet} &\scriptsize39.9 &\scriptsize 206.7 &\scriptsize43.8 &\scriptsize62.6 &\scriptsize47.7 &\scriptsize26.4 &\scriptsize47.1 &\scriptsize58.0 \\
\scriptsize ResNet-50*~\cite{resnet} &\scriptsize39.9 &\scriptsize 206.7 &\scriptsize44.5 &\scriptsize63.6 &\scriptsize48.7 &\scriptsize27.1 &\scriptsize47.6 &\scriptsize59.6 \\
\scriptsize DGMN2-Tiny &\scriptsize \textbf{24.0} &\scriptsize \textbf{163.6} &\scriptsize44.4 &\scriptsize63.5 &\scriptsize48.3 &\scriptsize 26.4 &\scriptsize48.0 &\scriptsize57.7 \\
\scriptsize DGMN2-Small &\scriptsize32.9 &\scriptsize 205.1 &\scriptsize47.3 &\scriptsize66.9 &\scriptsize51.4 &\scriptsize28.9 &\scriptsize51.2 &\scriptsize62.1 \\
\scriptsize DGMN2-Medium &\scriptsize 47.7 &\scriptsize 268.9 &\scriptsize \textbf{48.4} &\scriptsize \textbf{68.0} &\scriptsize \textbf{53.0} &\scriptsize \textbf{29.9} &\scriptsize \textbf{52.1} &\scriptsize \textbf{64.0} \\
\hline
\scriptsize ResNet-50+~\cite{resnet} &\scriptsize41.0 &\scriptsize 215.9 &\scriptsize46.2 &\scriptsize65.2 &\scriptsize50.0 &\scriptsize28.8 &\scriptsize49.2 &\scriptsize61.7 \\
\scriptsize ResNet-50*+~\cite{resnet} &\scriptsize41.0 &\scriptsize 215.9 &\scriptsize46.9 &\scriptsize65.6 &\scriptsize51.0 &\scriptsize29.6 &\scriptsize50.1 &\scriptsize61.6 \\
\scriptsize DGMN2-Small+ &\scriptsize \textbf{34.0} &\scriptsize \textbf{214.3} &\scriptsize \textbf{48.5} &\scriptsize \textbf{68.0} &\scriptsize \textbf{52.5} &\scriptsize \textbf{30.6} &\scriptsize \textbf{52.1} &\scriptsize \textbf{63.9} \\
\end{tabular}
}
\end{table}

\begin{table}[t]
\centering
\caption{\small Object detection performance using our DGMN2 with Sparse R-CNN~\cite{sun2021sparse} on COCO validation set. All the models use 300 learnable proposal boxes.}
\label{tab:dgmn2_coco_object_detection_sparse_rcnn_val}
\scalebox{0.9}{
\tablestyle{1pt}{1}
\begin{tabular}{l|x{40}|x{40}|x{16}x{16}x{16}x{16}x{16}x{16}}
 \scriptsize Backbone &\scriptsize Params (M) &\scriptsize FLOPs (G) &\scriptsize $\rm AP$ &\scriptsize $\rm AP_{50}$ &\scriptsize $\rm AP_{75}$ &\scriptsize $\rm AP_S$ &\scriptsize $\rm AP_M$ &\scriptsize $\rm AP_L$ \\
\shline

\scriptsize ResNet-50~\cite{resnet} &\scriptsize106.1 &\scriptsize \textbf{112.82} &\scriptsize45.0 &\scriptsize63.4 &\scriptsize48.2 &\scriptsize26.9 &\scriptsize47.2 &\scriptsize59.5 \\
\scriptsize DGMN2-Small &\scriptsize \textbf{102.6} &\scriptsize 114.86 &\scriptsize \textbf{48.2} &\scriptsize \textbf{67.7} &\scriptsize \textbf{52.7} &\scriptsize \textbf{32.3} &\scriptsize \textbf{51.0} &\scriptsize \textbf{63.2} \\
\end{tabular}
}
\end{table}

\par\noindent\textbf{Implementation details.} 
In this section, we use DGMN2 as the backbone to perform object detection experiments.
We evaluate our backbone on three diverse object detection architectures: RetinaNet~\cite{lin2017focal}, Mask R-CNN~\cite{maskrcnn} and Deformable DETR~\cite{zhang2020dynamic} on the challenging COCO benchmark~\cite{COCO_dataset}.
In all of our experiments, use the ImageNet pretrained weights (from Table~\ref{tab:dgmn2_imagenet_val}) to initialize the backbone and train all models on COCO Train2017, and evaluate on COCO Val2017.
For fair comparisons, we follow the training and test hyperparameters of our baselines.
Specifically, we use AdamW~\cite{loshchilov2017decoupled} optimizer and a batch size of 16 on 8 NVIDIA GPUs to train the model with the initial learning rate of $1 \times 10^{-4}$. 
We use the 1$\times$ schedule (12 epochs with the learning rate decayed by 10$\times$ at epochs 9 and 11) to train the models. The training image is resized to the shorter side of 800 and the longer side does not exceed 1333. During testing, the shorter side of input image is set to 800. We also report the parameters and FLOPs of our models. The FLOPs is calculated using the input resolution of 800 $\times$ 1333.
For the experiments with Deformable DETR, we follow its implementation and use the Adam optimizer with initial learning rate of $2 \times 10^{-4}$ and the weight decay of $10^{-4}$. Note that we use the official implementation of Deformable DETR on GitHub which has improved performance with a fixed bug.

\par\noindent\textbf{Results with RetinaNet~\cite{lin2017focal}.} Table~\ref{tab:dgmn2_coco_object_detection_val} lists the parameters and the bounding box AP of different backbones when using RetinaNet~\cite{lin2017focal}, showing that our DGMN2 models can consistently outperforms ResNet~\cite{resnet} and PVT~\cite{wang2021pyramid} when using similar parameters. Specifically, our DGMN2-Tiny significantly surpasses ResNet-18 by 4.9\% AP and PVT-Tiny by 3.0\% AP using a similar parameters.
Our DGMN2-Small outperforms ResNet-50 by 6.2\% AP and PVT-Small by 2.1\% AP with fewer parameters. 
Moreover, our DGMN2-Medium surpasses ResNet-101 by 5.2\% AP and PVT-Medium by 1.8\% AP with fewer parameters.
Finally, DGMN2-Large surpasses ResNeXt-101-64x4d by 3.7\% AP and surpass PVT-Large by 2.1\% AP with fewer parameters and FLOPs.

\par\noindent\textbf{Results with Mask R-CNN~\cite{maskrcnn}.} Table~\ref{tab:dgmn2_coco_instance_segmentation_val} lists the parameters, the bounding box AP and the mask AP of different backbones when using Mask R-CNN~\cite{maskrcnn}, which also demonstrates the similar results.
Our DGMN2-Tiny surpasses ResNet-18~\cite{resnet} by 6.2\% on bounding box AP ($\rm AP^b$) and 6.0\% on mask AP ($\rm AP^m$) and even surpasses ResNet-50 by about 2.2\% on $\rm AP^b$ and 2.8\% on $\rm AP^m$. When compared to PVT, our DGMN-Tiny outperforms PVT-Tiny by 3.5\% on $\rm AP^b$ and 2.1\% on $\rm AP^m$. Our DGMN2-Small achieves 43.4\% $\rm AP^b$ and 39.7\% $\rm AP^m$, which is 5.4\% and 5.3\% higher than ResNet-50.
Our DGMN2-Small also outperforms PVT-Small by 3.0\% on $\rm AP^b$ and 1.9\% on $\rm AP^b$. 
DGMN2-Medium achieves 44.4\% $\rm AP^b$ and 40.2\% $\rm AP^m$, which is 4.0\% and 3.8\% higher than ResNet-101 and also 2.4\% and 1.2\% higher than PVT-Medium.
Finally, DGMN2-Large achieves 46.2\% $\rm AP^b$ and 41.6\% $\rm AP^m$, which is 3.4\% and 3.2\% higher than ResNeXt-101-64x4d and also 3.3\% and 2.1\% higher than PVT-Large.

\par\noindent\textbf{Results with Deformable DETR~\cite{zhu2020deformable}.} When using a Transformer-based detector (such as Deformable DETR~\cite{zhu2020deformable}), our DGMN2 still have performance improvement under comparable parameters. Table~\ref{tab:dgmn2_coco_object_detection_deformable_detr_val} lists the parameters and the bounding box AP of different backbones when using Deformable DETR, showing that our DGMN2-Tiny achieves comparable performance with ResNet-50~\cite{resnet} with much fewer parameters and FLOPs. For our DGMN2-Small, we surpass ResNet-50 for about 2.8\% AP and also have fewer parameters.

\par\noindent\textbf{Results with Sparse R-CNN~\cite{sun2021sparse}.} When using Sparse R-CNN~\cite{sun2021sparse} as the detector, our DGMN2 models can achieve high performance under comparable parameters. Table~\ref{tab:dgmn2_coco_object_detection_sparse_rcnn_val} lists the parameters and the bounding box AP of different backbones when using Sparse R-CNN. For example, our DGMN2-Small surpass ResNet-50~\cite{resnet} for about 3.2\% AP.

\subsection{Semantic segmentation}

\begin{table}[t]
\centering
\caption{\small Semantic segmentation performance using our DGMN2 on Cityscapes validation set. ``SS'' denotes single-scale testing and ``MS'' denotes multi-scale testing. ``21K'' means the backbone is pretrained on ImageNet-21K, otherwise on ImageNet-1K.}
\label{tab:dgmn2_cityscapes_val}
\scalebox{0.9}{
\tablestyle{1pt}{1}
\begin{tabular}{l|x{64}|x{36}|x{36}|x{56}}
 Backbone &\scriptsize Decoder Head &\scriptsize Params (M) &\scriptsize FLOPs (G) &\scriptsize mIoU (SS / MS) \\
\shline
\scriptsize DGMN2-Tiny &\scriptsize Semantic FPN~\cite{kirillov2019panoptic} &\scriptsize \textbf{15.84} &\scriptsize \textbf{76.35} &\scriptsize78.09 / 79.4 \\
\scriptsize DGMN2-Small &\scriptsize Semantic FPN~\cite{kirillov2019panoptic} &\scriptsize24.71 &\scriptsize 99.57 &\scriptsize80.65 / 81.58 \\
\scriptsize DGMN2-Medium &\scriptsize Semantic FPN~\cite{kirillov2019panoptic} &\scriptsize39.55 &\scriptsize 136.01 &\scriptsize80.60 / 81.79 \\
\scriptsize DGMN2-Large &\scriptsize Semantic FPN~\cite{kirillov2019panoptic} &\scriptsize52.02  &\scriptsize 175.94 &\scriptsize \textbf{81.75} / \textbf{82.64} \\
\hline
\scriptsize SETR (T-Base, 21K)~\cite{zheng2021rethinking} &\scriptsize Naive~\cite{zheng2021rethinking} &\scriptsize88.27 &\scriptsize 295.44 &\scriptsize75.54 / - \\
\scriptsize SETR (T-Large, 21K)~\cite{zheng2021rethinking} &\scriptsize Naive~\cite{zheng2021rethinking} &\scriptsize306.58 &\scriptsize 953.44 &\scriptsize77.37 / - \\
\scriptsize SETR (T-Base, DeiT)~\cite{zheng2021rethinking} &\scriptsize Naive~\cite{zheng2021rethinking} &\scriptsize88.27 &\scriptsize 295.44 &\scriptsize77.85 / - \\
\scriptsize DGMN2-Tiny &\scriptsize Naive~\cite{zheng2021rethinking} &\scriptsize \textbf{11.60} &\scriptsize \textbf{117.52} &\scriptsize77.23 / 78.23 \\
\scriptsize DGMN2-Small &\scriptsize Naive~\cite{zheng2021rethinking} &\scriptsize20.47 &\scriptsize 204.97 &\scriptsize80.31 / 81.04 \\
\scriptsize DGMN2-Medium &\scriptsize Naive~\cite{zheng2021rethinking} &\scriptsize35.31 &\scriptsize 347.76 &\scriptsize80.83 / 81.39 \\
\scriptsize DGMN2-Large &\scriptsize Naive~\cite{zheng2021rethinking} &\scriptsize47.78 &\scriptsize 467.46 &\scriptsize \textbf{81.80} / \textbf{82.61} \\
\hline
\scriptsize SETR (T-Base, 21K)~\cite{zheng2021rethinking} &\scriptsize PUP~\cite{zheng2021rethinking} &\scriptsize98.12 &\scriptsize 414.15 &\scriptsize76.71 / - \\
\scriptsize SETR (T-Large, 21K)~\cite{zheng2021rethinking} &\scriptsize PUP~\cite{zheng2021rethinking} &\scriptsize319.11 &\scriptsize 1079.18 &\scriptsize78.39 / - \\
\scriptsize SETR (T-Base, DeiT)~\cite{zheng2021rethinking} &\scriptsize PUP~\cite{zheng2021rethinking} &\scriptsize98.12 &\scriptsize 414.15 &\scriptsize78.79 / - \\
\scriptsize DGMN2-Tiny &\scriptsize PUP~\cite{zheng2021rethinking} &\scriptsize \textbf{13.95} &\scriptsize \textbf{242.51} &\scriptsize78.25 / 79.26 \\
\scriptsize DGMN2-Small &\scriptsize PUP~\cite{zheng2021rethinking} &\scriptsize22.82 &\scriptsize 329.96 &\scriptsize79.78 / 80.73 \\
\scriptsize DGMN2-Medium &\scriptsize PUP~\cite{zheng2021rethinking} &\scriptsize37.67 &\scriptsize 472.75 &\scriptsize80.97 / 81.80 \\
\scriptsize DGMN2-Large &\scriptsize PUP~\cite{zheng2021rethinking} &\scriptsize50.14 &\scriptsize 592.45 &\scriptsize \textbf{81.58} / \textbf{82.27} \\
\hline
\scriptsize SETR (T-Base, 21K)~\cite{zheng2021rethinking} &\scriptsize MLA~\cite{zheng2021rethinking} &\scriptsize92.50 &\scriptsize 309.45 &\scriptsize75.60 / - \\
\scriptsize SETR (T-Large, 21K)~\cite{zheng2021rethinking} &\scriptsize MLA~\cite{zheng2021rethinking} &\scriptsize310.81 &\scriptsize 973.68 &\scriptsize76.65 / - \\
\scriptsize SETR (T-Base, DeiT)~\cite{zheng2021rethinking} &\scriptsize MLA~\cite{zheng2021rethinking} &\scriptsize92.50 &\scriptsize 309.45 &\scriptsize78.04 / - \\
\scriptsize DGMN2-Tiny &\scriptsize MLA~\cite{zheng2021rethinking} &\scriptsize \textbf{13.37} &\scriptsize \textbf{141.72} &\scriptsize78.25 / 79.32 \\
\scriptsize DGMN2-Small &\scriptsize MLA~\cite{zheng2021rethinking} &\scriptsize22.24 &\scriptsize 229.17 &\scriptsize80.79 / 81.62 \\
\scriptsize DGMN2-Medium &\scriptsize MLA~\cite{zheng2021rethinking} &\scriptsize37.08 &\scriptsize 371.96 &\scriptsize81.09 / \textbf{82.00} \\
\scriptsize DGMN2-Large &\scriptsize MLA~\cite{zheng2021rethinking} &\scriptsize49.56 &\scriptsize 491.66 &\scriptsize \textbf{81.55} / 81.98 \\
\end{tabular}
}
\end{table}

\par\noindent\textbf{Implementation details.} 
We use our DGMN2 as the backbone to perform the semantic segmentation experiments on the Cityscapes dataset~\cite{Cityscapes}. 
In all of our experiments on semantic segmentation, we use ImageNet pretrained weights (from Table~\ref{tab:dgmn2_imagenet_val}) to initialize the backbone and train all models on the training set and evaluated on the validation set and test set.
For the decoder head, it is randomly initialized.
Following the default setting (\eg, data augmentation and training schedule) of public codebase mmsegmentation~\cite{mmseg2020}, 
we set the number of training iterations to 40,000 and the input size to 769 $\times$ 769 for all our experiments.
We use AdamW~\cite{loshchilov2017decoupled} optimizer and a batch size of 8 to train the model with the initial learning rate of $1 \times 10^{-4}$. 
For fair comparison with the Transformer-based semantic segmentation model SETR~\cite{zheng2021rethinking}, we also use the decoder heads of SETR-Naive, SETR-PUP and SETR-MLA. 
As the DGMN2 is a pyramidal architecture and
in order to adapt our model to the semantic segmentation task with  higher resolution outputs, we set the stride of convolution in the patch embedding of the third and fourth stages to 1. 
Note that we can still can use the ImageNet pretrained weights to initialize the backbone, as we are simply changing the strides, but not any of the parameters.
For all experiments on semantic segmentation, we do not use the auxiliary loss or complementary tricks like OHEM~\cite{ohem} in model training.

For the evaluation on test set, we submit the result to Cityscapes test server. 
We use the same training schedule as SETR~\cite{zheng2021rethinking} to train the model: 
(1) We first train the model on Cityscapes training and validation set for 100,000 iterations with the initial learning rate of $1 \times 10^{-4}$; 
(2) We then fine tune the model on Cityscapes coarse set for 50,000 iterations with the initial learning rate of $1 \times 10^{-5}$; 
(3) We finally fine tune the model on Cityscapes training and validation set for 20,000 iterations with the initial learning rate of $1 \times 10^{-5}$.

\par\noindent\textbf{Results on Cityscapes.}
Table~\ref{tab:dgmn2_cityscapes_val} and~\ref{tab:dgmn2_cityscapes_test} 
show the comparative results on the validation and test set of Cityscapes respectively.
We can make following observations:
\textbf{(i)}
SETR with backbone of DeiT~\cite{touvron2020training} with ImageNet-1K pretraining even outperforms that of ViT~\cite{dosovitskiy2020image} with ImageNet-21K pretraining and T-large configuration;
\textbf{(ii)}
Our model consistently outperforms SETR~\cite{zheng2021rethinking} across all of the decoder heads (\eg,~Naive, PUP, MLA), scales (\eg,~T-Base, T-Large) and its strong version DeiT~\cite{touvron2020training}, with fewer parameters and FLOPs.
Note that our DGMN2 is pretrained on ImageNet-1K, much less data training than the SETR ImageNet-21K configuration.
This is again showing the effectiveness of our dynamic graph message passing module is superior to the fully connected graph in dense prediction task.
\textbf{(iii)}
Particularly, with the advantage of multi-scale pyramidal architecture design, DGMN2 achieves superior results with decoder head of Semantic FPN~\cite{kirillov2019panoptic}.
\textbf{(iv)}
Our DGMN2-Large achieves mIoU 82.46\% on test set with the MLA~\cite{zheng2021rethinking} decoder head.
which is on par with the best results reported so far.

\begin{table}[t]
\centering
\caption{\small Semantic segmentation performance using our DGMN2 on Cityscapes test set.}
\label{tab:dgmn2_cityscapes_test}
\scalebox{1}{
\tablestyle{1pt}{1}
\begin{tabular}{l|x{60}|x{36}|x{36}|x{36}}
 Backbone &\scriptsize Decoder Head &\scriptsize Params (M) &\scriptsize FLOPs (G) &\scriptsize mIoU \\
\shline
\scriptsize SETR (T-Large)~\cite{zheng2021rethinking} &\scriptsize PUP~\cite{zheng2021rethinking} &\scriptsize319.11 &\scriptsize 1079.18 &\scriptsize81.64 \\
\hline
\scriptsize DGMN2-Large &\scriptsize Semantic FPN~\cite{kirillov2019panoptic} &\scriptsize52.02 &\scriptsize 175.94 &\scriptsize81.84 \\
\scriptsize DGMN2-Large &\scriptsize Naive~\cite{zheng2021rethinking} &\scriptsize \textbf{47.78} &\scriptsize 467.46 &\scriptsize81.33 \\
\scriptsize DGMN2-Large &\scriptsize PUP~\cite{zheng2021rethinking} &\scriptsize50.14 &\scriptsize 592.45 &\scriptsize81.86 \\
\scriptsize DGMN2-Large &\scriptsize MLA~\cite{zheng2021rethinking} &\scriptsize49.56 &\scriptsize 491.66 &\scriptsize \textbf{82.46} \\
\end{tabular}
}
\end{table}

\subsection{Visualization}

\begin{figure*}[h]
	\centering
	\hspace{-12mm}
	\subfloat{
	\includegraphics[width=0.3\linewidth]{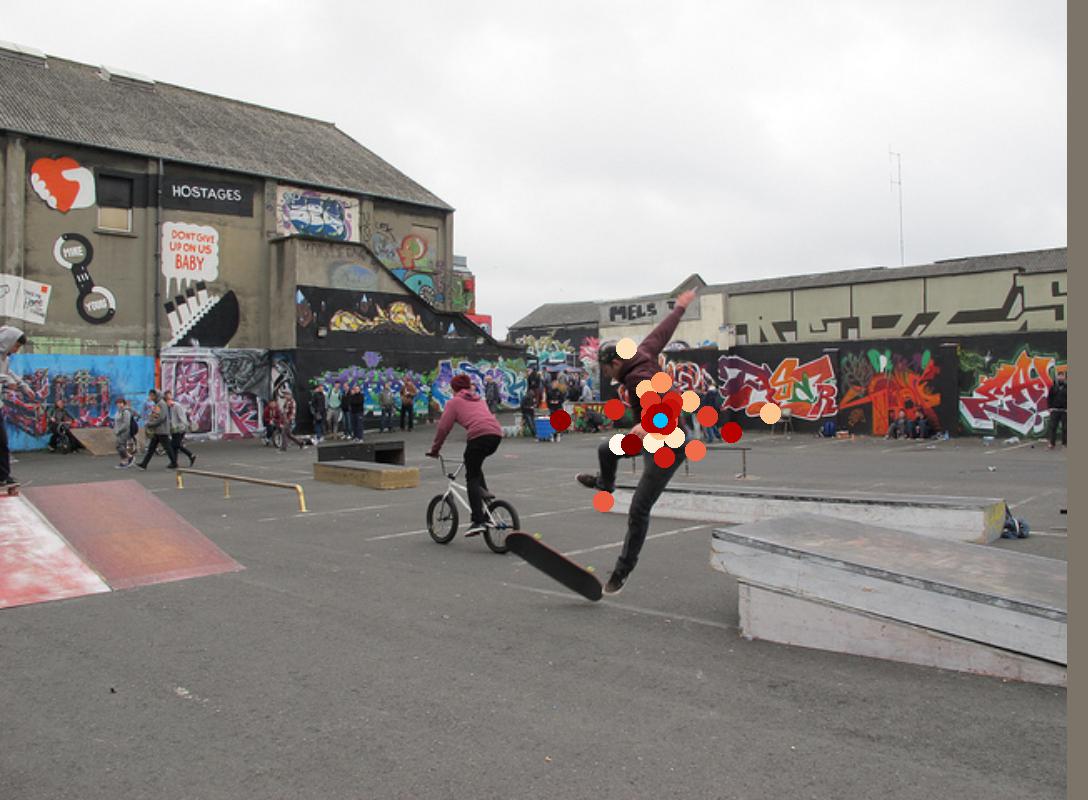}
	}
	\subfloat{
	\includegraphics[width=0.3\linewidth]{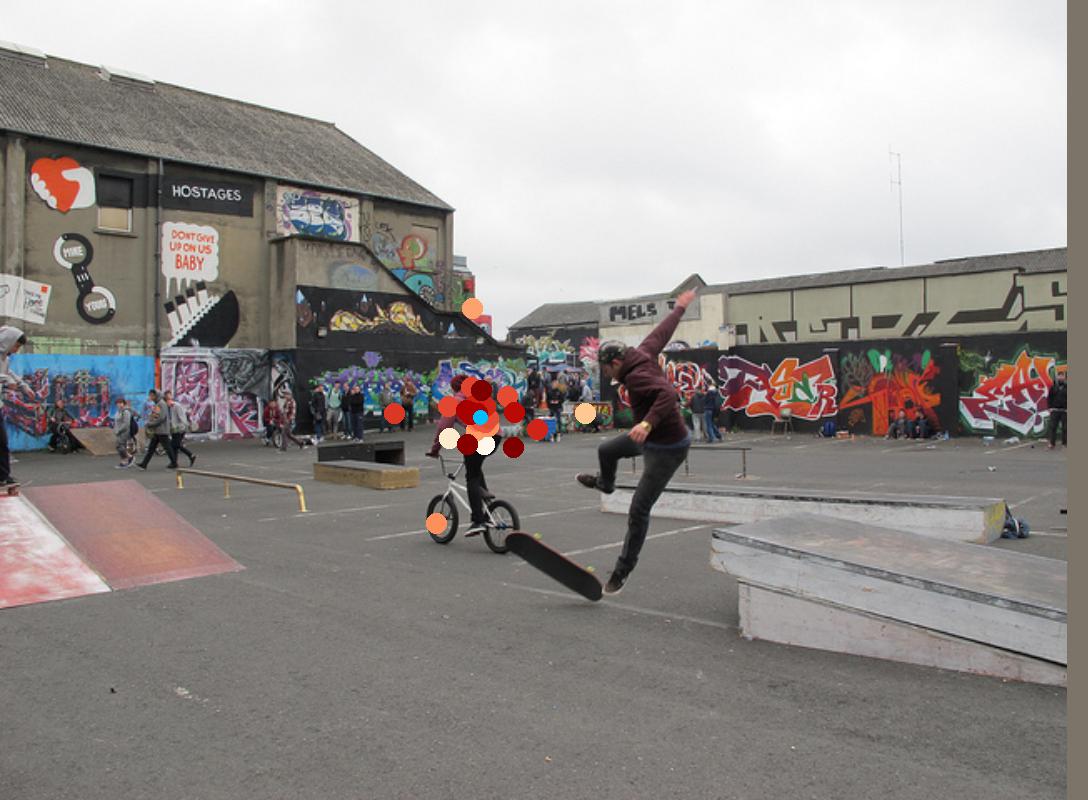}
	}
	\subfloat{
	\includegraphics[width=0.3\linewidth]{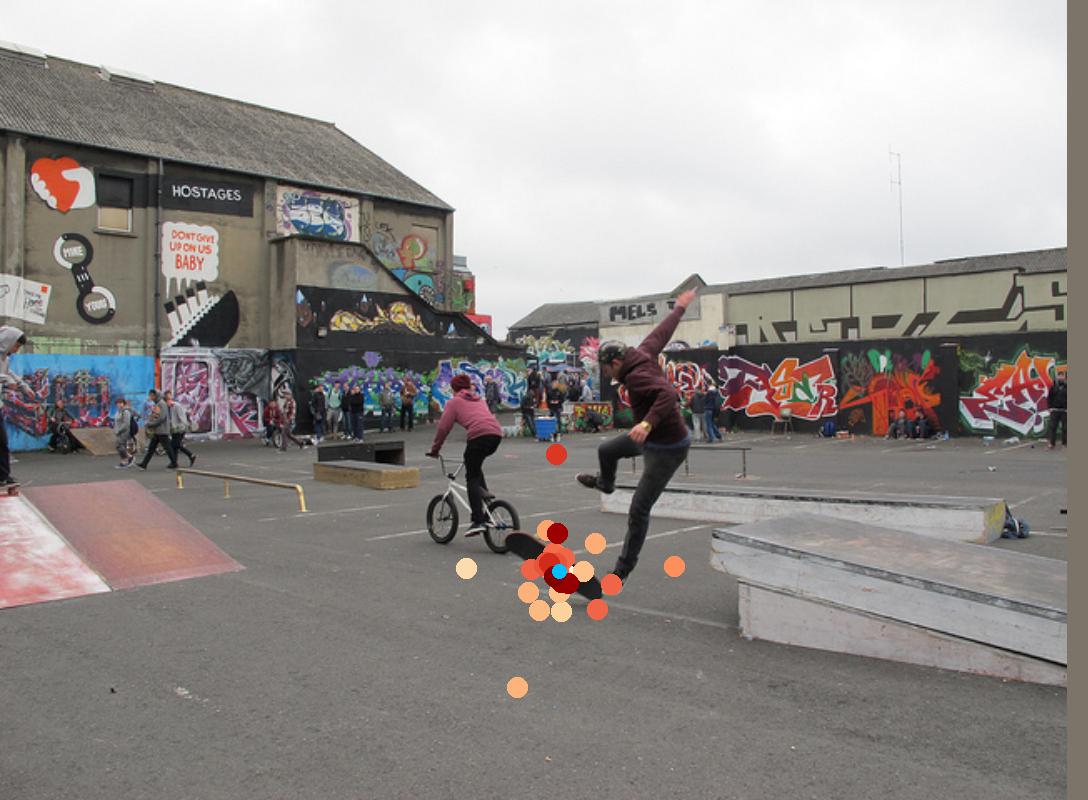}
	} \\
	\hspace{-12mm}
	\subfloat{
	\includegraphics[width=0.3\linewidth]{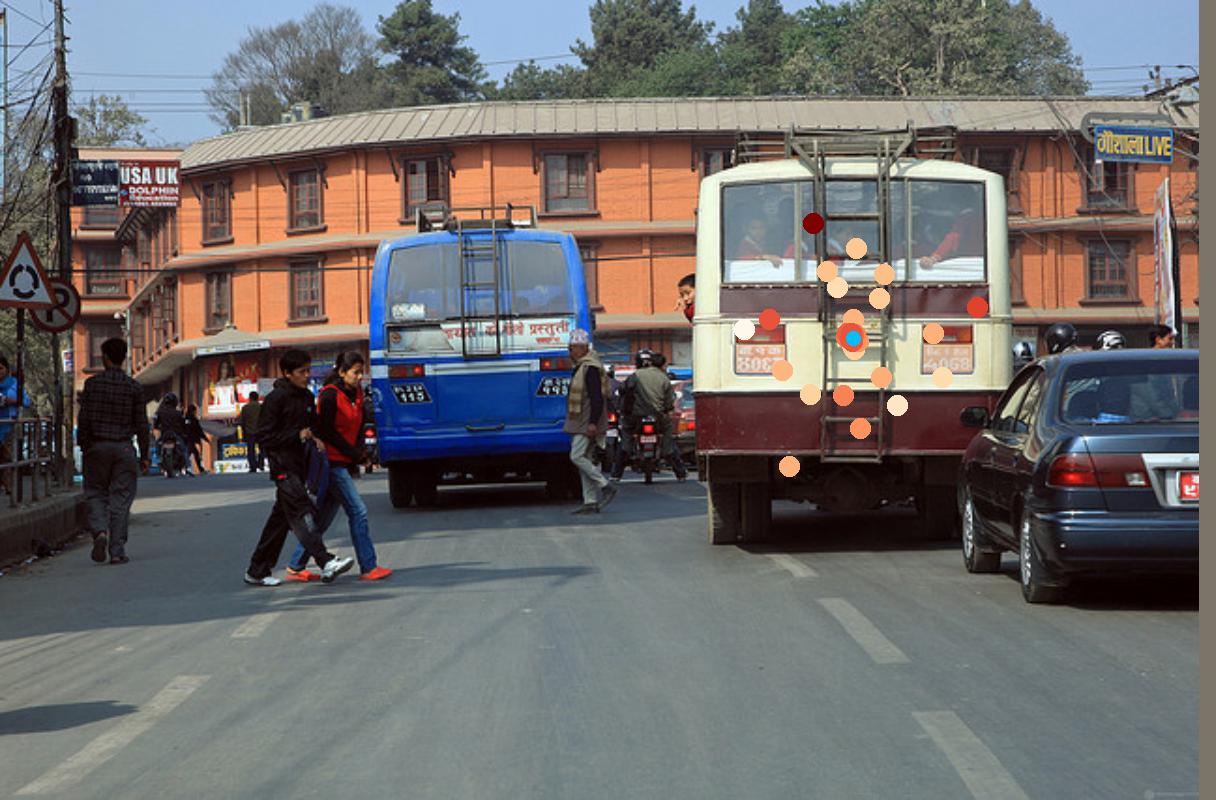}
	}
	\subfloat{
	\includegraphics[width=0.3\linewidth]{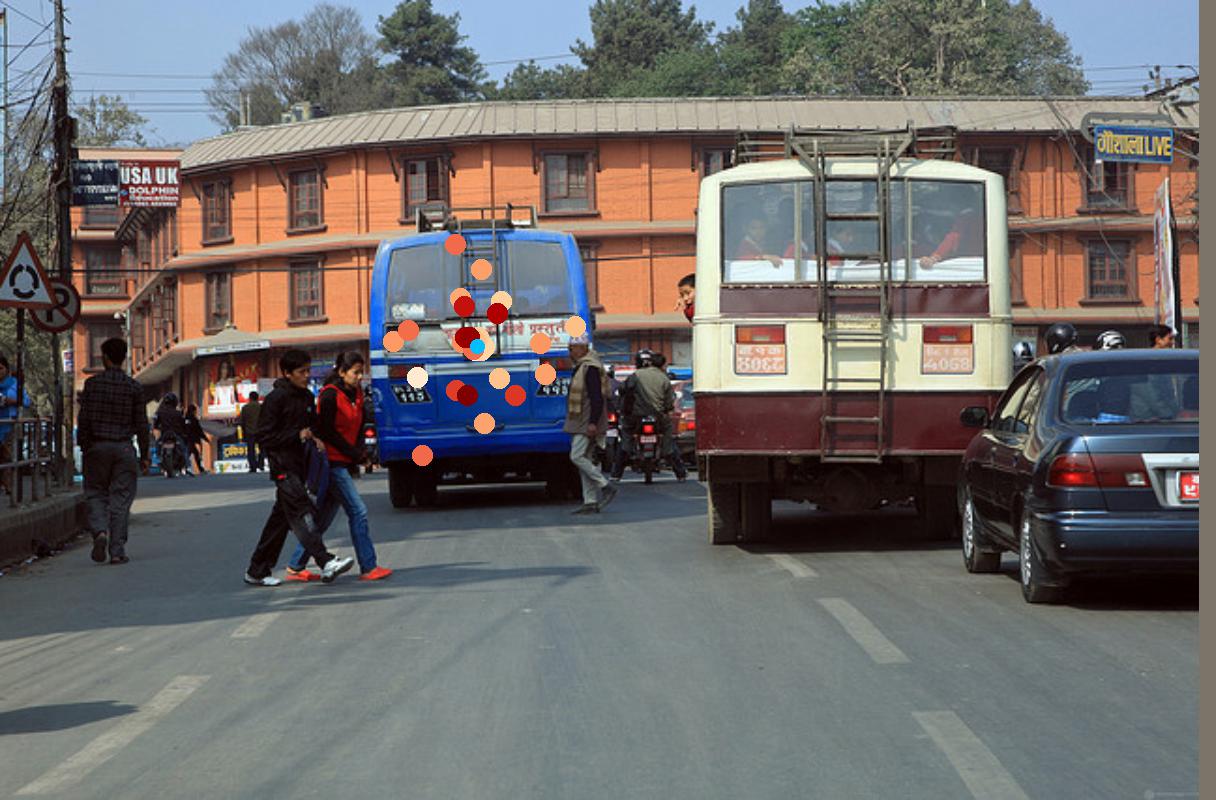}
	}
	\subfloat{
	\includegraphics[width=0.3\linewidth]{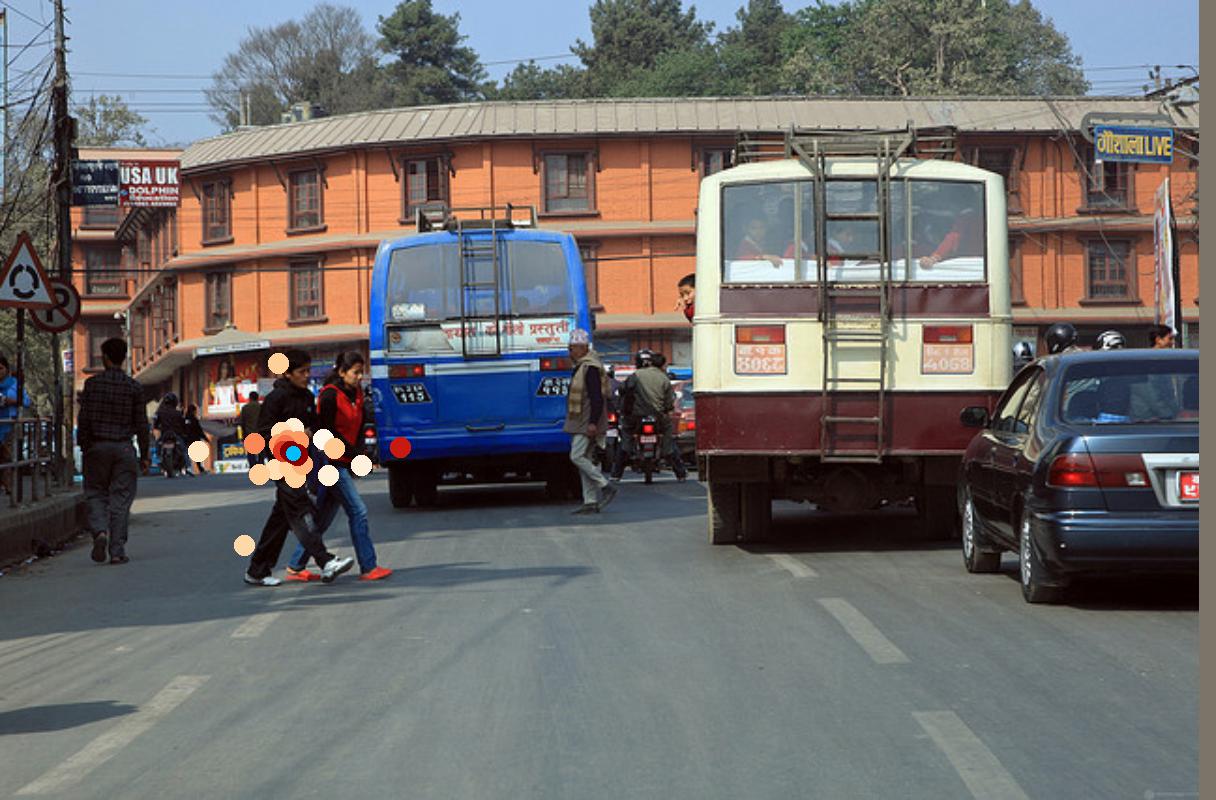}
	} \\
	\subfloat{
	\includegraphics[width=0.3\linewidth]{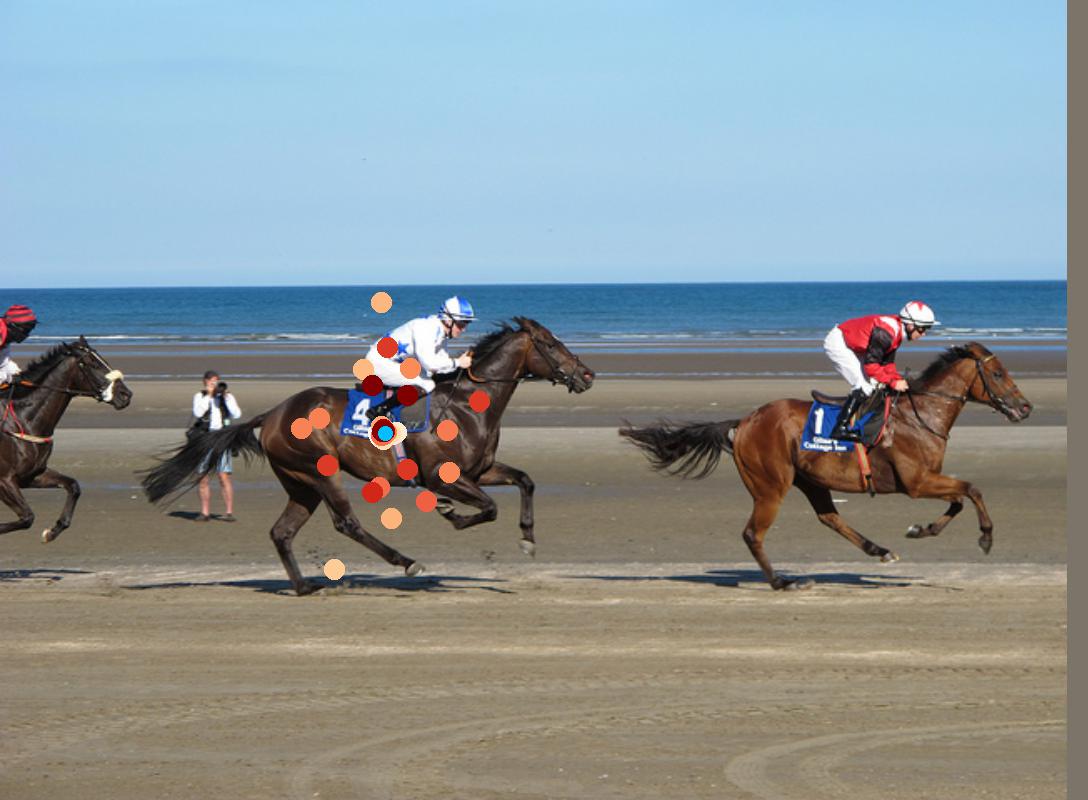}
	}
	\subfloat{
	\includegraphics[width=0.3\linewidth]{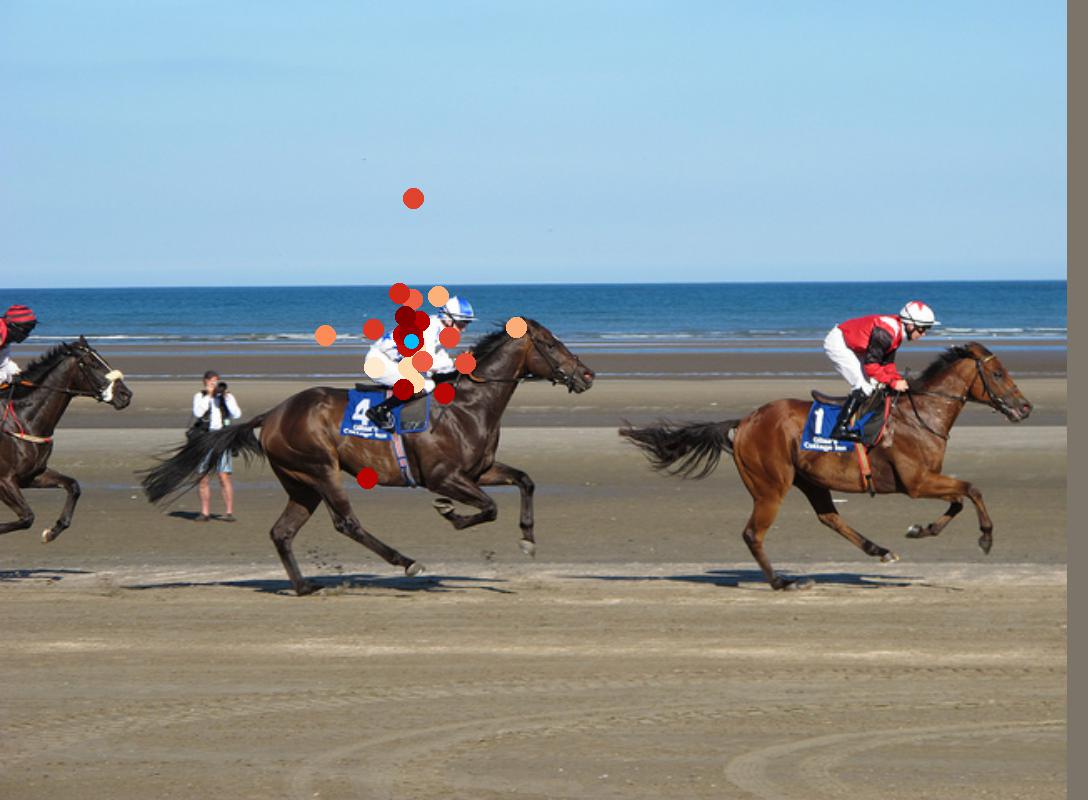}
	}
	\subfloat{
	\includegraphics[width=0.3\linewidth]{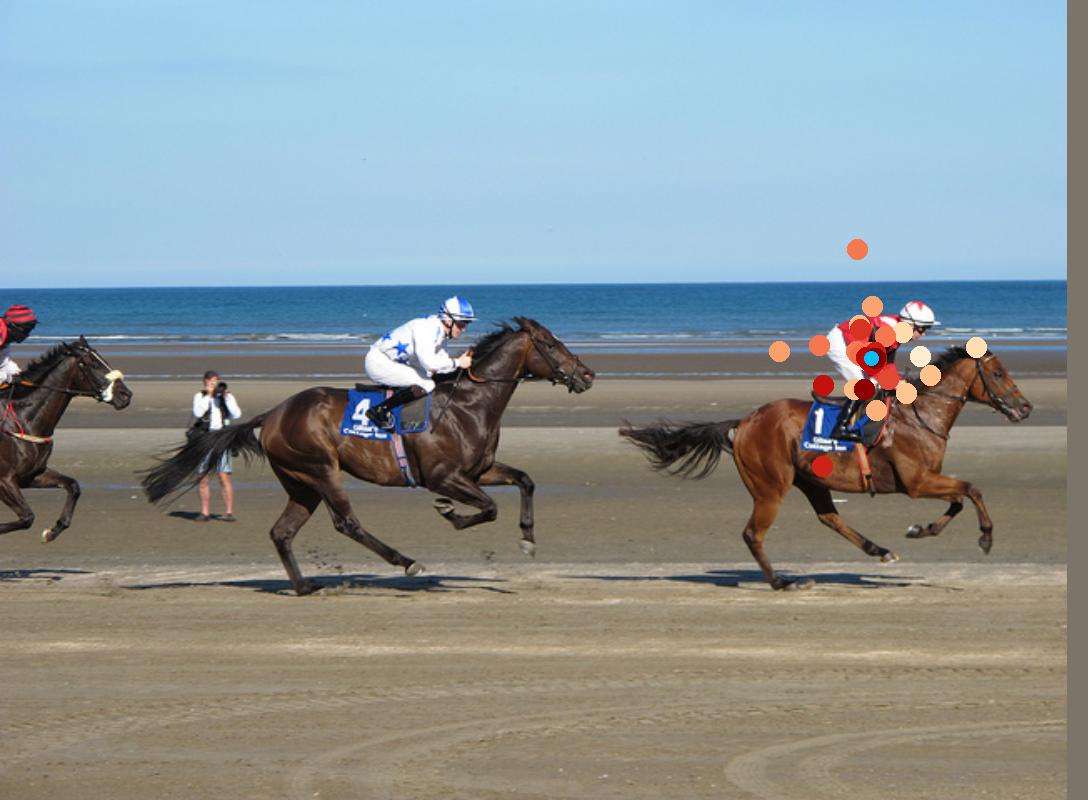}
	}
	\subfloat{
	\includegraphics[width=0.05\linewidth]{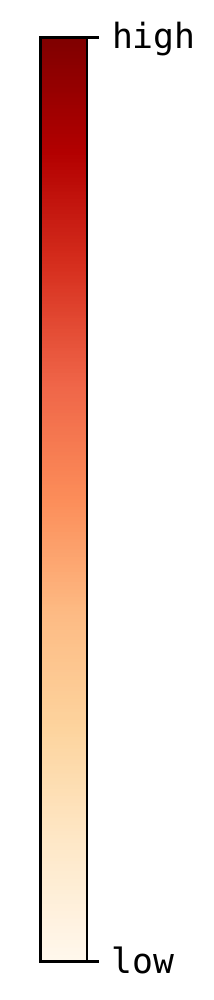}
	} \\
	\hspace{-12mm}
	\subfloat{
	\includegraphics[width=0.3\linewidth]{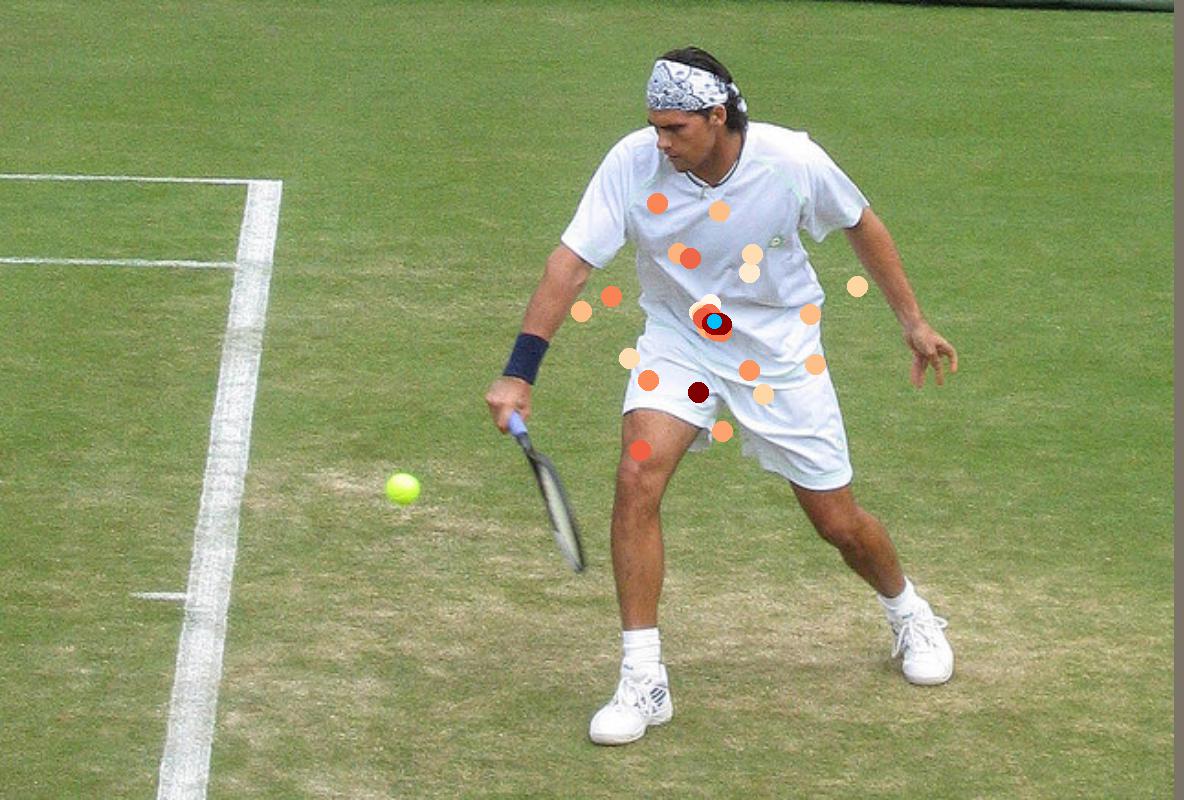}
	}
	\subfloat{
	\includegraphics[width=0.3\linewidth]{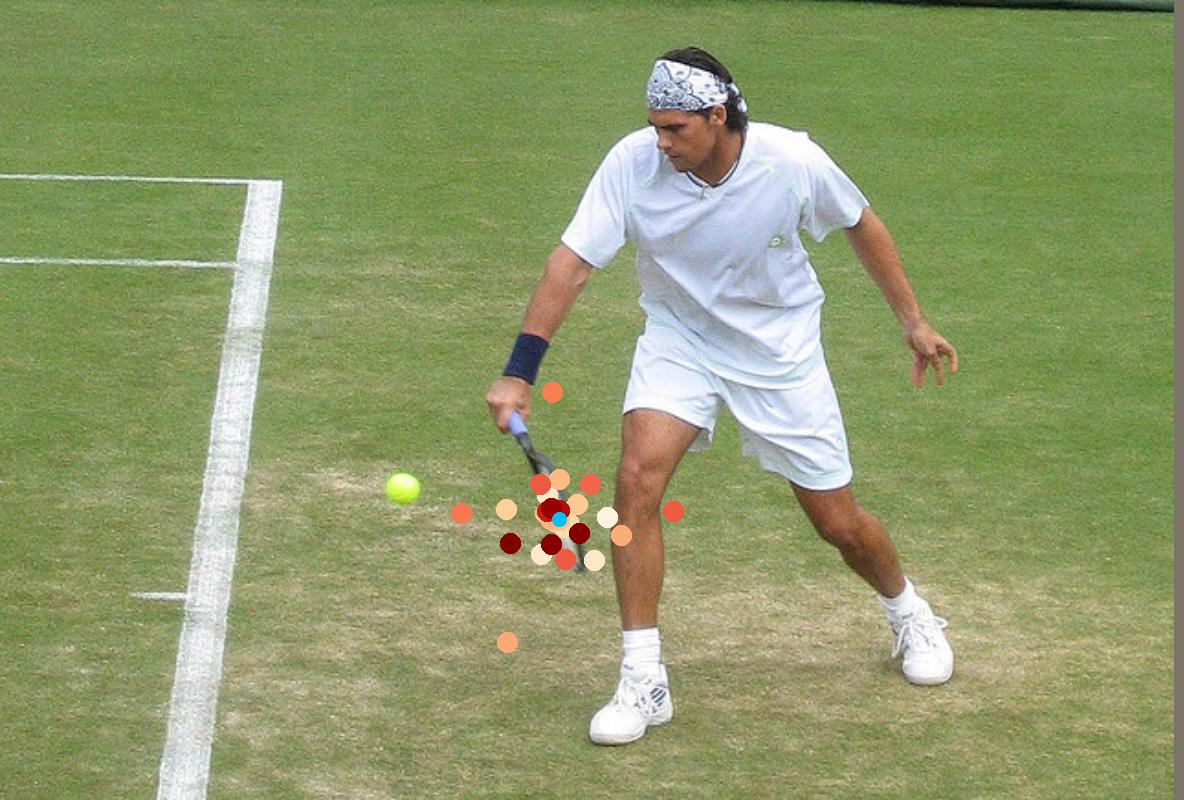}
	}
	\subfloat{
	\includegraphics[width=0.3\linewidth]{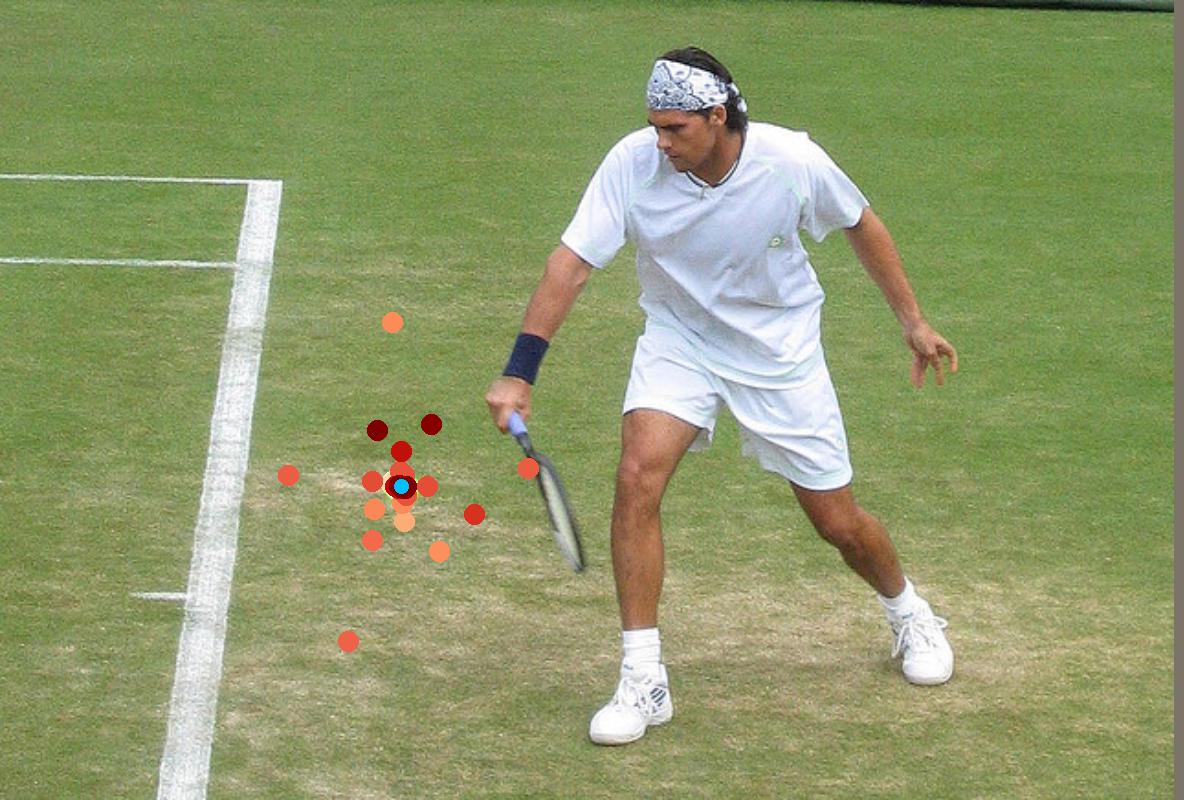}
	} \\
	\hspace{-12mm}
	\subfloat{
	\includegraphics[width=0.3\linewidth]{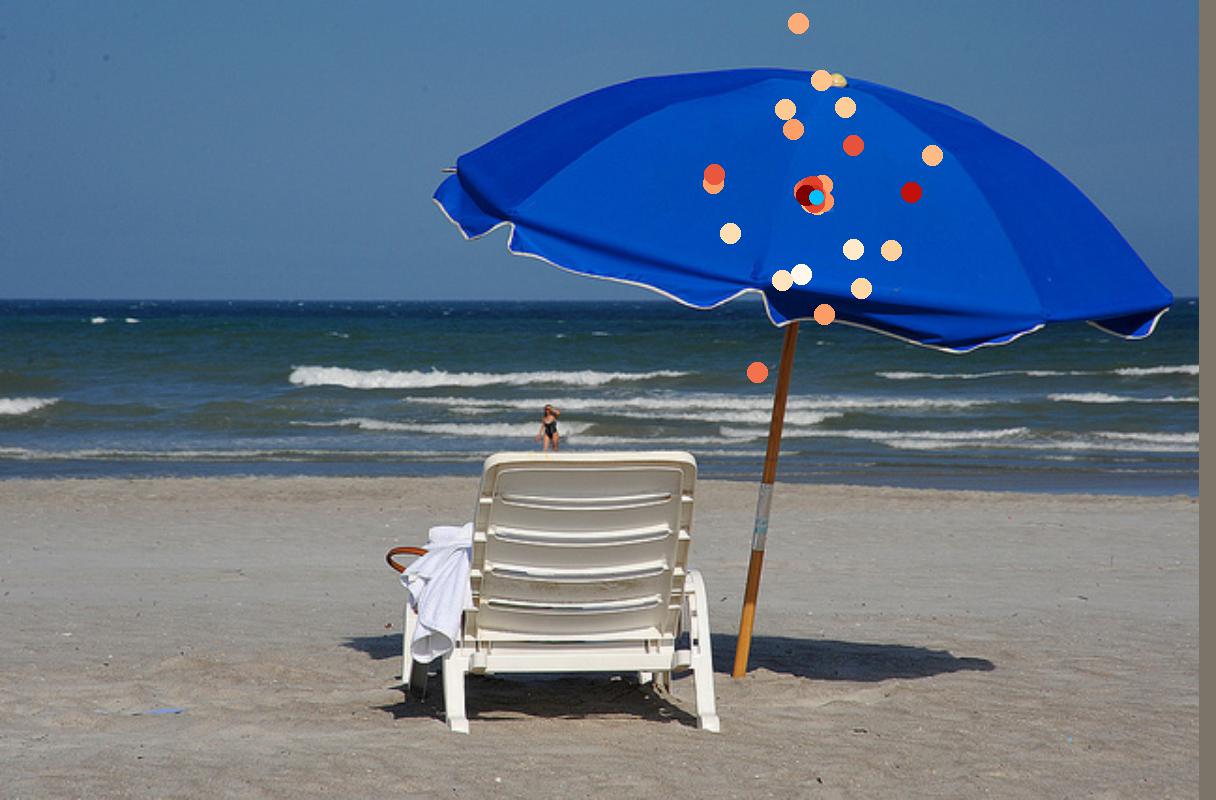}
	}
	\subfloat{
	\includegraphics[width=0.3\linewidth]{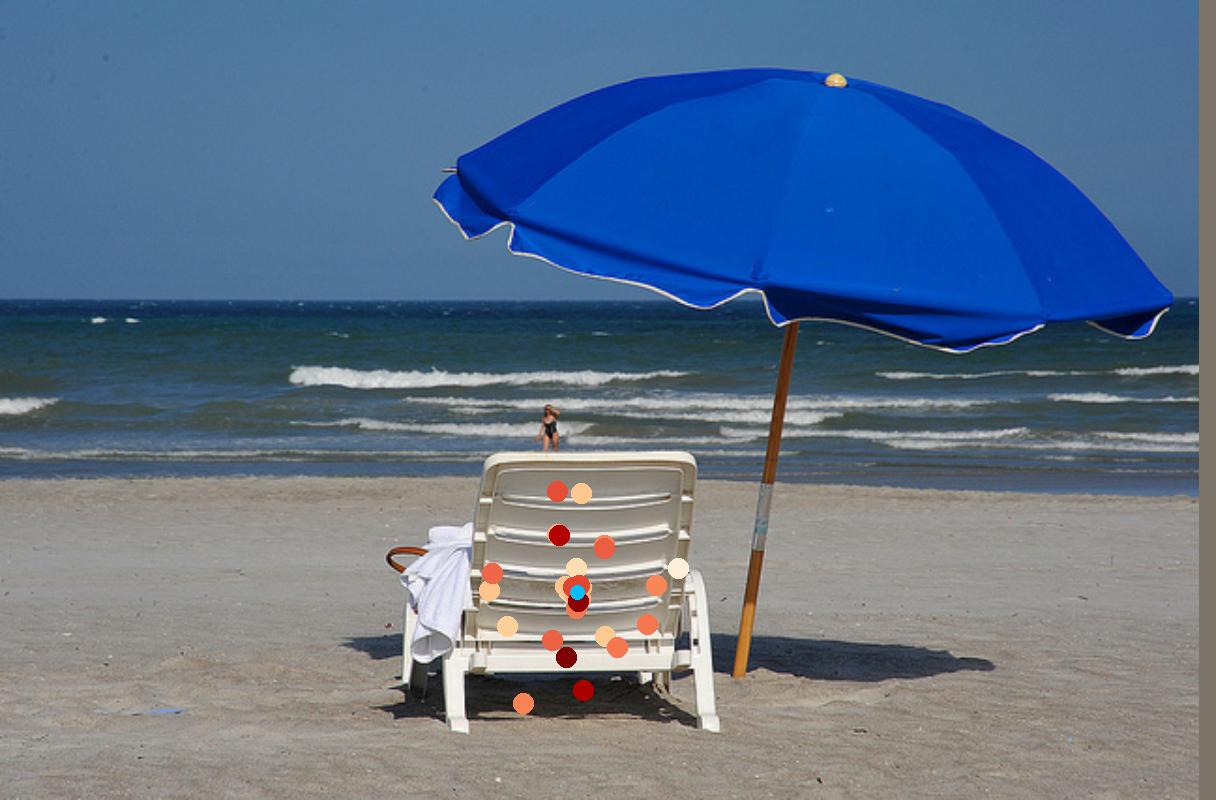}
	}
	\subfloat{
	\includegraphics[width=0.3\linewidth]{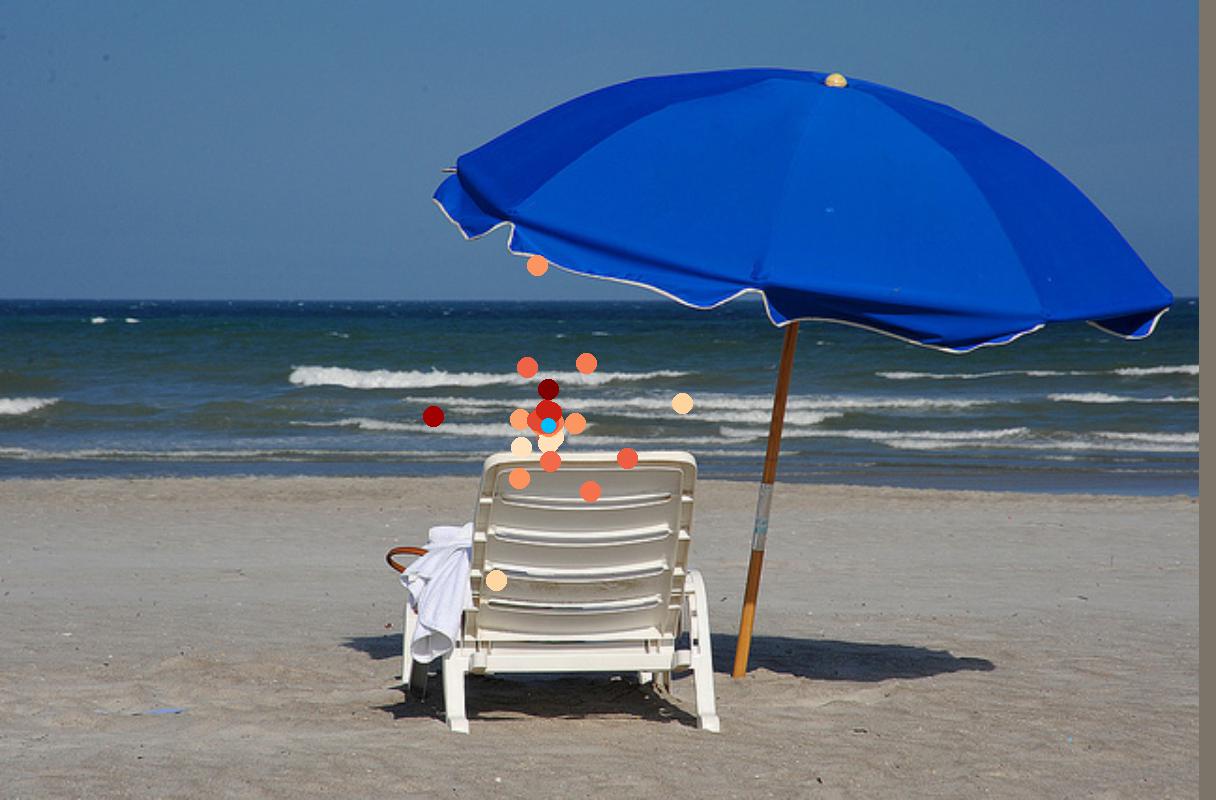}
	}
	\caption{
	\small
	Visualization of dynamic sampled graph nodes with our DGMN2 (trained for object detection on COCO dataset). The blue circle denotes a selected position on observed feature and the orange circles denote the corresponding sampled graph nodes. 
	The color shades of the orange circles indicate the intensity of the dynamic filters of the sampling graph nodes.
	The color bar is shown at the \textit{right}.
	}
	\label{fig:dgmn2_sampled_node_visualization_coco}
\end{figure*}

We visualize the dynamic sampled graph nodes using the DGMN2 model trained for object detection on COCO dataset. 
We select a point on image and find the corresponding position on the observed feature and visualize their corresponding sampled graph nodes. 
For better understanding, we draw all sampled graph nodes from the last layer of each of the four stages in our DGMN2.
As shown in Fig.~\ref{fig:dgmn2_sampled_node_visualization_coco}, the blue circles denotes a selected position on observed feature and the orange circles denote the corresponding sampled graph nodes. 
The color shades of the orange circles indicates the intensity of the dynamic filters on the sampling graph nodes. 
This visualization demonstrates that our proposed DGMN2 is able to adaptively sample nodes and use dynamic filters with appropriate intensity at different positions according to context feature of different objects.

\section{Conclusion}
\label{sec:conclusion}

We proposed Dynamic Graph Message passing Networks (DGMN), a novel graph neural network module that dynamically determines the graph structure for each input.
It learns dynamic sampling of a small set of relevant neighbours for each node, and also predicts the weights and affinities dependant on the feature nodes to propagate information through this sampled neighbourhood.
This formulation significantly reduces the computational cost of static, fully-connected graphs
which contain many redundancies.
We further recast our formulation to take a Transformer perspective and instantiate an efficient and effective Transformer layer and present a new Transformer backbone network DGMN2 in supporting various vision downstream tasks.
This is demonstrated by the fact that we are able to clearly improve upon the accuracy of several state-of-art baselines, on four complex scene understanding tasks.



%

\ifCLASSOPTIONcompsoc
  \section*{Acknowledgments}
\else
  \section*{Acknowledgment}
\fi

We thank Professor Dan Xu and Professor Andrew Zisserman for valuable discussions.
This work was supported in part by 
National Natural Science Foundation of China (Grant No. 6210020439),
Lingang Laboratory (Grant No. LG-QS-202202-07) and
Natural Science Foundation of Shanghai (Grant No. 22ZR1407500).

\ifCLASSOPTIONcaptionsoff
  \newpage
\fi



%

{\small
\bibliographystyle{ieee_fullname}
\bibliography{biblio}
}





\section{Additional experiments}
\label{sec:appendix}

In this supplementary material, we report additional details about the experiments in our paper (Sec.~\ref{sec:app_exp_details}), and also conduct further ablation studies (Sec.~\ref{sec:app_ablation}).

\subsection{Additional experimental details}
\label{sec:app_exp_details}

\subsubsection{Semantic segmentation on Cityscapes}
\label{sec:app_implementation_cityscapes}

For the semantic segmentation task on Cityscapes, we follow \cite{pspnet} and use a polynomial learning rate decay with an initial learning rate of 0.01.
The momentum and the weight decay are set to 0.9 and 0.0001 respectively. 
We use 4 Nvidia V100 GPUs, batch size 8 and train for $40 000$ iterations from an ImageNet-pretrained model.
For data augmentation, random cropping with a crop size of 769 and random mirror-flipping are applied on-the-fly during training. 
Note that following common practice \cite{pspnet,ocnet,psanet,rota2018place} we used synchronised batch normalisation for better estimation of the batch statistics for the experiments on Cityscapes.
When predicting dynamic filter weights, we use the grouping parameter $G=4$. 
For the experiments on Cityscapes, we use a set of the sampling rates of $\varphi = \{1, 6, 12, 24, 36\}$. 

\subsubsection{Object detection and instance segmentation on COCO}
\label{sec:app_implementation_coco}

Our models and all baselines are trained with the typical ``$1\mathrm{x}$'' training settings from the public Mask R-CNN benchmark~\cite{massa2018mrcnn} for all experiments on COCO.
More specificially, 
the backbone parameters of all the models in the experiments are pretrained on ImageNet classification. 
The input images are resized such that their shorter side is of 800 pixels and the longer side is limited to 1333 pixels.
The batch size is set to 16. 
The initial learning rate is set to 0.02 with a decrease by a factor of 0.1 after $60 000$ and $80 000$ iterations, and finally terminates at $90 000$ iterations. 
Following~\cite{massa2018mrcnn, goyal2017accurate}, the training warm-up is employed by using a smaller learning rate of 0.02 $\times$ 0.3 for the first 500 iterations of training. 
All the batch normalisation layers in the backbone are ``frozen'' during fine-tuning on COCO. 

When predicting dynamic filter weights, we use the grouping parameter $G=4$. 
For the experiments on COCO, a set of the sampling rates of $\varphi = \{1, 4, 8, 12\}$ is considered.
We train models on only the COCO training set and test on the validation set and test-dev set.

\subsection{Additional ablation studies}
\label{sec:app_ablation}

\par\noindent\textbf{Effectiveness of different training and inference strategies.} 
When evaluating models for the Cityscapes test set, we followed common practice and employed several complementary strategies used to improve performance in semantic segmentation, including Online Hard Example Mining (OHEM)~\cite{ohem,pohlen2017full,li2017holistic,ocnet,bisenet}, Multi-Grid~\cite{deeplabv3,DAnet,graph_reason} and Multi-Scale (MS) ensembling~\cite{deeplabv1,deeplabv3p,pspnet,ocnet,boxsup}. 
The contribution of each strategy is reported in Table~\ref{tab:improve} on the validation set. 

\par\noindent\textbf{Inference time}
We tested the average run time on the Cityscapes validation set with a Nvidia Tesla V100 GPU. 
The Dilated FCN baseline and the Non-local model take 0.230s and 0.276s per image, respectively, while our proposed model uses 0.253s,
Thus, our proposed method is more efficient than Non-local~\cite{wang2018nonlocal} in execution time, FLOPs and also the number of parameters.

\par\noindent\textbf{Effectiveness of different sampling rate $\varphi$ and group of predicted weights $G$ (Section 3.3 and 3.4 in main paper)}.
For our experiments on Cityscapes, where are network has a stride of 8, the sampling rates are set to  $\varphi = \{1, 6, 12, 24, 36\}$.
For experiments on COCO, where the network stride is 32, we use smaller sampling rates of $\varphi = \{1, 4, 8, 12\}$ in C5.
We keep the same sampling rate in C4 when DGMN modules are inserted into C4 as well.

Unless otherwise stated, all the experiments in the main paper and supplementary used $G=4$ groups as the default.
Each group of $C/G$ feature channels shares the same set of filter parameters~\cite{chollet2017xception}.

The effect of different sampling rates and groups of predicted filter weights are studied in Table~\ref{tab:segdilation}, for semantic segmentation on Cityscapes, and Table~\ref{tab:detdilation}, for object detection and instance segmentation on COCO.

\begin{table}[t]
	\tablestyle{2pt}{1.1}
	\centering
	\caption{\small
	Ablation studies of different training and inference strategies.
    Our method (DGMN w/ DA+DW+US) is evaluated under single scale mIoU with ResNet-101 backbone on Cityscapes validation set.
    }\label{tab:improve}
	\begin{tabular}{   l  |x{42}|x{52}|x{32}|x{42}}
		& OHEM & Multi-grid &MS &mIoU (\%)   \\
		\shline
		FCN w/ DGMN   &\xmark  &\xmark &\xmark & 79.2 \\
        FCN w/ DGMN & \cmark   & \xmark &\xmark  &79.7   \\
        FCN w/ DGMN &\cmark   &\cmark  & \xmark  &80.3   \\
        FCN w/ DGMN &\cmark  &\cmark & \cmark   &\bf 81.1   \\
	\end{tabular}
\end{table}

\begin{table*}[t]
\tablestyle{2pt}{1.2}
	\centering
	\begin{tabular}{   l  |x{32}|x{32}|x{32}|x{32}}
     &DA &DW& DS & mIoU (\%)   \\
    \shline
    Dilated FCN    &\xmark &\xmark &\xmark & 75.0 \\
    \hline
    + DGMN ($\varphi =\{1\}$)  & \cmark &\xmark &\xmark  &  76.5 \\
    + DGMN ($\varphi =\{1\}$)  & \cmark & \cmark  &\xmark &  79.1 \\
    + DGMN ($\varphi = \{1, 1, 1, 1\}$)  & \cmark & \cmark &\cmark & 79.2 \\
    + DGMN ($\varphi = \{1, 6, 12\}$)  & \cmark & \cmark & \cmark  &  79.7 \\
    + DGMN ($\varphi = \{1, 6, 12, 24, 36\}$)  & \cmark & \cmark &\cmark  &80.4  \\
    \end{tabular}
\caption{\small 
Quantitative analysis on different sampling rates of our dynamic sampling strategy in the proposed DGMN model on the Cityscapes validation set. 
We report the mean IoU and use a ResNet-101 as backbone. 
All methods are evaluated using a single scale.}
\label{tab:segdilation}
\end{table*}


\begin{table*}[h!]
\tablestyle{2pt}{1.2}
	\centering
    \begin{tabular}{   l  |x{32}|x{32}|x{32}|x{32}|x{32}}
     &DA &DW & DS & $\mathrm{AP^{box}}$ & $\mathrm{AP^{mask}}$  \\
    \shline
    Mask R-CNN baseline   &\xmark &\xmark &\xmark  & 37.8 & 34.4\\
    \hline
    + DGMN ($\varphi = \{1, 4, 8, 12\}$, $G = 0$)  & \cmark & \xmark &\xmark &39.4 &35.6 \\
    + DGMN ($\varphi = \{1, 4, 8, 12\}$, $G = 0$)  & \cmark & \xmark &\cmark &39.9 &35.9 \\
    + DGMN ($\varphi = \{1, 4, 8\}$, $G = 2$)  & \cmark & \cmark  & \cmark &39.5 &35.6 \\
    + DGMN ($\varphi = \{1, 4, 8\}$, $G = 4$)  & \cmark & \cmark  & \cmark &39.8 &35.9  \\
    + DGMN ($\varphi = \{1, 4, 8, 12\}$, $G = 4$)  & \cmark & \cmark & \cmark  &\bf 40.2  &\bf 36.0 \\
    \end{tabular}
\caption{\small
Quantitative analysis on different numbers of filter groups ($G$) and sampling rates ($\varphi$) for the proposed DGMN model on the COCO 2017 validation set. 
All methods are based on the Mask R-CNN detection pipeline with a ResNet-50 backbone, and evaluated on the COCO validation set.
Modules are inserted after all the $3 \times 3$ convolution layers of C5 (\emph{res5}) of ResNet-50.}
\label{tab:detdilation}
\end{table*}

\par\noindent\textbf{Effectiveness of feature learning with DGMN on stronger backbones.}
Table 4 of the main paper showed that our proposed DGMN module still provided substantial benefits on the more powerful backbones such as ResNet-101 and ResNeXt 101 on the COCO test set.
Table~\ref{tab:strongbackbone} shows this for the COCO validation set as well.
By inserting DGMN at the convolutional stage C5 of ResNet-101, DGMN (C5) outperforms the Mask R-CNN baseline with 1.6 points on the  AP$^{\mathrm{box}}$ metricand by 1.2 points on the AP$^{\mathrm{mask}}$ metric. 
On ResNeXt-101, DGMN (C5) also improves by 1.5 and 0.9 points on the AP$^{\mathrm{box}}$ and the AP$^{\mathrm{mask}}$, respectively.

\begin{table*}[h!]
\tablestyle{2pt}{1.2}
	\centering
	\begin{tabular}{   l  |x{52}|x{32}x{32}x{32}|x{32}x{32}x{32}}
Model & Backbone & $\mathrm{AP^{box}}$ & $\mathrm{AP^{box}_{50}}$ & $\mathrm{AP^{box}_{75}}$ &$\mathrm{AP^{mask}}$ &$\mathrm{AP^{mask}_{50}}$ &$\mathrm{AP^{mask}_{75}}$  \\
\shline
Mask R-CNN baseline & \multirow{2}{*}{ResNet-101} &40.1 &61.7 &44.0 &36.2  &58.1 &38.3 \\
+ DGMN (C5) &  &\bf 41.7 &\bf 63.8 &\bf 45.7 &\bf 37.4  &\bf 60.4 &\bf 39.8 \\
\hline 
Mask R-CNN baseline & \multirow{2}{*}{ResNeXt-101}   &42.2 &63.9 &46.1 &37.8  &60.5 &40.2 \\
+ DGMN (C5)  &  &\bf 43.7 &\bf 65.9 &\bf 47.8 &\bf 38.7  &\bf 62.1 &\bf 41.3 \\
\end{tabular}
\caption{\small
Quantitative results via applying the proposed DGMN module into different strong backbone networks for object detection and instance segmentation on the COCO 2017 validation set.}
\label{tab:strongbackbone}
\end{table*} 

\subsection{State-of-the-art comparison on COCO}
\label{sec:coco_sota}
Table~\ref{tab:coco_sota} shows comparisons to the state-of-the-art on the COCO \texttt{test-dev} set.
When testing, we process a single scale using a single model.
We do not perform any other complementary performance-boosting ``tricks''.
Our DGMN approach outperforms one-stage detectors including the most recent CornerNet~\cite{law2018cornernet} by 2.1 points on box Average Precision (AP). 
DGMN also shows superior performance compared to two-stage detectors including Mask R-CNN~\cite{maskrcnn} and Libra R-CNN \cite{pang2019libra} using the same ResNeXt-101-FPN backbone.

\begin{table*}[!t]
	\tablestyle{2pt}{1.1}
	\centering
	\begin{tabular}{   l  |c|x{22}x{22}x{32}|x{32}x{32}x{32}}
		& Backbone
		& $\mathrm{AP^{box}}$ & $\mathrm{AP^{box}_{50}}$ & $\mathrm{AP^{box}_{75}}$ &$\mathrm{AP^{mask}}$ &$\mathrm{AP^{mask}_{50}}$ &$\mathrm{AP^{mask}_{75}}$  \\ [.1em]
		\shline
		\emph{One-stage detectors} & & & & & & & \\
		~YOLOv3 \cite{redmon2018yolov3} & Darknet-53 & 33.0 &57.9 &34.4 & -& - & - \\
		~SSD513 \cite{liu2016ssd} & ResNet-101-SSD & 31.2 & 50.4 & 33.3 & - & - & - \\
		~DSSD513 \cite{fu2017dssd} & ResNet-101-DSSD & 33.2 & 53.3 & 35.2 & - & - & - \\
		~{RetinaNet} \cite{lin2017focal} & ResNeXt-101-FPN &40.8 &61.1 &44.1 & & & \\
		~CornerNet \cite{law2018cornernet} & Hourglass-104 &42.2 &57.8 &45.2 & & &    \\
		\hline
		\emph{Two-stage detectors} & & & & & & & \\
		~Faster R-CNN+++ \cite{resnet} & ResNet-101-C4 & 34.9 & 55.7 & 37.4 &- &- & -\\
		~Faster R-CNN w FPN \cite{fpn} & ResNet-101-FPN & 36.2 & 59.1 & 39.0 & - & - & -\\
		~R-FCN \cite{dai2016r} & ResNet-101 & 29.9 & 51.9  & - & - & - & - \\
		~Mask R-CNN \cite{maskrcnn} & ResNet-101-FPN &40.2 &61.9 &44.0 &36.2  &58.6 &38.4 \\
	    ~Mask R-CNN \cite{maskrcnn}  & ResNeXt-101-FPN &42.6 &64.9 &46.6 &38.3  &61.6 &40.8 \\
	    ~Libra R-CNN \cite{pang2019libra}& ResNetX-101-FPN  & 43.0 &64.0 &47.0 & - & - & - \\
		\hline 
		~\textbf{DGMN} (ours) & ResNeXt-101-FPN&\bf 44.3   &\bf 66.8 &\bf 48.4  &\bf 39.5 &\bf 63.3 &\bf 42.1 \\
	\end{tabular}
	\caption{
		Object detection and instance segmentation performance using a \textit{single-model} on the COCO \texttt{test-dev} set. 
		We use \textit{single scale} testing.
	}\label{tab:coco_sota}
\end{table*}

\clearpage

%




\end{document}